\documentclass[journal,twoside,final,letterpaper,twocolumn]{IEEEtran}

\ifCLASSINFOpdf
   \usepackage[pdftex]{graphicx}
\else

\fi

\usepackage{epstopdf}
\usepackage{epsfig}
\usepackage{float}
\usepackage{makecell}

\usepackage{amssymb,amsbsy,subfigure,amsmath,amsfonts,multirow,color,url,bm,cite,cases,eqparbox}
\usepackage{algorithmic}
\usepackage{algorithm}
\usepackage[mathscr]{eucal}

\newcommand{\tablesize}{\fontsize{7.5pt}{12pt}\selectfont}

\newcommand{\tablesizethree}{\fontsize{9pt}{12.5pt}\selectfont}

\def\0{\mathbf{0}}
\def\1{\mathbf{1}}

\def\A{\mathbf{A}}

\def\Q{\mathbf{Q}}
\def\P{\mathbf{P}}

\def\S{\mathbf{S}}

\def\X{\mathbf{X}}

\def\Trm{\mathrm{T}}

\def\f{\mathbf{f}}
\def\g{\mathbf{g}}
\def\h{\mathbf{h}}

\def\s{\mathbf{s}}

\def\v{\mathbf{v}}
\def\w{\mathbf{w}}
\def\x{\mathbf{x}}
\def\y{\mathbf{y}}

\def\1{\mathbf{1}_{I\times J}}

\def\I{\mathbf{I}}

\def\Rbb{\mathbb{R}}

\def\Mcal{\mathcal{M}}

\def\Tcal{\mathcal{T}}

\begin{document}

\title{Material Based Object Tracking in Hyperspectral Videos: Benchmark and Algorithms}

\author{Fengchao Xiong, Jun Zhou, and Yuntao Qian
}
 \maketitle

\begin{abstract}
Traditional color images only depict color intensities in red, green and blue channels, often making object trackers fail in challenging scenarios, e.g., background clutter and rapid changes of target appearance. Alternatively, material information of targets contained in a large amount of bands of hyperspectral images (HSI) is more robust to these difficult conditions. In this paper, we conduct a comprehensive study on how material information can be utilized to boost object tracking from three aspects: benchmark dataset, material feature representation and material based tracking. In terms of benchmark, we construct a dataset of fully-annotated videos, which contain both hyperspectral and color sequences of the same scene. Material information is represented by spectral-spatial histogram of multidimensional gradient, which describes the 3D local spectral-spatial structure in an HSI, and fractional abundances of constituted material components  which encode the underlying material distribution. These two types of features are embedded into correlation filters, yielding material based tracking. Experimental results on the collected benchmark dataset show the potentials and advantages of material based object tracking.
\end{abstract}

\section{Introduction}
Object tracking is one of the fundamental tasks in computer vision, particularly for video surveillance and robotics. It is a challenging task to detect the size and location of an object in video frames given a bounding box of the object in the first frame. Many efforts have been made to obtain informative visual cues, such as color intensities~\cite{Bolme2010,Henriques2012}, color names~\cite{Danelljan2014}, texture~\cite{Henriques2015,Danelljan2015} and more for object tracking. However, tracking in traditional color videos has its inherent limitations. Trackers tend to drift in some challenging scenarios, e.g., background clutter, similar color between tracking target and background, object rotation and deformation~\cite{Mueller2017, Danelljan2016a, Zhang2017, Choi2016, Gao2017, Feng2019}, as also exemplified by the recent failure of autonomous cars\footnote{https://www.tesla.com/blog/tragic-loss}. In these cases, however, the underlying material information of foreground and background is distinct, which can be used for distinguishing the target from the surrounding environment~\cite{Liang2018, Schwartz2013}.

\begin{figure}[!tp]
  \centering
\subfigure[RGB]{\includegraphics[width=0.46\linewidth, height=0.32\linewidth]{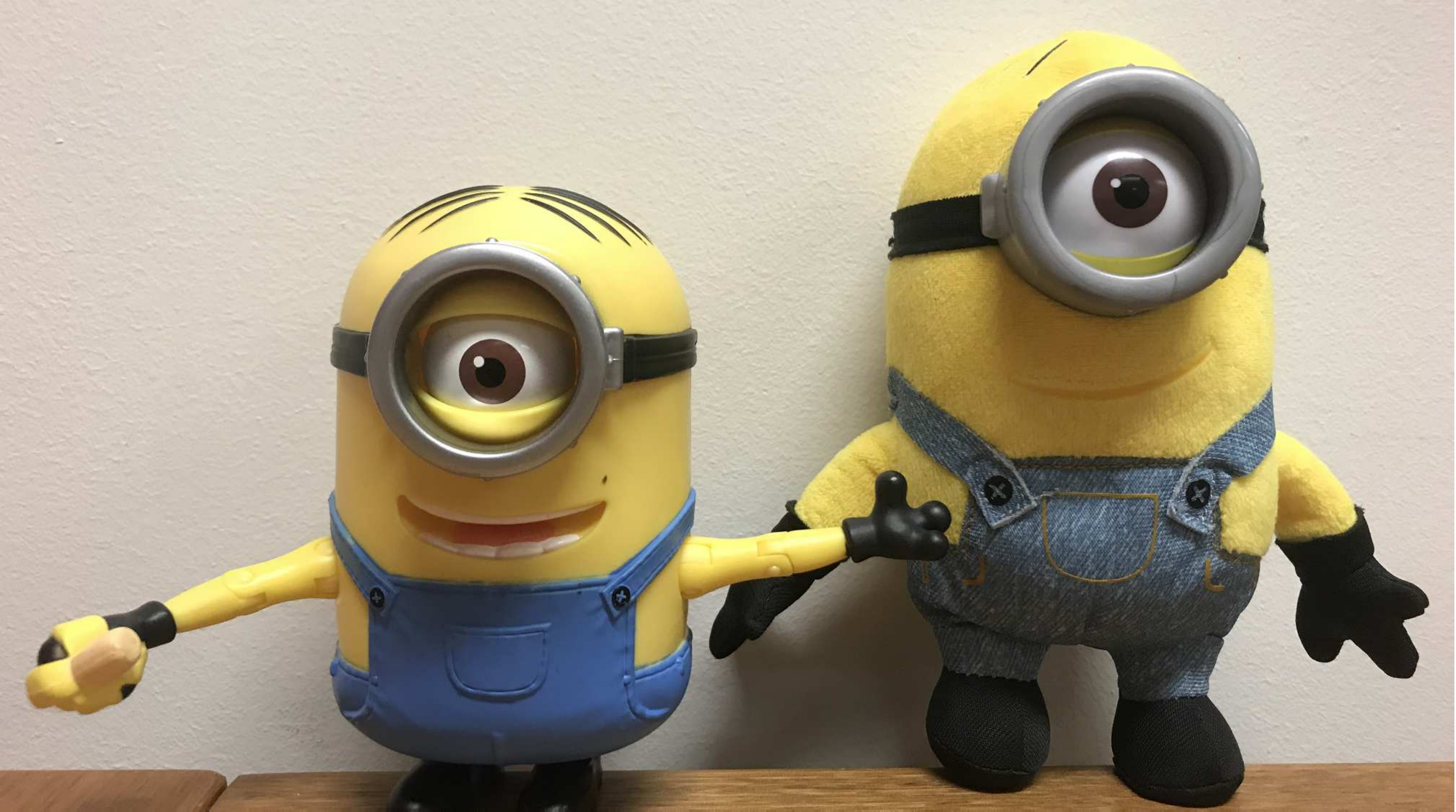}} \label{fig:hsi1}
\subfigure[HSI]{\includegraphics[width=0.46\linewidth, height=0.32\linewidth]{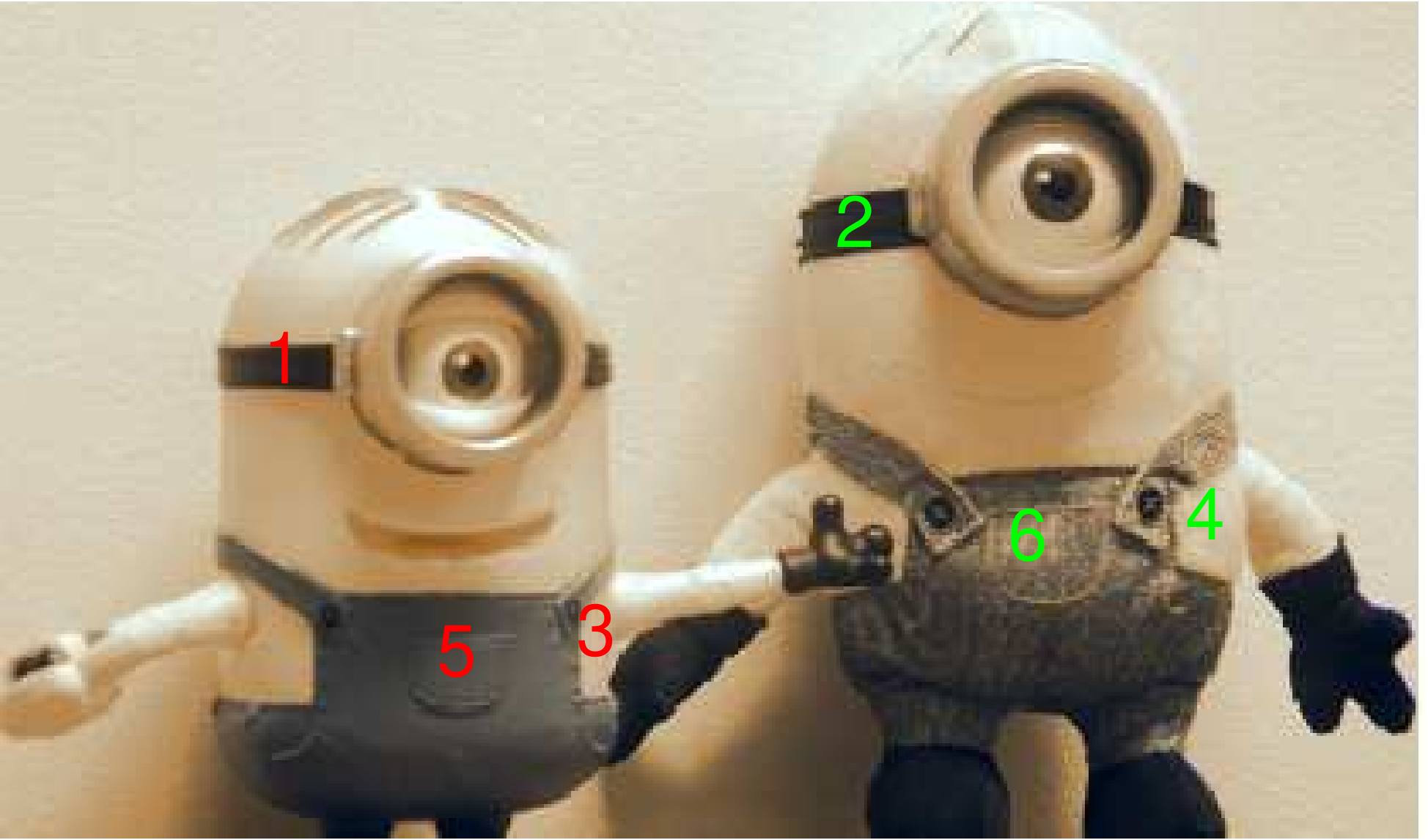}}\label{fig:hsi2}\\
\subfigure[Spectrum (1, 2)]{\label{fig:hsi3}\includegraphics[width=0.31\linewidth]{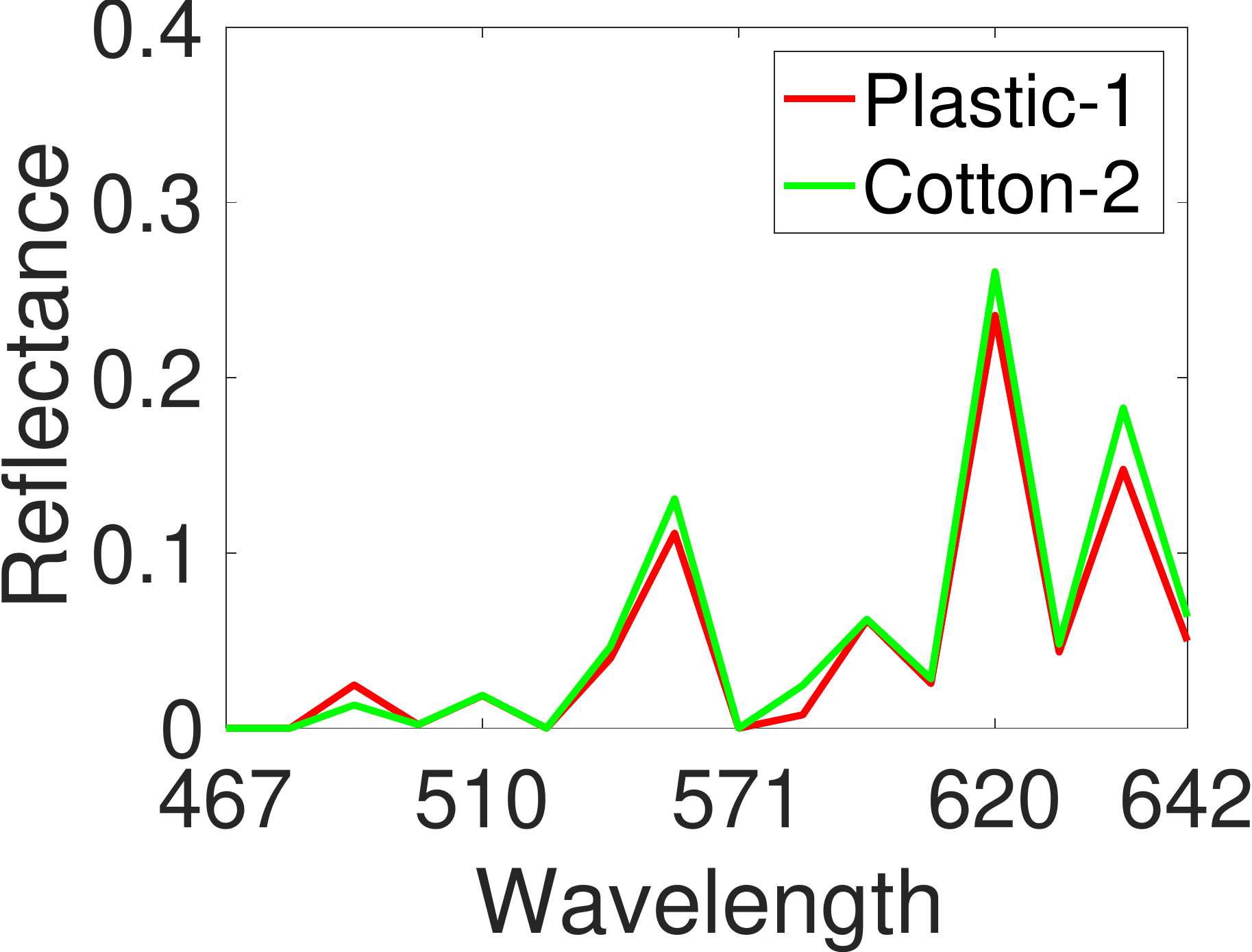}}
\subfigure[Spectrum (3, 4)]{\label{fig:hsi4}\includegraphics[width=0.31\linewidth]{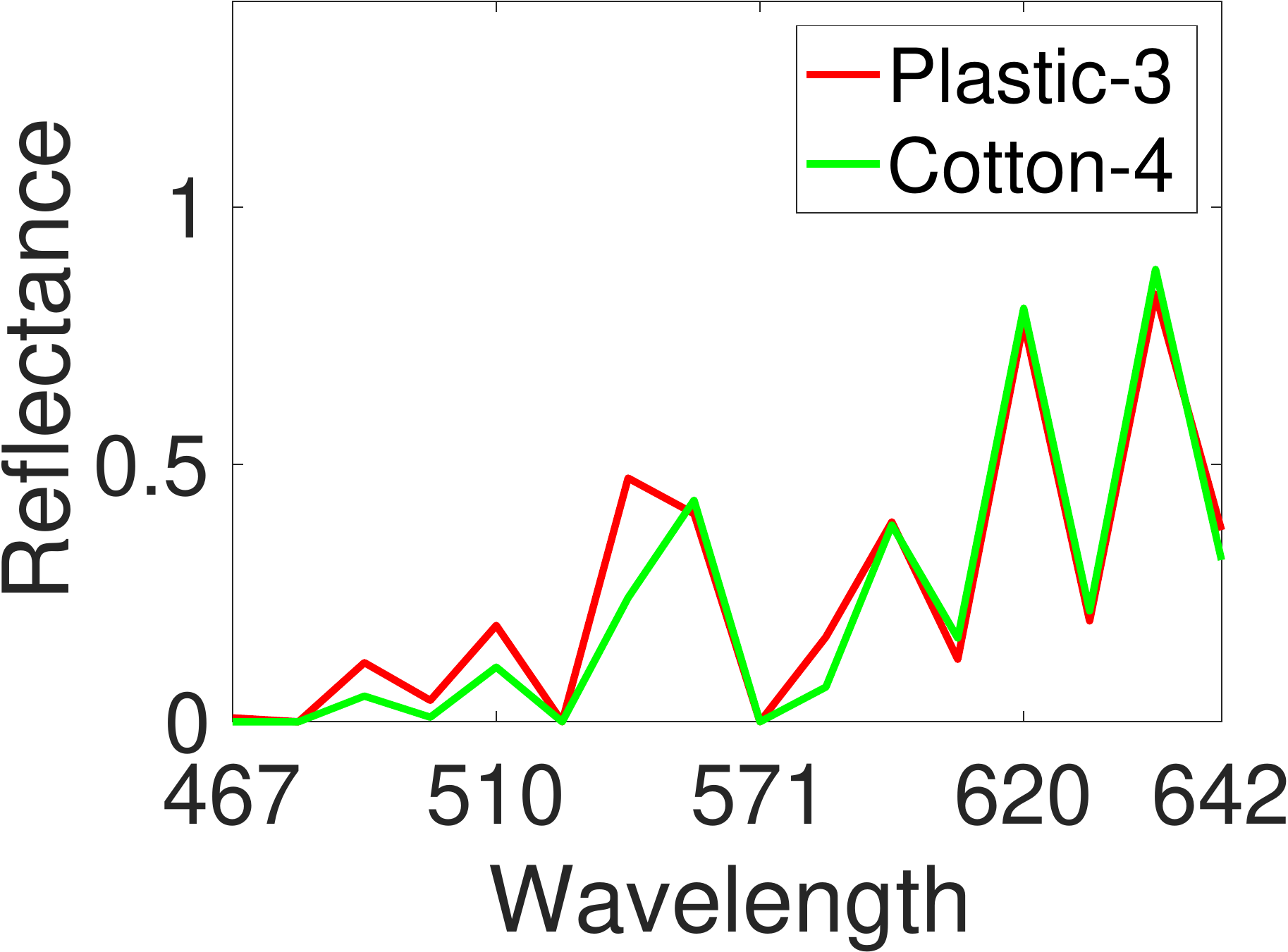}}
\subfigure[Spectrum (5, 6)]{\label{fig:hsi5} \includegraphics[width=0.31\linewidth]{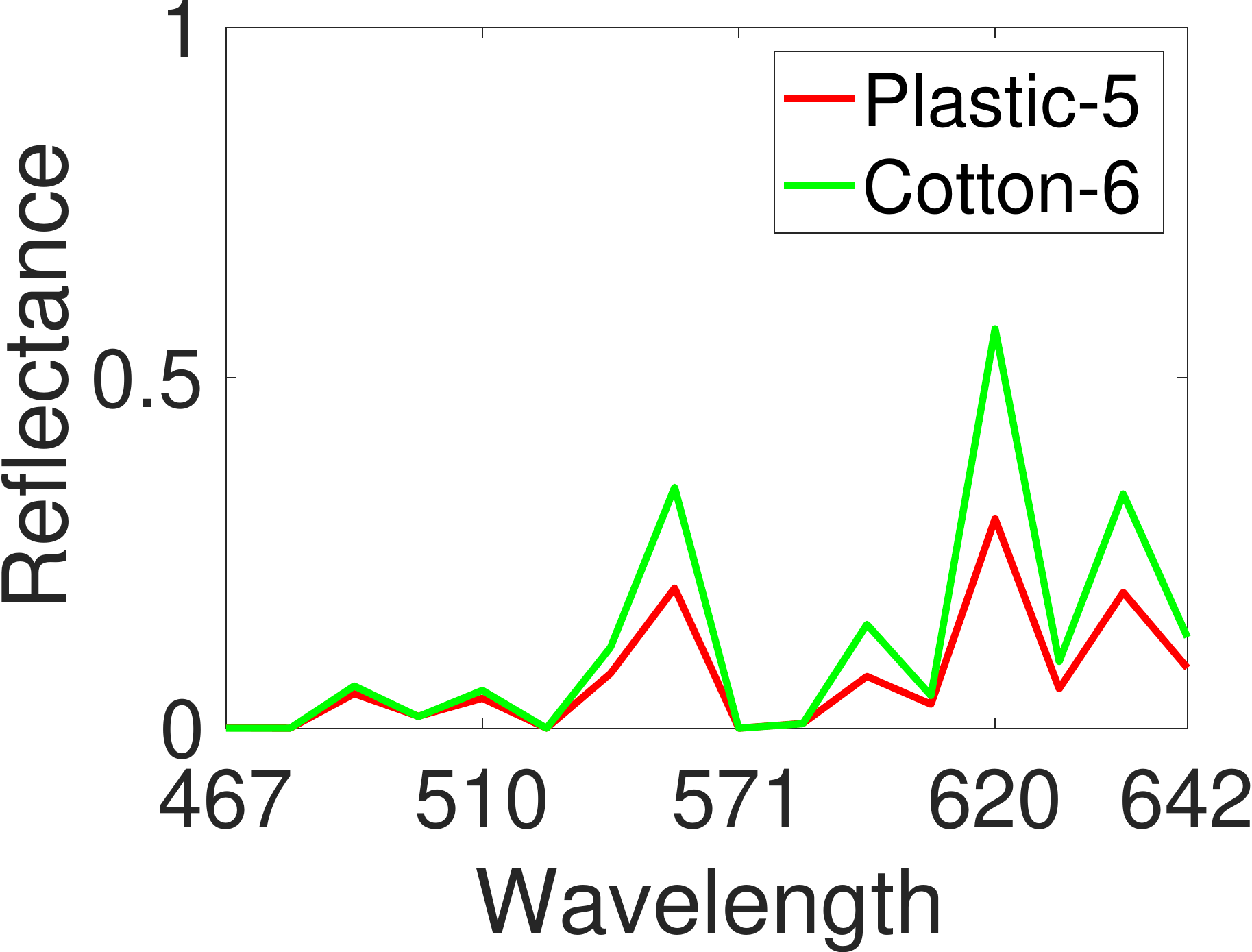}
}
\caption{An example of HSIs for material identification. (a) shows plastic (left) and cloth (right) toys. (b) shows their corresponding false-color image generated from an hyperspectral image. (c)-(e) demonstrate the spectral responses at several pixels. Though these pixels are similar in color, their spectra are different.} \label{fig:hsi}
 \end{figure}

Material information can be broadly obtained by two groups of methods, discriminative and generative. Discriminative approaches extract material information using visual appearance, e.g., color, shape, local textures, etc., in a single RGB image~\cite{Schwartz2015, Schwartz2013, Degol2016, Bell2015, Cimpoi2015}. Such methods fail to identify similar appearances of different materials~\cite{Schwartz2013}, for example artificial plastic fruits and real fruits, due to the limitation of color images in depicting the full physical property of surface reflectance. Alternatively, generative methods represent the material information based on the intrinsic reflectance of the material. Therefore, optical images such as spectral images and time-of-flight images are vastly used thanks to more captured information on the material of the recorded objects~\cite{Shiradkar2014,TominagaandTakahikoHoriuchi2010, Tanaka2017, Su2016}.  Hyperspectral images (HSI) is such a spectral image that records continuous spectrum information instead of monochrome or color intensities on each object. The spectrum information provides details on the material constitution of contents in the scene and increases the inter-object discrimination capability~\cite{Uzair2015}. Fig.~\ref{fig:hsi} shows a sample HSI for material identification from two toy minions built from different materials. Though pixels 1 and 2 are both black in color, their recorded spectral responses are different. The same observation can be made from Figs.~\ref{fig:hsi4} and~\ref{fig:hsi5}. Benefit from the superior advantages of material identification along the spectral dimension, HSIs have enabled many unique applications in remote sensing~\cite{Ye2017} and computer vision~\cite{Liang2018,Uzair2015,Al-khafaji2018}.
				
Despite the material identification ability of HSIs, how to effectively extract and use this information for visual object tracking is faced with many challenges, making this topic barely investigated. First, the high dimensionality nature of HSIs caused by large number of spectral bands and lower spatial resolution because of the imaging mechanism bring difficulties for robust material information extraction. Traditional feature extractors developed for monochrome or color images may not provide highly discriminative material information for hyperspectral data because of the ignorance of valuable spectral information. Pixel-wise spectral reflectance was adopted as the feature for object tracking in early attempts~\cite{Wang2010, Banerjee2009, Nguyen2010,Uzkent2016a}, but the spatial structure is ignored. Alternatively, Uzkent \emph{et al.}~\cite{Uzkent2018} proposed a deep kernelized correlation filter based method (DeepHKCF) for aerial object tracking at the sacrifice of valuable spectral information, in which an HSI was converted to false-color image before passing to a deep convolutional neural network. Qian \emph{et al.}~\cite{Qian2018} selected a set of patches as convolutional kernels for each band to extract features, but the correlations among bands were neglected. It is known that local structure inside an object region facilitates exploiting material information~\cite{Hu2011, Bell2015}. Therefore, spectral-spatial feature extraction methods, which simultaneously consider local spatial information, local spectral information and joint local spectral-spatial information in an HSI, are more favorable for material information representation.
	
Second, there is a lack of benchmark datasets with high diversity to support hyperspectral object tracking. One reason is that with the limitation of most existing hyperspectral sensors, it is difficult to collect real-time hyperspectral videos at high frame rate with high signal-to-noise ratio and high spatial and spectral resolutions. Recently, Uzkent \emph{et al.}~\cite{Uzkent2018} introduced a synthetic aerial dataset generated by Digital Imaging and Remote Sensing (DIRSIG) software~\cite{Uzkent2018} at 1.42 fps. However, the objects are described at very low frame rate and in the remote sensing setting, making it difficult to cover the challenges in close-range computer vision setting, i.e. rotation, deformation, illumination variation, etc.

\begin{figure*}[t]
    \centering
    \graphicspath{{figure/}}
    \includegraphics[width=0.7\linewidth]{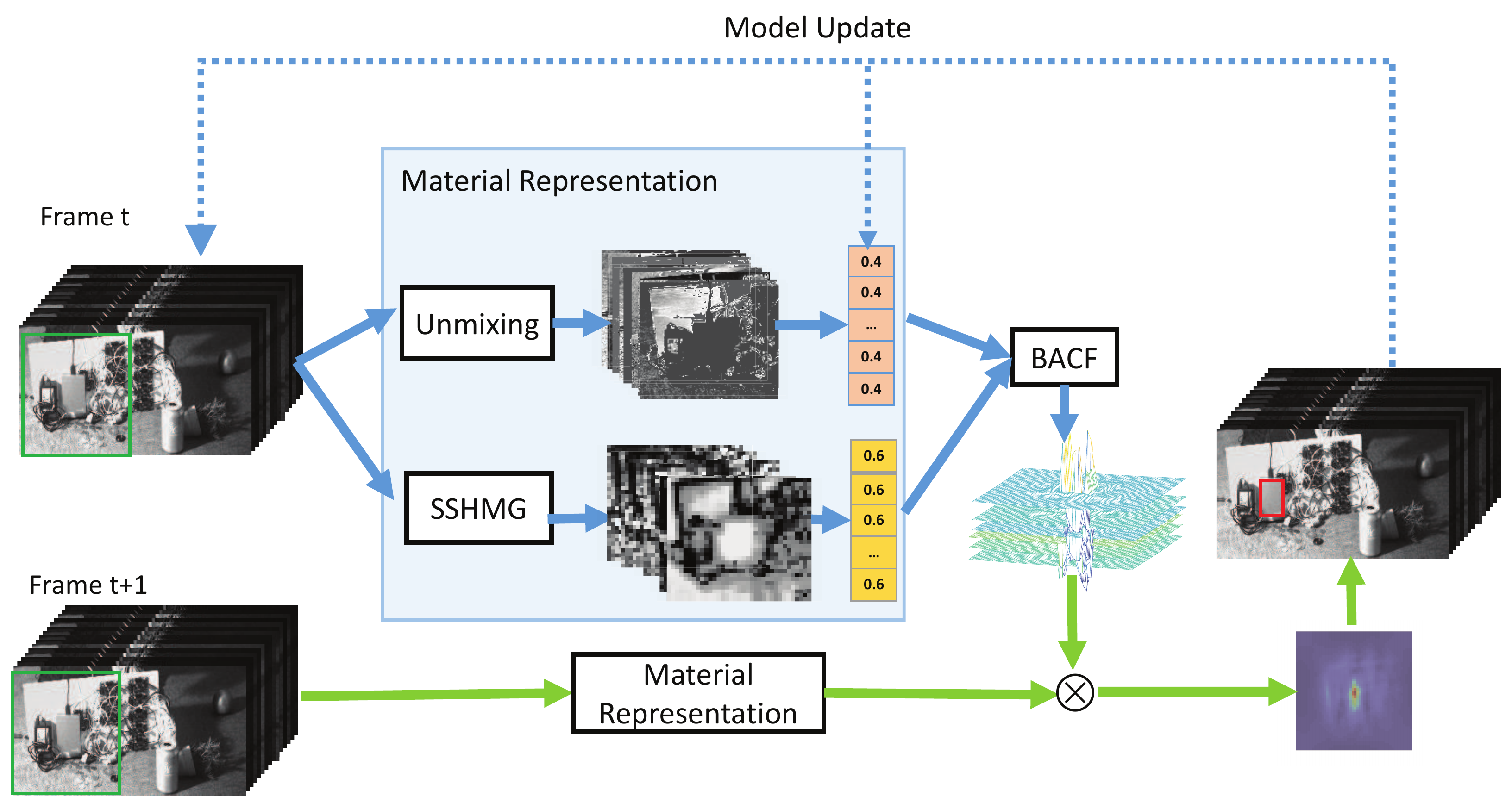}
    \caption{Framework of the proposed MHT tracker. SSHMG and abundances are first calculated, then weighted according to their reliability in order to learn correlation filters. These filters are used to convolve with the candidate patches, yielding multi-channel responses. The sum of the responses is calculated to get the final response map, whose maximum value indicates the predicted location of the object. Finally, the reliability of each feature is updated according to the responses, and the filters are updated by BACF.
    }\label{fig:msshogFramework}
\end{figure*}

In this paper, we tackle object tracking problem from a new perspective in which the material properties are considered. The framework of the proposed material based object tracking is shown in Fig.~\ref{fig:msshogFramework}.  Our original and novel contributions lie in three aspects. First, we develop two feature extractors, local spectral-spatial histogram of multidimensional gradients (SSHMG) and spatial distribution of materials, to capture the material properties in an HSI. SSHMG captures the local spectral-spatial texture information in terms of the spatial and spectral gradients orientations. The distributions of underlying constitute materials in the scene are encoded by abundances. Abundances behave much more like bag-of-words where it implicitly discover recognition-related structures in the spectral space, making it superior to spectral representation. They are obtained by hyperspectral unmixing which decomposes an HSI into constitute spectral (or endmembers) and their corresponding fractions (or abundances). Moreover, we develop an online learning approach to adaptively adjust their relative importance in object state translation. Second, to evaluate the material based tracking, we introduce a fully annotated dataset with 35 hyperspectral videos captured by a commercial high-speed hyperspectral camera. To our best of knowledge, this is the first large-scale fully-annotated  hyperspectral video dataset in close-range computer vision setting. We also provide color videos captured and annotated on the same scene. Third, extensive experiments with detailed analysis are carried out to explore a better understanding of how material information can be used to promote object tracking. The experimental results show that the proposed material-based tracking even outperforms several state-of-the-art deep learning based trackers. Given increasing adoption of HSIs in computer vision tasks, we expect our dataset and material based tracker will provide baselines for future research on object tracking.

\begin{table*}[t]
\centering

\caption{Distribution of dataset attributes}\label{tab:attribute}
\centering
\tablesizethree{
\begin{tabular}{lccccccccccc}
\hline
\hline
Attributes &\textbf{SV}&\textbf{MB}&\textbf{OCC}&\textbf{FM}&\textbf{LR}&\textbf{IPR}&\textbf{OPR}&\textbf{DEF}&\textbf{BC}&\textbf{IV}&\textbf{OV}\\
\hline
 \textbf{SV}&\textbf{22}&4&9&2&1&9&14&5&7&7&2\\
\textbf{MB}&4&\textbf{5}&2&1&1&2&3&1&1&0&0\\
\textbf{OCC}&9&2&\textbf{14}&2&1&2&4&1&5&5&2\\
\textbf{FM}&2&1&2&\textbf{4}&2&1&1&0&2&1&0\\
\textbf{LR}&1&1&1&2&\textbf{3}&0&0&0&2&0&0\\
\textbf{IPR}&9&2&2&1&0&\textbf{15}&14&4&6&1&0\\
\textbf{OPR}&14&3&4&1&0&14&\textbf{19}&7&9&2&2\\
\textbf{DEF}&5&1&1&0&0&4&7&\textbf{7}&3&0&0\\
\textbf{BC}&7&1&5&2&2&6&9&3&\textbf{17}&1&1\\
\textbf{IV}&7&0&5&1&0&1&2&0&1&\textbf{7}&1\\
\textbf{OV}&2&0&2&0&0&0&2&0&1&1&\textbf{2}\\
\hline
\hline
\end{tabular}}
\end{table*}

\section{Hyperspectral Tracking Benchmark Dataset}\label{sec:data}
In this section, we provide the details about the hyperspectral tracking dataset including data collection, data annotation and attribute statistics.
\subsection{Data Collection}
Recent progress on sensors makes it possible to collect hyperspectral sequences at video rate. In our research, a snapshot mosaic hyperspectral camera\footnote{https://www.imec-int.com/en/hyperspectral-imaging} was used to collect videos. This camera can acquire videos up to 180 hyperspectral cubes per second, each of which contains $512\times256$ pixels and 16 bands in the wavelength from 470nm to 620nm. In our data collection, we captured videos at 25 frames per second, where a frame refers to a 3D hyperspectral cube with two dimensions indexing the spatial location and the third dimension indexing the spectral band. For fair comparison with color-based tracking methods, RGB videos were also acquired at the same frame rate in a very close view point as the hyperspectral videos. The preprocessing steps include spectral calibration, image registration and color conversion. Their details are described in the appendix.

\subsection{Data Annotation}
We manually drew a single upright bounding box for each frame to obtain the ground truth location of the target. The bounding box is represented by the centre location and its height and width. Though we tried our best to make the hyperspectral video and color video capture the same scene, the detailed location of the target may differ slightly in two types of videos. To make the ground truth as precise as possible, we labelled hyperspectral and color videos independently. Each video was labelled by three volunteers and one of them was an expert to check the bounding boxes and adjust them if necessary. With the above strategy, the quality of data labels is guaranteed.

\subsection{Data Attributes}
The whole dataset contains 35 color videos and 35 hyperspectral videos with an average of 500 frames in each sequence. Following the challenging factors listed in~\cite{Wu2015}, we carefully collected videos to include multiple target categories, diverse scenarios, rich activities and diverse content etc., guaranteeing the generality and complexity of the dataset. For example, the tracking objects are of high diversity, including vehicles, faces, people, generic objects, animals, etc.  Additionally, we collected objects in different scenes (indoor and outdoor) and different viewpoints.

Each video is also labelled with associated challenging factors out of eleven attributes~\cite{Wu2015}, including illumination variation (IV), scale variation (SV), occlusion (OCC), deformation (DEF), motion blur (MB), fast motion (FM), in-plane rotation (IPR), out-of-plane rotation (OPR), out-of-view (OV), background clutters (BC), and low resolution (LR). Table~\ref{tab:attribute} shows the number of coincident attributes across all videos.  As can be seen, all the tracking factors are considered and the most common challenging factors are SV, OPR, IPR, BC and OCC. These factors remain difficult for color tracking. With the usage of the material identification ability of HSIs, it is expected that our dataset will enable improved tracking performance in some scenes, such as IPR, OPR and BC, where the shape of objects changes but not the materials.

\section{Material Based Tracking} \label{sec:method}
In this section, we give the details of the proposed material based tracking method, including material feature representation and feature reliability learning.

\subsection{Spectral-spatial Histogram of Multidimensional Gradients}

\begin{figure*}[t]
    \centering
    \graphicspath{{figure/}}
    \includegraphics[width=0.7\linewidth]{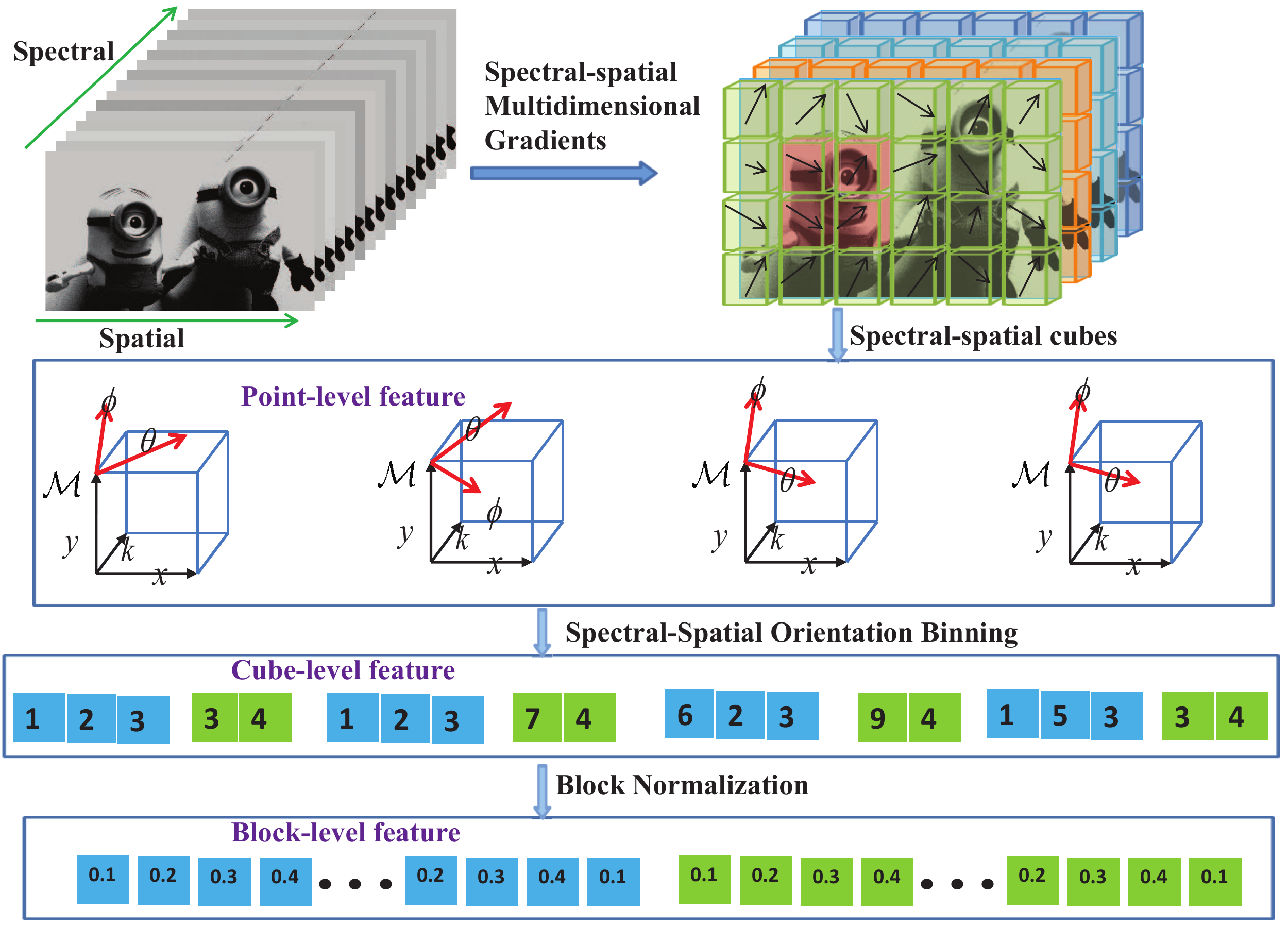}
    \caption{Framework of the proposed SSHMG. We first calculate the spectral-spatial multidimensional gradients in an HSI in both spatial and spectral directions, which is represented by $(\Mcal, \theta, \phi)$ in a spherical coordinate. After that, all the points in a cube are aggregated in spatial and spectral orientations, yielding cube-level features. Finally the cube-level features are normalized in a local block, resulting block-level features. }\label{fig:sshogFramework}
\end{figure*}

As analysis earlier, an HSI is a 3D cube, comprising two spatial dimensions and one spectral dimension.  Therefore, it can be naturally modelled by a third-order tensor without structural information loss. Previous works show that 3D spectral-spatial features are effective for HSI processing~\cite{Qian2013, Jia2017}. Instead of 2D patches, these features are constructed in a local 3D spectral-spatial  neighborhood. Therefore, spectral information, spatial information and joint spectral-spatial information are simultaneously taken into consideration. In addition, since local information is beneficial to object tracking~\cite{Sun2018, Sui2018, Choi2018} and material information representation~\cite{Schwartz2015, Schwartz2013}, we build a 3D local spectral-spatial histogram of multidimensional gradient (SSHMG) descriptor for an HSI to exploit local information of materials, which is also less sensitive to illumination condition. As shown in Fig.~\ref{fig:sshogFramework}, given an HSI $\Tcal \in \Rbb^{W \times H \times K}$ containing $W \times H $ pixels and $K$ bands, SSHMG is constructed as follows.

First, we convolve $\Tcal$  in three modes using a finite difference filer [-1, 0, 1] to obtain its multidimensional gradients $\nabla \Tcal_x, \nabla \Tcal_y, \nabla \Tcal_k$. Transferring the gradients to a spherical coordinate system, they can be identically represented by $(\Mcal, \theta, \phi)$, where $\Mcal$ represents the spectral-spatial  multidimensional gradient magnitude, $\theta$ indicates the spatial gradient orientation, and $\phi$ denotes the spectral gradient orientation. Their definitions are given as follows:
 \begin{equation}
 \begin{split}
 		&\Mcal(x, y, k)=\sqrt{\nabla \Tcal_x^2+\nabla \Tcal_y^2+\nabla \Tcal_k^2}\\
 		&\theta(x, y, k)=\arctan\bigg(\frac{\nabla \Tcal_y}{\nabla \Tcal_x}\bigg)\\
 		&\phi(x, y, k)=\arctan\bigg(\frac{\nabla \Tcal_k}{ \sqrt{\nabla \Tcal_x^2+\nabla \Tcal_y^2}}\bigg) \\
 \end{split}
\end{equation}
Subsequently, the tensor is divided to a number of 3D cubes with the size of $z\times z \times z$. For each position in a local cube, its gradient orientation is then quantized in both spatial dimension ($B_{\theta}$) and spectral dimension ($B_{\phi}$), i.e.,
\begin{equation}
\begin{split}
	B_{\theta}(x,y,k)&= \text{round} \left(\frac{n_\theta \theta(x, y, k)}{2\pi}\right) \text{mod}\: n_\theta\\
	B_{\phi}(x,y,k)&=\text{round}\left(\frac{n_\phi \phi(x, y, k)}{\pi} \right) \text{mod}\: n_\phi
\end{split}
\end{equation}where $n_\theta$ and $n_\phi$ respectively denote the number of bins in the spatial and spectral directions. Based on such quantization, we can define point-level spatial feature $F_{\theta}(x, y, k)$ and spectral feature $F_{\phi}(x, y, k)$ for each position as follows:
\begin{equation}
	F_{\theta}(x, y, k)_b=\left\{
             \begin{array}{lr}
             \Mcal(x, y, k), \: \mathrm{if}\: b=B_\theta(x, y, k) \\
          \qquad 0, \:   \mathrm{otherwise}
             \end{array}
             \right.
\end{equation}

\begin{equation}
	F_{\phi}(x, y, k)_b=\left\{
             \begin{array}{lr}
             \Mcal(x, y, k), \: \mathrm{if}\: b=B_\phi(x, y, k) \\
          \qquad 0, \:   \mathrm{otherwise}
             \end{array}
             \right.
\end{equation} where $b$ indexes gradient orientation.
Then we aggregate point-level features to obtain cube-level features $C_{\theta}(i, j, s)$ and $C_{\phi}(i, j, s)$ by summarising point-level features in a cube, where $0 \leq i \leq \lfloor (W-1)/z \rfloor$, $0 \leq j \leq \lfloor  (H-1)/z  \rfloor$ and $0 \leq s \leq \lfloor (K-1)/z  \rfloor$.

Afterwards, we get the block-level feature $\v$ by concatenating the cube-level feature over overlapping $2\times2$ spatial blocks to reduce the sensitivity of feature descriptor to illumination and foreground- background contrast.  $\v$ is then normalized to unit length by dividing with its $L_2$ norm. An element-wise truncation function $L_{\alpha}$ is applied to the normalized block-level feature $\v$ to limit the maximum values in a feature vector to $\alpha$, which is then renormalized. With such step, the proposed description is less sensitive to the illumination changes. Finally, the block-level features are concatenated in the spectral direction, yielding proposed SSHMG.

\subsection{Material Distribution Learning}
\begin{figure}[!htp]
  \centering
\subfigure[RGB]{\includegraphics[width=0.45\linewidth, clip=true]{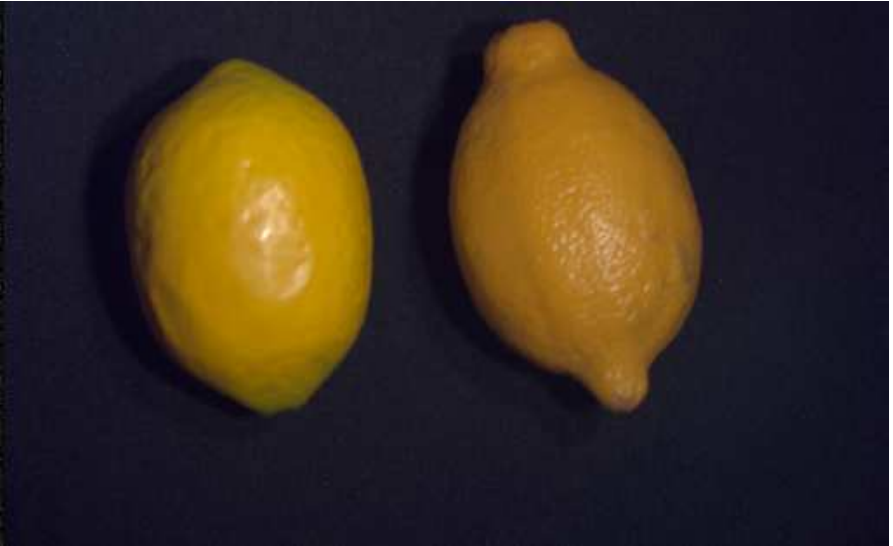}}
\subfigure[Plastic]{\includegraphics[width=0.45\linewidth, clip=true]{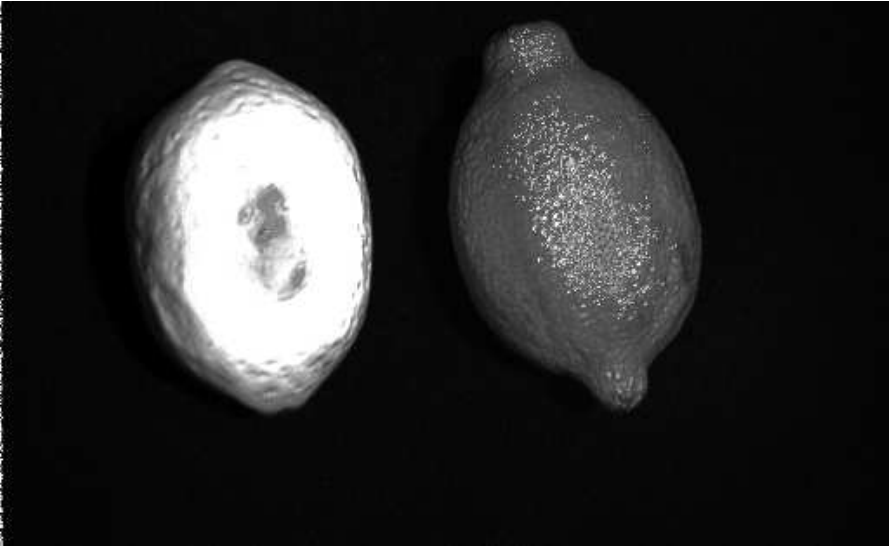}}\\
\subfigure[Background]{\includegraphics[width=0.45\linewidth, clip=true]{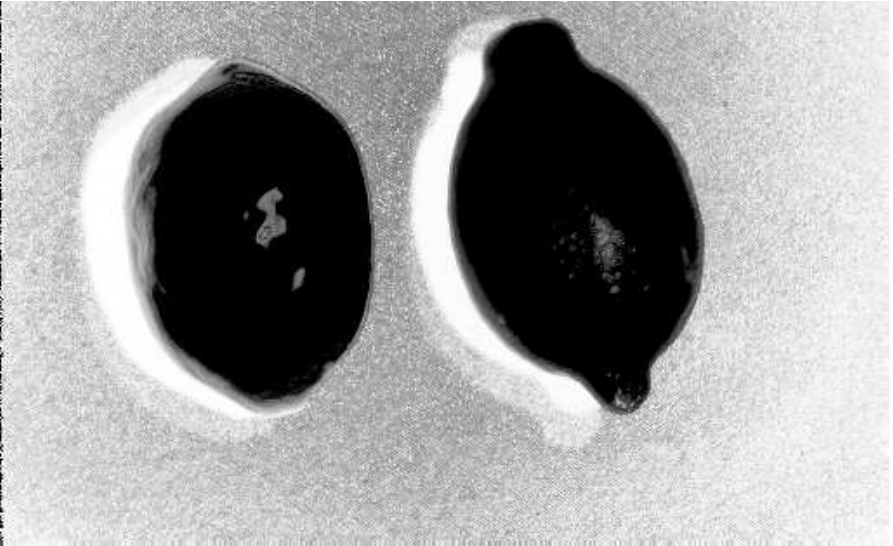}}
\subfigure[Real]{\includegraphics[width=0.45\linewidth, clip=true]{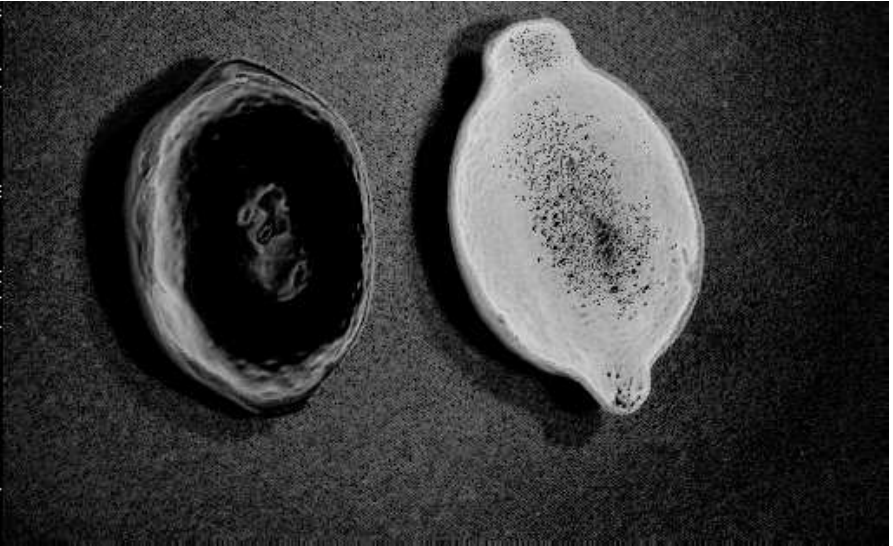}}
\caption{Unmixing results of a hyperspectral scene. (a) shows the color images of two lemons. The left lemon is plastic and the right one is real. (b)-(d) give the abundances generated by hyperspectral unmixing. } \label{fig:unmixing}
 \end{figure}

Here, we introduce a useful feature, abundances from hyperspectral unmixing, to describe detailed material distribution in a scene. Low spatial resolution of sensor and long distance from camera to imaging targets cause overlapped spectral responses of neighboring but different surface materials, leading to ``mixed" pixels. Hyperspectral unmixing decomposes these pixels into a collection of spectral signatures, or endmembers, and associated proportions or abundances at each pixel. Fig.~\ref{fig:unmixing} provides an example of hyperspectral unmixing. In this scene, the abundances produced by hyperspectral unmixing clearly demonstrate the underlying distributions of three materials, namely, plastic lemon, background, and real lemon.

Modern unmixing methods can be broadly categorised into three groups, supervised, unsupervised  and semi-supervised. Supervised unmixing breaks this problem into two tasks, endmember extraction~\cite{Nascimento2005, Chang2010} and abundance estimation~\cite{Chouzenoux2014,Heylen2011}. Despite fast speed, supervised methods fail to deal with  complicated situations where pure endmembers are unavailable. Alternatively, unsupervised methods integrate both tasks into a single problem to simultaneously estimate endmembers and abundances~\cite{Wang2015c,Qian2017}, but their performance is limited by time-consuming optimization steps. Recently, semi-supervised unmixing has been introduced which assumes the endmembers are a part of a predefined spectral library~\cite{Iordache2014,Zhang2018a}. Since the endmembers are selected from the library, the unmixing results are more accurate and closer to the spectral signatures of real-world materials.

Considering the requirements on tracking speed and endmember accuracy, we adopt semi-supervised methods to select the endmembers from an offline library built from a large set of HSIs. These endmembers are used for abundance estimation in all subsequent frames. The spectral library is constructed using K-means method to cluster the endmembers extracted by vertex component analysis (VCA)~\cite{Nascimento2005} from a collection of HSIs into a set of clusters. The spectral reflectance of each cluster center is considered as an atom in the library. In terms of endmember selection, let $\X\in \Rbb^{L\times N}$ be the target object in the bounding box, covering $L$ bands and $N$ pixels, the endmembers are selected using CLSUnSAL~\cite{Iordache2014} by enforcing group sparsity on the abundance matrix $\S$, which is formulated as:
	\begin{equation}
	\min_{\X} \|\X-\A\S\|_F^2+\lambda \|\S\|_{2,1}  \quad s.t.  \quad \S \geq 0
	\end{equation}
where $\A\in \Rbb^{L\times M }$ is the predefined spectral library with $M$ materials. When the CLSUnSAL is converged, the sum of each row of $\S$ is calculated, yielding a vector $\s$ with $M$ entries. Each element in $\s$ can be regarded as the total contributions of a specified material in $\X$. The spectral signatures of top $R$ elements are then selected, determined by HySime~\cite{Bioucas-Dias2008}, whose entries are considered as the constitute endmembers in the scene. Once the endmembers are available, a simplex-projection unmixing (SPU)~\cite{Heylen2011} method is used for abundance estimation because of its computational efficiency and superior performances.

\subsection{Material Based Object Tracking Method}
Discriminative correlation filter (DCF) is widely used in object tracking due to its competitive performance and computational efficiency enabled by fast Fourier transform (FFT). DCF produces filters by minimizing the output sum of squared error~\cite{Bolme2010} for all circular shifts of a training sample. The periodic assumption on training samples causes unwanted boundary effects, which can be alleviated by adding spatial regularization~\cite{Danelljan2015, Galoogahi2017, Lukezic2017, Mueller2017}. Concretely, background-aware correlation filter (BACF) considers all background patches as negative samples by using a rectangular mask covered central part of the circular samples~\cite{Galoogahi2017}. Benefit from alternating direction method of multipliers (ADMM) and FFT, BACF is also computationally efficient.

We adopt BACF~\cite{Galoogahi2017} as our tracking method considering its computational efficiency. Under the framework of BACF, our material based tracking learns the filters $\f_k$ by the following objective function:
\begin{equation}
\begin{split}
		\min_{\f}\frac{1}{2} \sum_{d=1}^{D} \| \y^d- \sum \limits_{i=1}^{N} w_i\sum_{k=1}^{K_i}\f_k^{\Trm} \P \x_k^{d} \|_2^2+
		\frac{\lambda}{2}\sum \limits_{i=1}^{N}  \sum_{k=1}^{K_i}\|\f_k\|_2^2 \label{eq:bacf}
\end{split}
\end{equation}
where $\x$ is the material feature, represented by SSHMG and abundances, $w_i$ contains their weights for determining the location of target, $\P$ is a binary matrix which crops the central patch of $\x_k$, $\otimes$ denotes the spatial convolution operator, $\lambda \geq 0$ is the regularization parameter, and $K_i$, $D$ represents feature dimensionality and number of pixels, respectively.

Taking advantages of FFT, the problem in  Equation~(\ref{eq:bacf}) can be reformulated in the frequency domain as:
\begin{equation}
	\min_{\f}\frac{1}{2}\|\hat{\y}-\hat{\X}\sqrt{D}(\Q\P^{\Trm}\otimes \I_K)\f\|_2^2+\frac{\lambda}{2}\|\f\|_2^2
\end{equation} where $\hat{\X}$ is a $D\times KD$ matrix concatenated with $K$ diagonal matrices $\mathrm{diag}(w_i\hat{\x_k})^{\Trm}$ and $\f=[\f_1^{\Trm}, \cdots, \f_k^{\Trm}]$. $\Q$ is a orthonormal $D\times D$ matrix, which maps any $D$ dimentional vectorised signal to the Fourier domain. $\I_K$ is a $K\times K$ identity matrix.  $\hat{}$ and $\otimes$  denote Discrete Fourier Transform of a signal and  Kronecker product respectively. By introducing auxiliary matrix $\hat{\g}=\sqrt{D}(\Q\P^{\Trm}\otimes \I_K)\f$, the Augmented Lagrangian function is:
\begin{equation}
\begin{split}
		L(\hat{\g}, \hat{\h}, \hat{\varsigma})&=\frac{1}{2}\|\hat{\y}-\hat{\X}\hat{\g} \|_2^2+\frac{\lambda}{2}\|\f\|_2^2\\
		&+ \hat{\varsigma}^{\Trm}(\hat{\g}-\sqrt{D}(\Q\P^{\Trm}\otimes \I_K)\f)\\
		&+\frac{\mu}{2}\|\hat{\g}-\sqrt{D}(\Q\P^{\Trm}\otimes \I_K)\f\|_2^2
\end{split}
\end{equation} where $\mu$ is regularization factor and $\hat{\varsigma}$ is the multiplier. This problem can be  divided into three subproblems with closed form solutions given by:
\begin{equation}
\begin{split}
	&\f \leftarrow  (\mu+\frac{\lambda}{\sqrt{D}})^{-1}\frac{\P\Q^{\Trm}\otimes \I_K}{\sqrt{D}} (\mu\hat{\g}+\hat{\varsigma})\\
		&\hat{\g}^d\leftarrow \frac{1}{\mu}(D\hat{\y}^d\hat{\x}^d-\hat{\varsigma}^d+\mu\hat{\f}^d)\\
&-\frac{\hat{\x}^d}{\mu (\hat{\x}^d)^{\Trm}\hat{\x}^d+T\mu}(D\hat{\y}^d(\hat{\x}^d)^{\Trm}\hat{\x}^d- (\hat{\x}^d)^T\hat{\varsigma}+\mu \hat{\x}^d\hat{\f} )\\
&\hat{\varsigma} \leftarrow  \hat{\varsigma}+\mu(\hat{\g}-(\Q\P^{\Trm}\otimes \I_K)\f) \label{eq:BACF}
	\end{split}
\end{equation}

Instead of estimating channel-wise reliability in~\cite{Lukezic2017}, we use group-wise reliability to represent the importance of features. Each group of features jointly represent one typical physical property. For example, all the channels in the abundances embody the distribution of underlying materials. Moreover, group-wise reliability differentiates the discriminative power of an individual feature. This enables the tracker to adaptively suppress the effect of less reliable features and enhance learning from more reliable features.

We evaluate the reliability of each feature using its expressiveness in object detection, including overlap reliability score, distance reliability score and self reliability score. The first two scores measure the contributions for final object localization. For overlapping reliability scores, we compute the overlap ratio between the bounding box from an individual feature $B_i^{t} $ and the final bounding box $ B^{t}$:
\begin{equation}
	O(B_i^{t}, B^{t})=\frac{B_i^{t} \cap B^{t}}{B_i^{t} \cup B^{t}}.
\end{equation}
Then, the overlap reliability score is determined by:
\begin{equation}
	O_{i}^{'t}=\mathrm{exp}(-(1-O(B_i^{t}, B^{t}))^2)
\end{equation}
In terms of distance reliability score, we also compare the differences between the predicted central location $C_i^{t}$ from one particular feature and the final position $ C^{t}$, formulated as:
\begin{equation}
	D_{i}^{'t}=\mathrm{exp}\big(-\|C_i^{t}- C^{t}\|_2^2\big).
\end{equation}
In addition, self reliability score measures the trajectory smoothness degree of feature. It records the shift between the previous bounding box $B_i^{t-1}$ and the current bounding box $B_i^{t}$, which is given by:
\begin{equation}
	S_{i}^{'t}=\mathrm{exp}\bigg(\frac{-\| C_i^{t}-C_i^{t-1} \|_2^2}{W(B_i^{t})+H(B_i^{t})}\bigg).
\end{equation}
where $C_i^{t-1}$ is the central location determined by the $i$-th feature at the $(t-1)$-th frame. $W(B_i^{t})$ and $H(B_i^{t})$ represent the width and height of the bounding box, respectively. With the above reliability measure, the final reliability weights of different features are given by:
\begin{equation}
	\w_i^{t}=\frac{O_{i}^{'t}+D_{i}^{'t}+S_{i}^{'t}}{\sum \limits_{i=1}^{N}  O_{i}^{'t}+D_{i}^{'t}+S_{i}^{'t}}\label{eq:weight1}
\end{equation}

The model optimization, target detection and model update  are as follows.

\textbf{Model Optimization}. The feature reliability is initialized with the same value for all the features at the initial frame. After each frame, the feature reliability is computed by Equation~(\ref{eq:weight1}). With these features, the correlation filters are learned by Equation~(\ref{eq:BACF}).

\textbf{Model Detection}. SSHMG and abundances extracted from the candidate area are re-scaled by their reliability weight. To deal with scale changes, multiple resolutions of the candidate  regions are used.  Afterwards, the weighted features are convoluted by filter $\hat{\g}$ to derive the channel-wise responses, which are then summed, resulting the final output response. The interpolation strategy in~\cite{Galoogahi2017} is employed to get the desired location and scale with sub-grid precision.

\textbf{Model Update}. The group reliability is computed by Equation~(\ref{eq:weight1}).  To account for appearance changes, for example scale and pose changes and rotation during tracking, group reliability scores and models are updated by an autoregressive model with learning rate $\eta$, formulated as
\begin{equation}
\left\{
             \begin{array}{lr}
	\w_{model}^{t}=(1-\eta)\w^{t-1}+\eta\w^{t}\\
	\hat{\x}_{model}^{l}= (1-\eta)\hat{\x}^{l-1}+\eta \hat{\x}
	\label{eq:updateM}
	 \end{array}
\right.
\end{equation}

\begin{figure*}[!htp]
  \centering
\subfigure[Precision plot ]{\includegraphics[width=0.3\linewidth, clip=true]{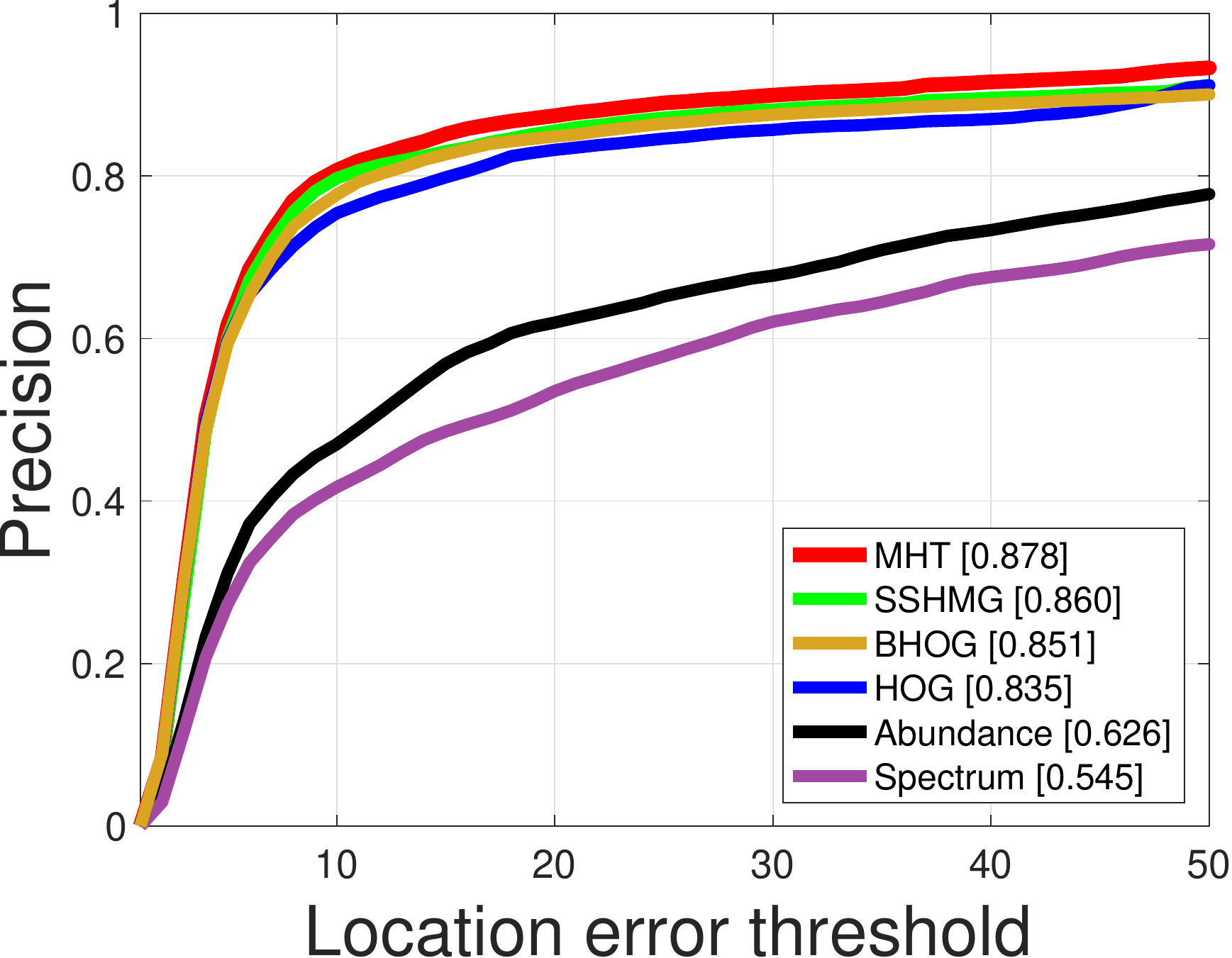}}
\subfigure[Success plot]{\includegraphics[width=0.3\linewidth, clip=true]{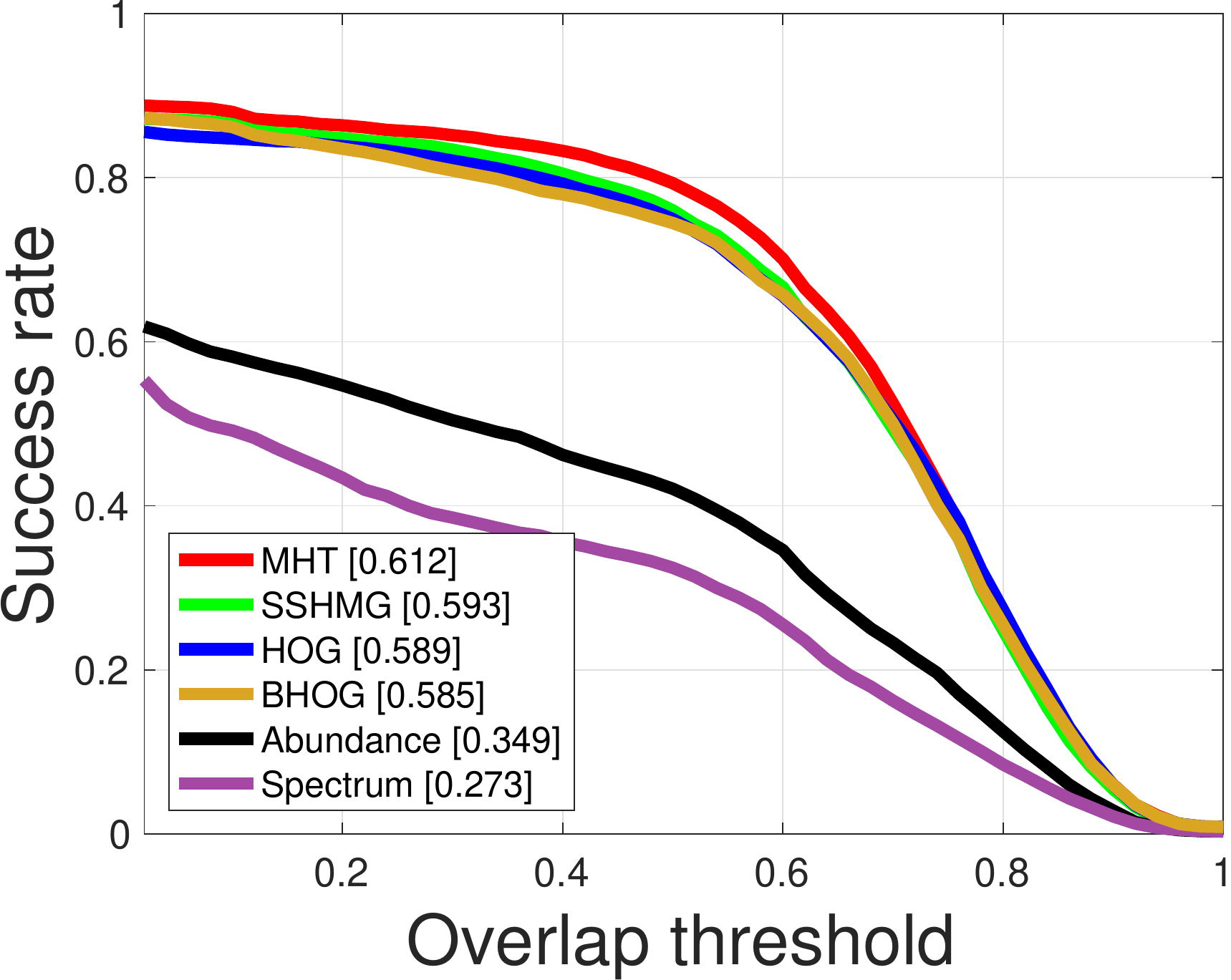}}
\caption{Precision plot and success plot of different features on the collected hyperspectral video dataset. The average distance precision scores at a threshold of 20 pixels and AUCs of the trackers are specified in the legends.} \label{fig:hsiFeature}
 \end{figure*}
\section{Experiments} \label{sec:exp}

In this section, we first investigate our material based tracking method from a feature representation perspective, then compare it with some state-of-the-art methods, including hand-crafted feature based trackers, deep feature based trackers and hyperspectral trackers.  The parameters of competing methods are automatically set as suggested in the original implementation.  Moreover, the attribute-based and quality-based comparison are also presented.

\subsection{Experimental Setting}

In our experiments, the parameters of SSHMG were set as $\alpha=0.2$, $z=4$, $n_{\theta}=9$, $n_{\phi}=4$. The learning rate $\eta$ was set to 0.0023. All the other parameters were set to the same as those in BACF~\cite{Galoogahi2017}. All methods were tested on a Windows machine equipped with Intel(R) Core(TM) i7-7800X CPU@3.50GHz and 128GB RAM.

Three evaluation protocols, including precision plot, success plot and area under curve (AUC), were adopted to describe the performance of all the trackers. Precision plot records the fractions of frames whose estimated location is within a given distance threshold to the ground truth. The average distance precision rate is reported at a threshold of 20 pixels.  Success plot shows the percentages of successful frames whose overlap ratio between the predicted bounding box and ground-truth is larger than a certain threshold varied from 0 to 1. AUC of each success plot is  also selected as an  evaluation measure to rank all trackers.  All the results  are presented one-pass evaluation (OPE), i.e., a tracker is run throughout a test sequence with initialization from the ground truth position in the initial frame.

\subsection{Effectiveness of  Proposed Material Feature}

In this experiment, we compare the effectiveness of five feature extractors, including spectrum, histogram of oriented gradients (HOG), band-wise HOG (BHOG)~\cite{Uzkent2018}, abundances, SSHMG, and SSHMG combined with material abundances (abbreviated as MHT). For spectrum, the raw spectral response at each pixel was employed as the feature for tracking. HOG was calculated following~\cite{Felzenszwalb2010}, whose cell size and orientations were set to $4\times4$ and 9 respectively. BHOG was constructed by concatenating the HOG feature  across all the bands. In this experiment, except for MHT in which the reliability weights were assigned in BACF, all other tracking steps were based on the framework of the original BACF.

Fig.~\ref{fig:hsiFeature} compares object tracking performance using different features. Spectral feature provides the worst accuracy among all the compared methods, as the raw spectrum is sensitive to illumination changes. Abundances exploit the underlying material distribution information, giving better performance. HOG and BHOG consider the local spatial structure information which is crucial for object tracking, therefore, produce more favorable results. SSHMG depicts local spectral-spatial structure information, yielding a gain of $0.4\%$ in AUC compared with HOG. Notably, replacing the SSHMG with the combination of abundances and SSHMG allows MHT to dominate SSHMG and HOG by a relatively large margin in average distance precision scores. This is owing to the hybrid benefit from SSHMG and abundances. From this experiment, we can infer that generic descriptors proposed in color images may not adapt well to robust material information representation of HSIs. In contrast, spectral-spatial material features are more effective in providing more discriminative information for target tracking.

\subsection{Quantitative Comparison with Hand-crafted Feature-based Trackers.}

In this experiment, we compare the proposed MHT tracker with ten state-of-the-art color trackers with hand-crafted features, including KCF~\cite{Henriques2015}, fDSST~\cite{Danelljan2017a}, SRDCF~\cite{Danelljan2015}, MUSTer~\cite{Hong2015}, SAMF~\cite{Li2014}, Struck~\cite{Hare2016}, CNT~\cite{Zhang2016}, BACF~\cite{Galoogahi2017}, CSR-DCF~\cite{Lukezic2017}, and MCCT~\cite{Wang2018}. The MHT tracker was tested on hyperspectral videos. Since all the alternative trackers were developed for color videos, they were run on both color videos and false-color videos generated from hyperspectral videos.
\begin{figure*}[!htp]
  \centering
\subfigure[Precision plot]{\includegraphics[width=0.3\linewidth, clip=true]{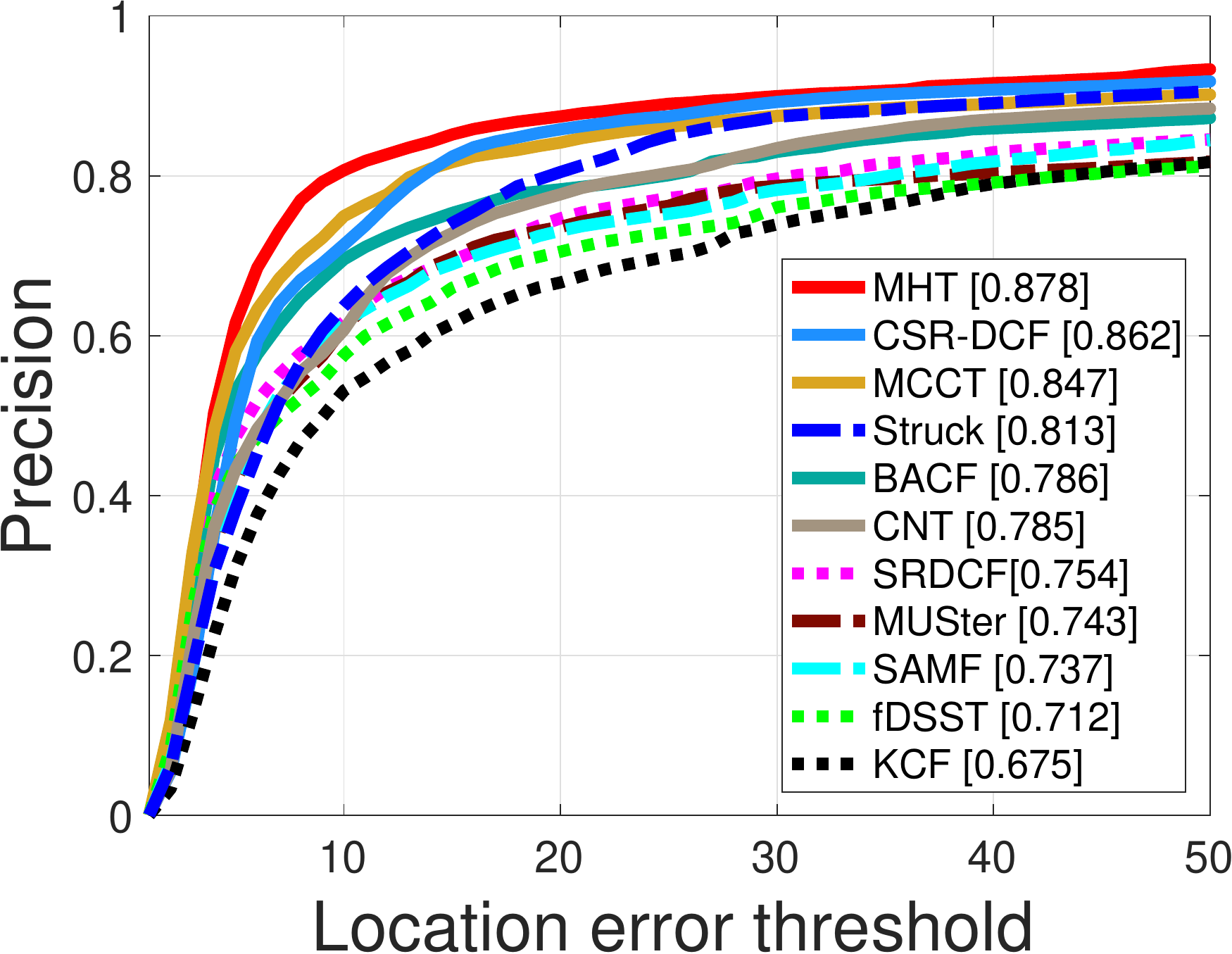}}
\subfigure[Success plot]{\includegraphics[width=0.3\linewidth, clip=true]{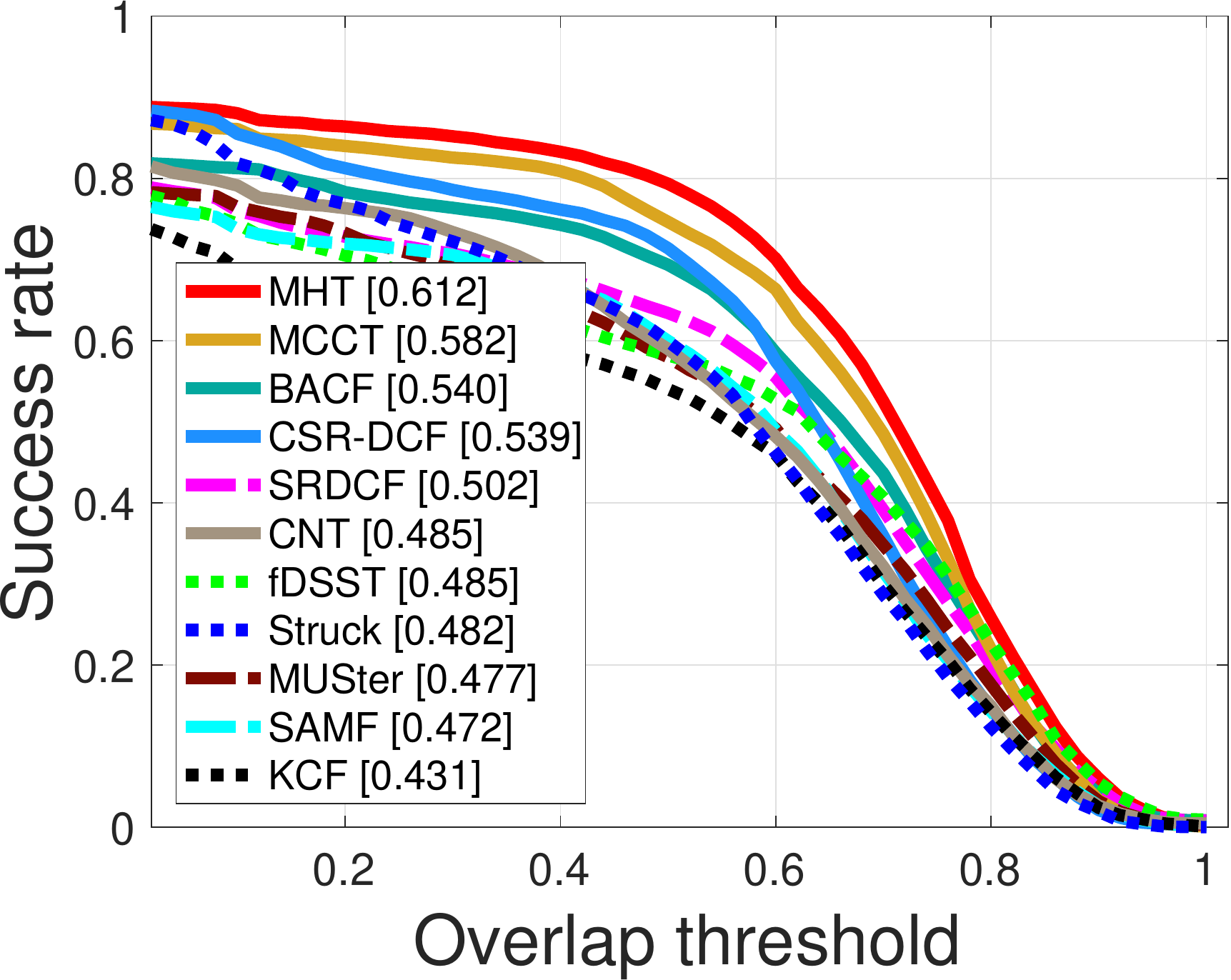}}
\caption{Comparisons with hand-crafted feature-based trackers on RGB videos. MHT outperforms all the other trackers. } \label{fig:colorvideo}
 \end{figure*}

\begin{figure*}[!htp]
\centering
\subfigure[Precision plot]{\includegraphics[width=0.3\linewidth, clip=true]{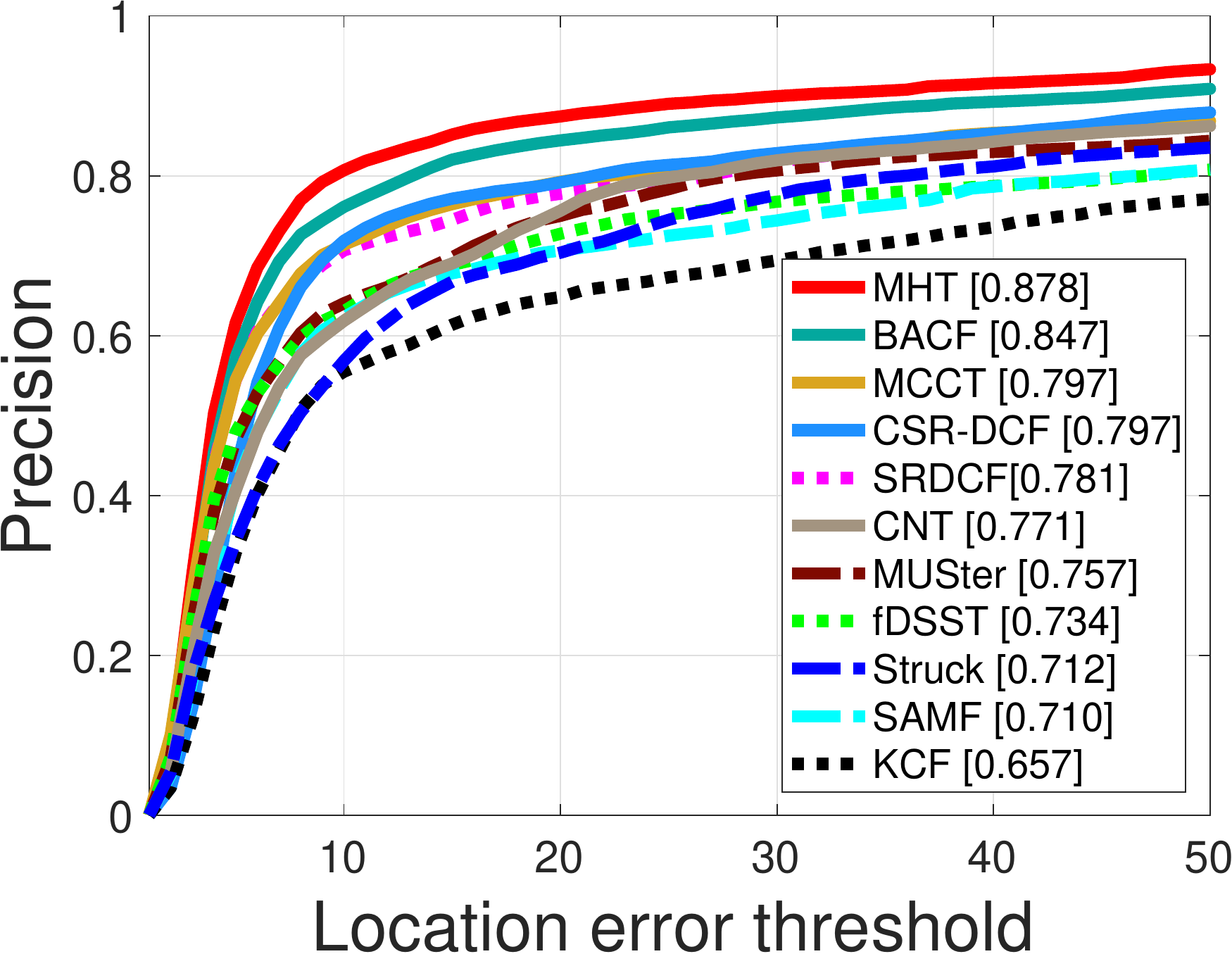}}
\subfigure[Success plot]{\includegraphics[width=0.3\linewidth, clip=true]{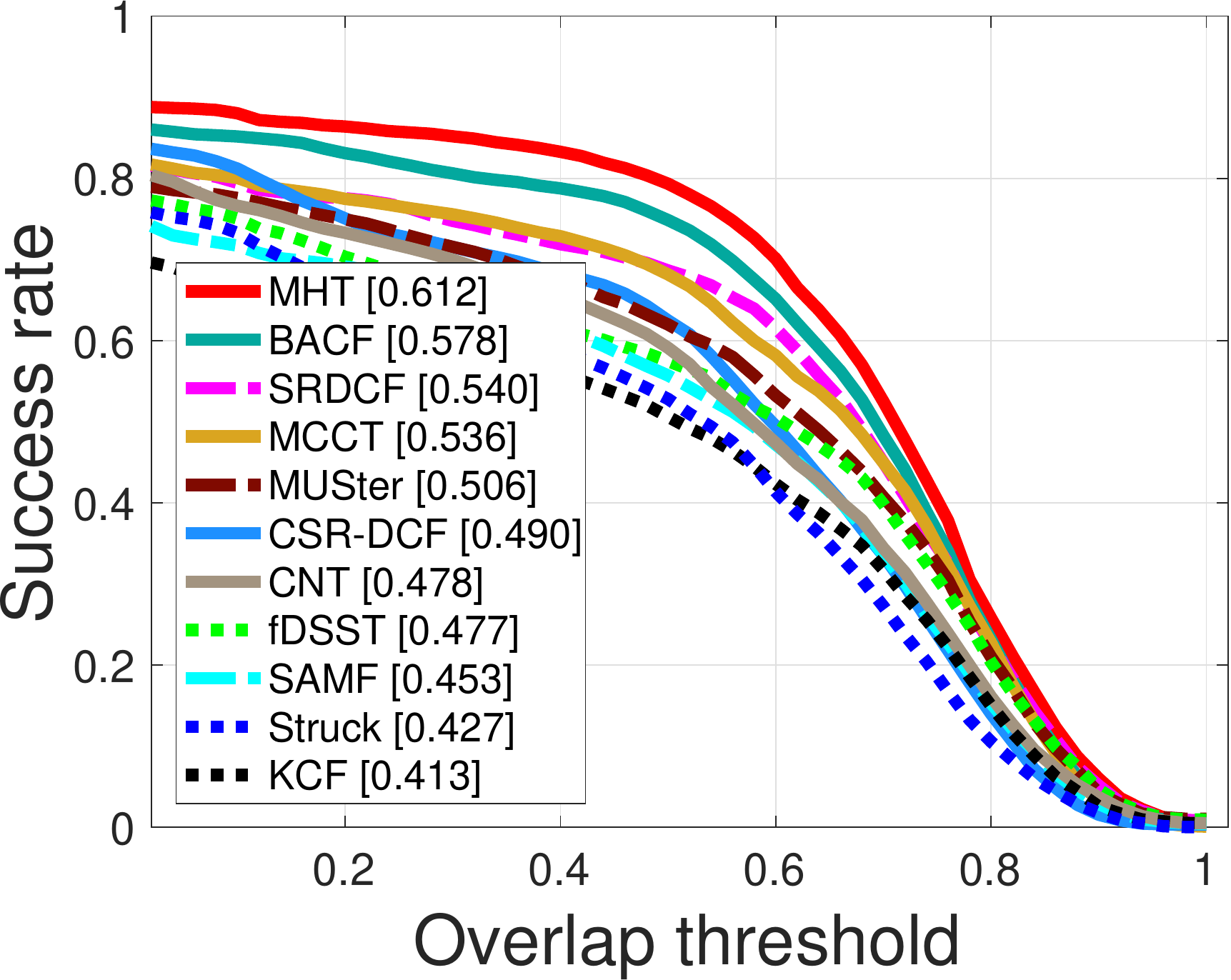}}
\caption{Comparisons with hand-crafted feature-based trackers on hyperspectral videos or corresponding false-color videos. MHT achieves the best accuracy with an AUC of $61.2\%$. } \label{fig:hsivideo}
\end{figure*}
Fig.~\ref{fig:colorvideo} and Fig.~\ref{fig:hsivideo} show the tracking performance of all methods. The results show that KCF and Struck give unsatisfying success scores because of limited consideration of scale estimation. MUSter also produces inferior performance due to the fact that it fails to detect the key points when the object shares similar appearance as the background. CSR-DCF, BACF and SRDCF integrate the background information to learn more discriminative filters, resulting in much better tracking performance. It is worth mentioning that the proposed MHT tracker achieves noticeable improvement over the original BACF, when original BACF was run on both color videos and hyperspectral videos respectively, providing a gain of $7.2\%$ and $3.4\%$ in AUC. In addition, compared with the other trackers, our approach ranks top over a range of thresholds, by achieving an AUC of $61.2\%$, followed by MCCT on color dataset. This implies that MHT well represents the image content using detailed constitute material distribution information and local spectral-spatial information contained in an HSI. Such information is beneficial to enabling a tracker to distinguish the target from the background. Furthermore, due to adaptive feature robustness learning, our tracker is able to make full use of more reliable features to learn correlation filters.

\subsection{Quantitative Comparison with Deep Feature-based Trackers}

\begin{table*}[tbp]
\centering
\caption{Performance comparison with deep trackers in terms of AUC. The top two values are highlighted by \textcolor{red}{red} and \textcolor{blue}{blue}.}\label{tab:deep}
\tablesize{
\begin{tabular}{cccccccccccc}
\hline
\hline
Video&MHT&C-COT~\cite{Danelljan2016}&ECO~\cite{Danelljan2017}&CF2~\cite{Ma2015}&CFNet~\cite{Valmadre2017a}&HDT~\cite{Qi2016}&DSiam~\cite{Guo2017a}&DeepSRDCF~\cite{Danelljan2015a}&TRACA~\cite{Choi2018}\\
Color&n/a&\textbf{\textcolor{red}{0.617}}&0.575&0.483&\textbf{\textcolor{blue}{0.580}}&0.453&0.564&0.571&0.570\\
Hyperspectral/False-color&\textbf{\textcolor{red}{0.612}}&0.561&\textbf{\textcolor{blue}{0.586}}&0.486&0.522&0.480&0.502&0.566&0.565\\
\hline
\hline
\end{tabular}}
\end{table*}
In this section, we select several state-of-the-art deep feature-based trackers for comparison, including ECO~\cite{Danelljan2017}, CF2~\cite{Ma2015}, TRACA~\cite{Choi2018}, CFNet~\cite{Valmadre2017a}, HDT~\cite{Qi2016}, DSiam~\cite{Guo2017a}, DeepSRDCF~\cite{Danelljan2015a}, and C-COT~\cite{Danelljan2016}. As reported in Table~\ref{tab:deep}, the proposed method achieves competitive performance with C-COT on color videos and significantly outperforms the other trackers on both color and false-color videos. The reason of lower performance of other methods is the challenging nature of the dataset which  contains background cluster, rotation and deformation, etc. In addition, most of deep trackers present lower AUCs in hyperspectral dataset. This means useful spectral information contained in hyperspectral videos is lost in  the false-color videos.

 \begin{figure*}[!htp]
  \centering
\subfigure[Precision plot]{\includegraphics[width=0.3\linewidth, clip=true]{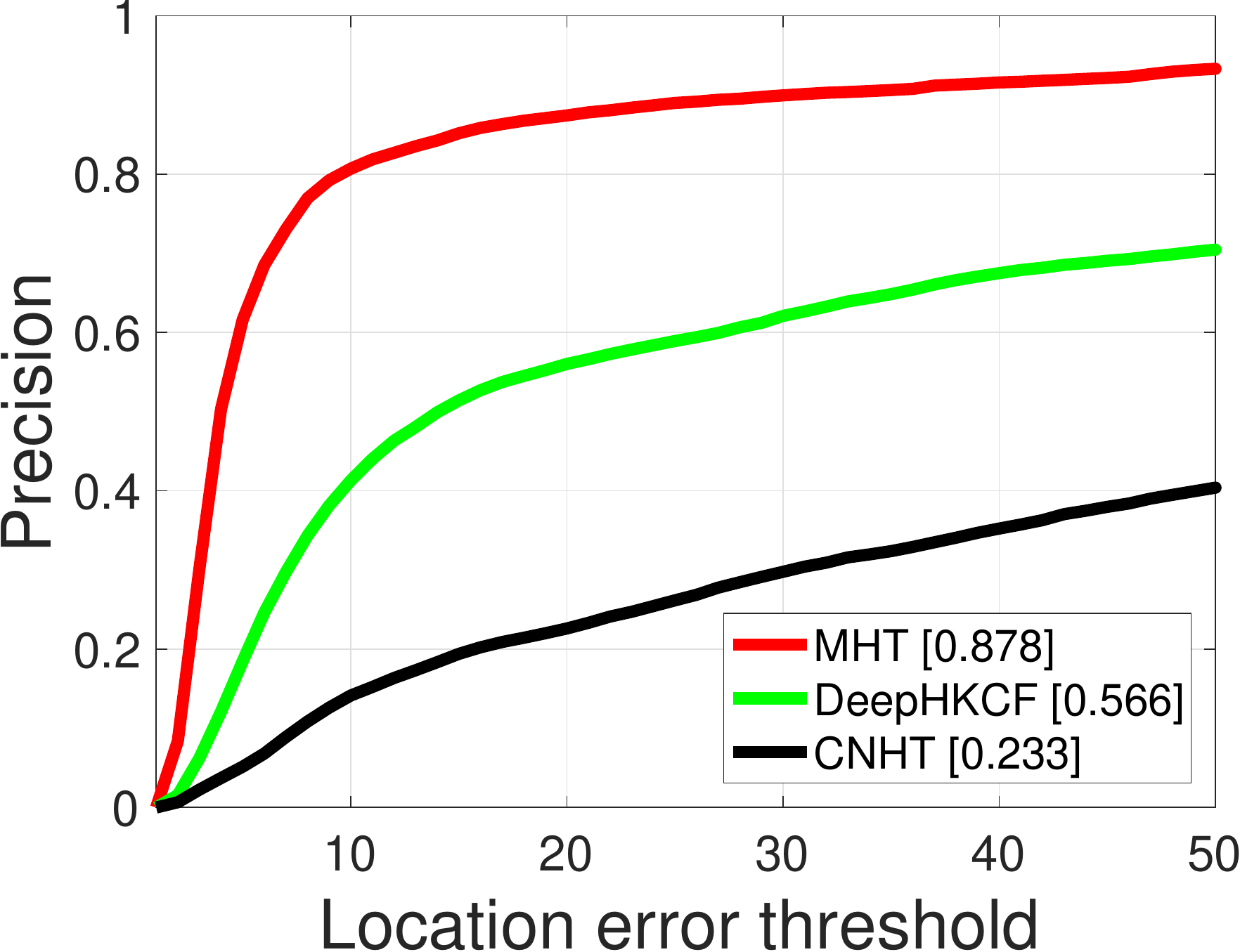}}
\subfigure[Success plot]{\includegraphics[width=0.3\linewidth, clip=true]{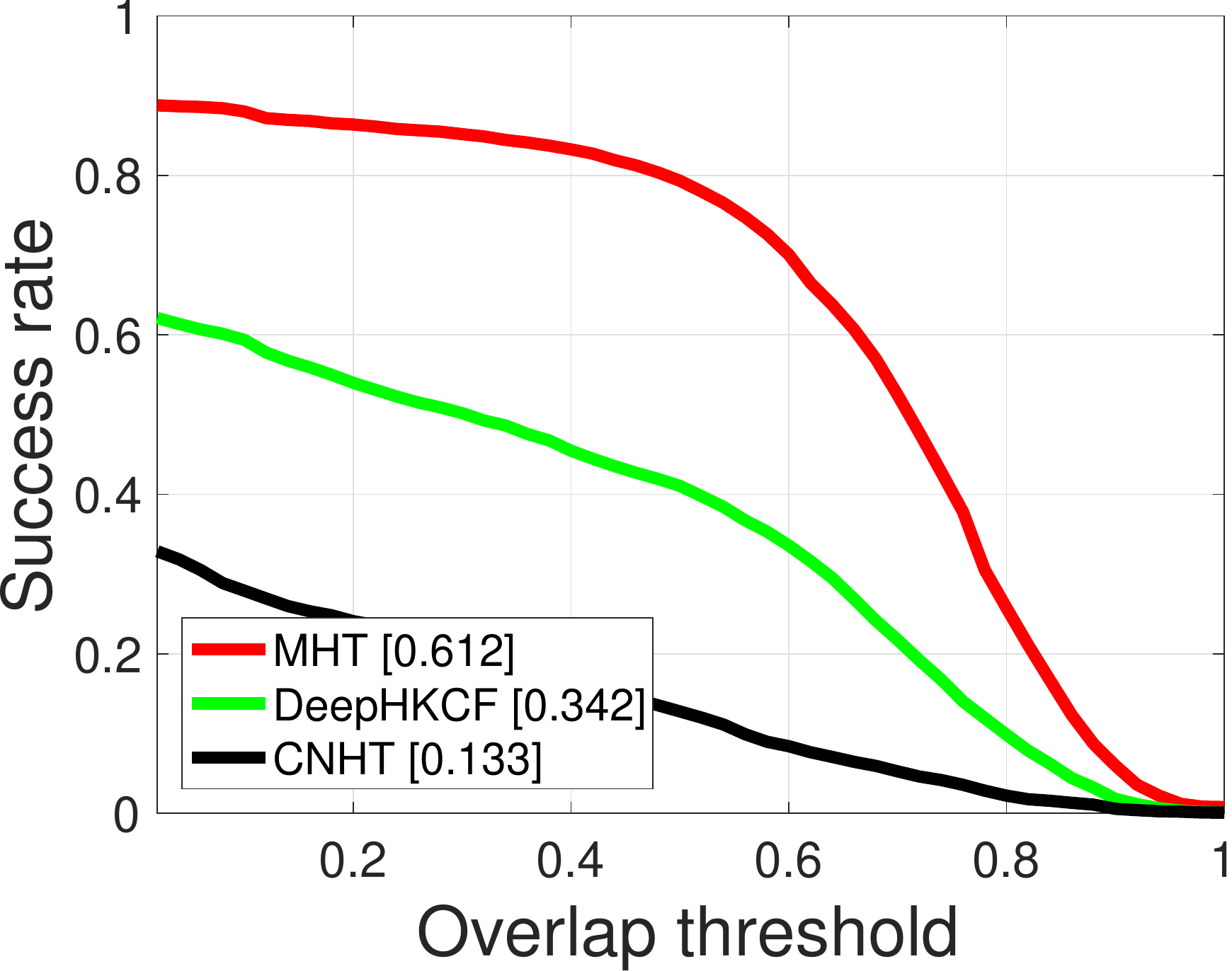}}
\caption{Comparison with hyperspectral trackers.} \label{fig:otherhsi}
 \end{figure*}

\begin{table*}[htbp]
\centering
\caption{Attribute-based comparison on hyperspectral/color videos. The best two results are shown in \textcolor{red}{red} and \textcolor{blue}{blue} fonts. Our tracker ranks the first on 5 out of 11 attributes: IPR, OPR, DEF, BC, and OV.}\label{tab:auc}
\centering
\tablesize{
\begin{tabular}{lcccccccccc}
\hline
\hline
Attributes &MHT&C-COT~\cite{Danelljan2016}&MCCT~\cite{Wang2018}&CFNet~\cite{Valmadre2017a}&ECO~\cite{Danelljan2017}&DeepSRDCF~\cite{Danelljan2015a}&TRACA~\cite{Choi2018}&DSiam~\cite{Guo2017a}&BACF~\cite{Galoogahi2017}&CSR-DCF~\cite{Lukezic2017}\\
SV&\textcolor{blue}{\textbf{0.609}}&\textcolor{red}{\textbf{0.613}}&0.563&0.591&0.577&0.531&0.526&0.561&0.525&0.518\\
MB&0.619&\textcolor{red}{\textbf{0.683}}&0.592&0.603&\textcolor{blue}{\textbf{0.626}}&0.614&0.584&0.605&0.526&0.615\\
OCC&0.508&\textcolor{red}{\textbf{0.564}}&0.468&0.489&0.523&\textcolor{blue}{\textbf{0.528}}&0.460&0.502&0.426&0.394\\
FM&0.579&\textcolor{blue}{\textbf{0.662}}&0.604&0.605&0.558&\textcolor{red}{\textbf{0.679}}&0.535&0.565&0.541&0.390\\
LR&0.512&\textcolor{red}{\textbf{0.566}}&\textcolor{blue}{\textbf{0.545}}&0.436&0.450&0.518&0.338&0.521&0.329&0.469\\
IPR&\textcolor{red}{\textbf{0.705}}&0.677&\textcolor{blue}{\textbf{0.687}}&0.664&0.639&0.653&0.650&0.656&0.658&0.641\\
OPR&\textcolor{red}{\textbf{0.707}}&\textcolor{blue}{\textbf{0.677}}&0.657&0.671&0.648&0.627&0.654&0.654&0.635&0.646\\
DEF&\textcolor{red}{\textbf{0.687}}&0.619&0.627&\textcolor{blue}{\textbf{0.678}}&0.604&0.558&0.668&0.643&0.577&0.655\\
BC&\textcolor{red}{\textbf{0.653}}&0.602&0.595&0.590&0.590&0.519&0.588&0.603&0.524&\textcolor{blue}{\textbf{0.604}}\\
IV
&0.490&\textcolor{red}{\textbf{0.558}}&0.441&0.497&0.549&\textcolor{blue}{\textbf{0.550}}&0.457&0.357&0.458&0.309
\\
OV&\textcolor{red}{\textbf{0.768}}&0.722&0.519&0.718&0.736&0.727&\textcolor{blue}{\textbf{0.740}}&0.666&0.666&0.726\\
\hline
\hline
\end{tabular}}
\end{table*}
 \begin{figure*}[!tp]
 \centering
 \includegraphics[width=0.2\linewidth, height=0.13\linewidth, clip=true]{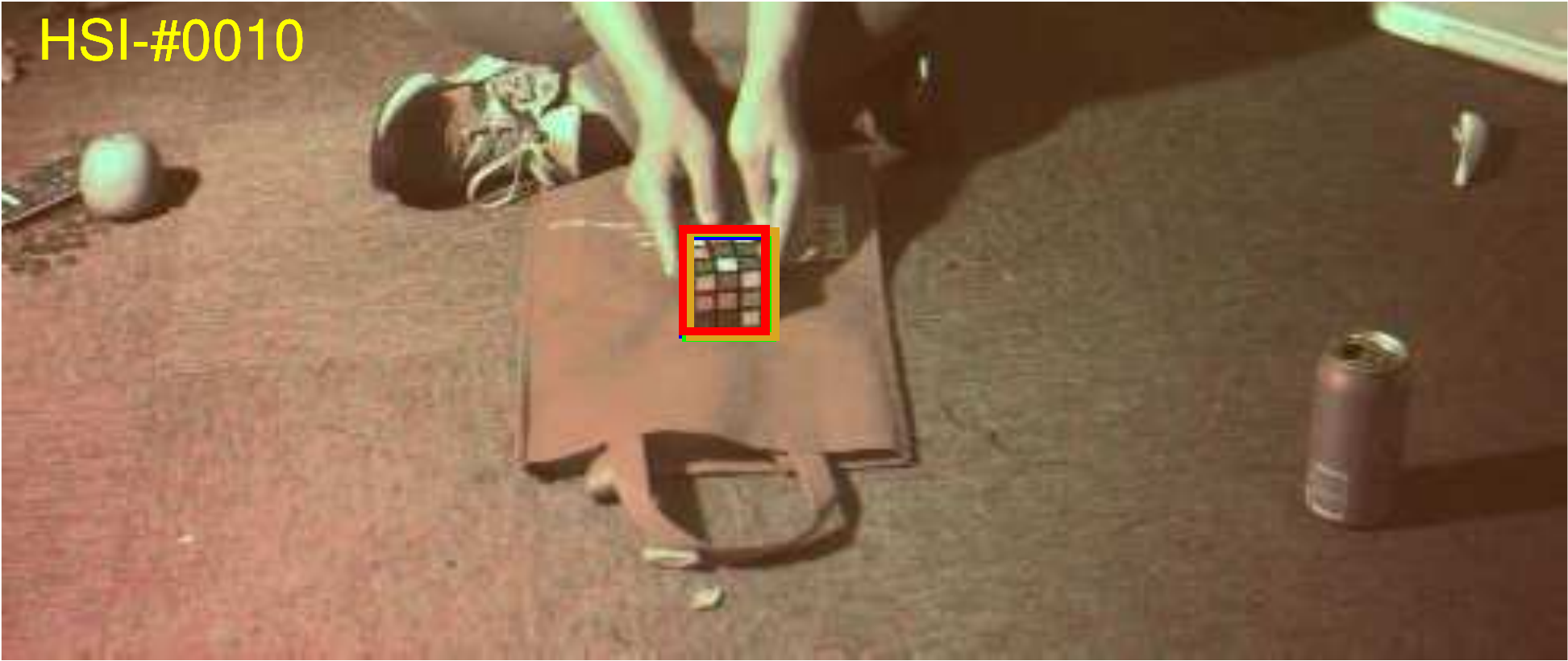}
\includegraphics[width=0.2\linewidth, height=0.13\linewidth, clip=true]{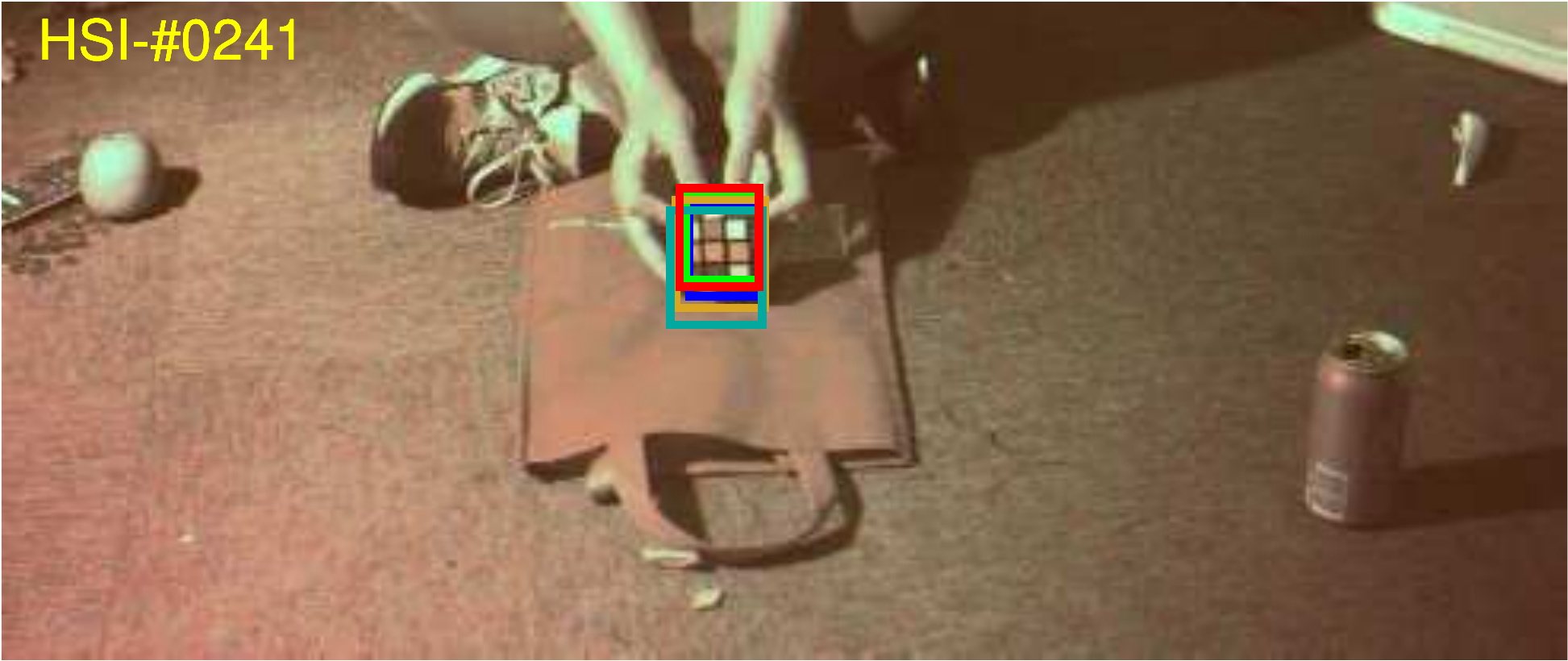}
\includegraphics[width=0.2\linewidth, height=0.13\linewidth, clip=true]{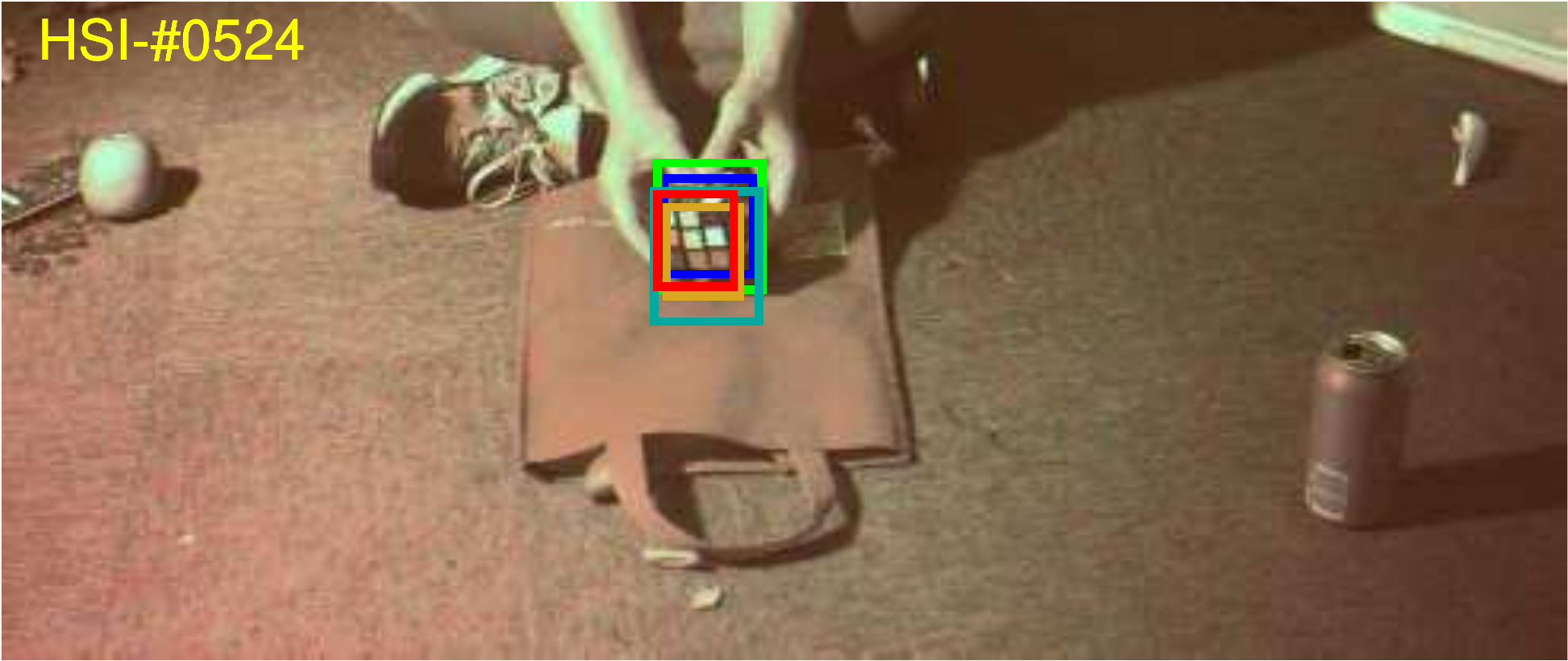}
\\
\includegraphics[width=0.2\linewidth, height=0.13\linewidth, clip=true]{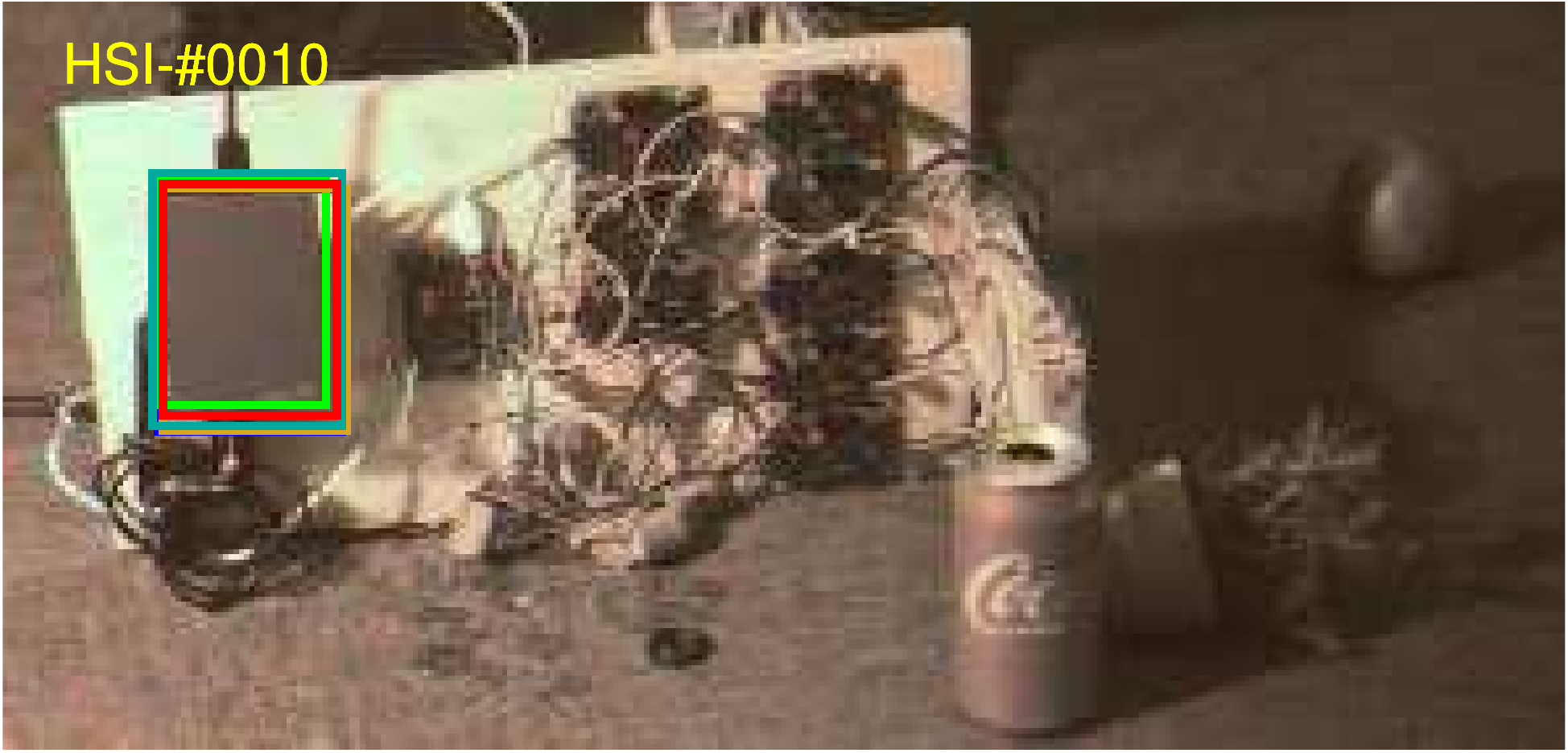}
\includegraphics[width=0.2\linewidth, height=0.13\linewidth, clip=true]{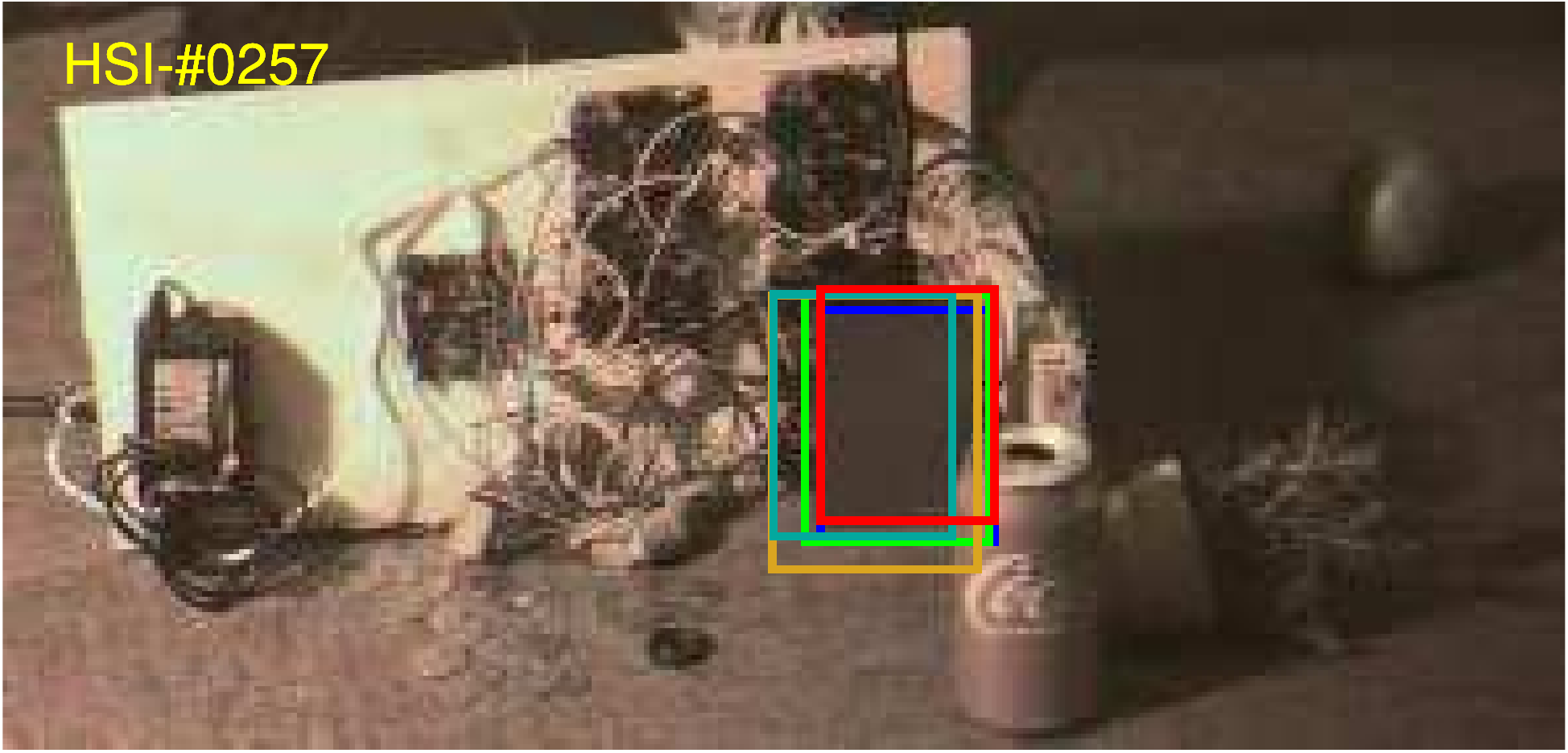}
\includegraphics[width=0.2\linewidth, height=0.13\linewidth, clip=true]{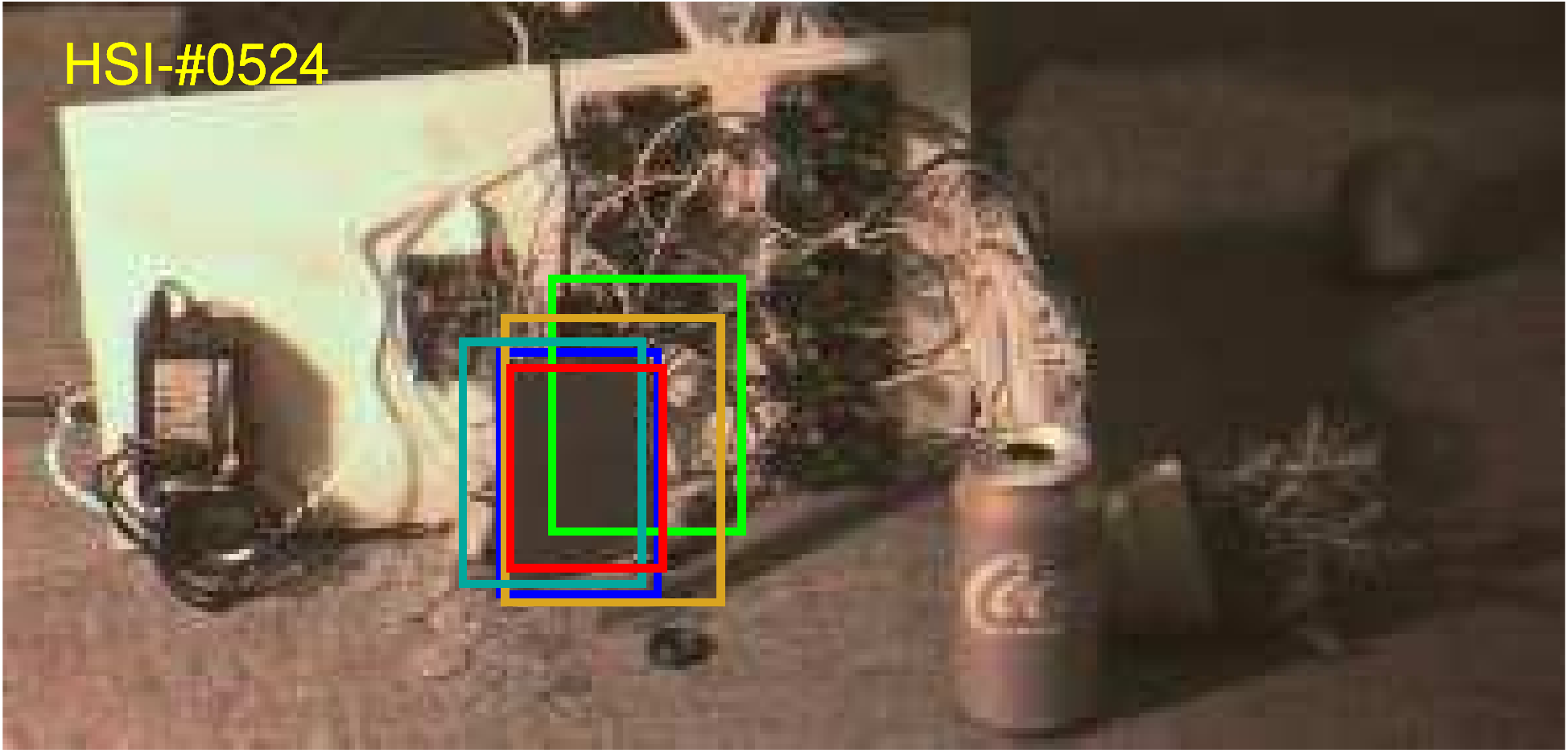}\\
\includegraphics[width=0.2\linewidth, height=0.13\linewidth, clip=true]{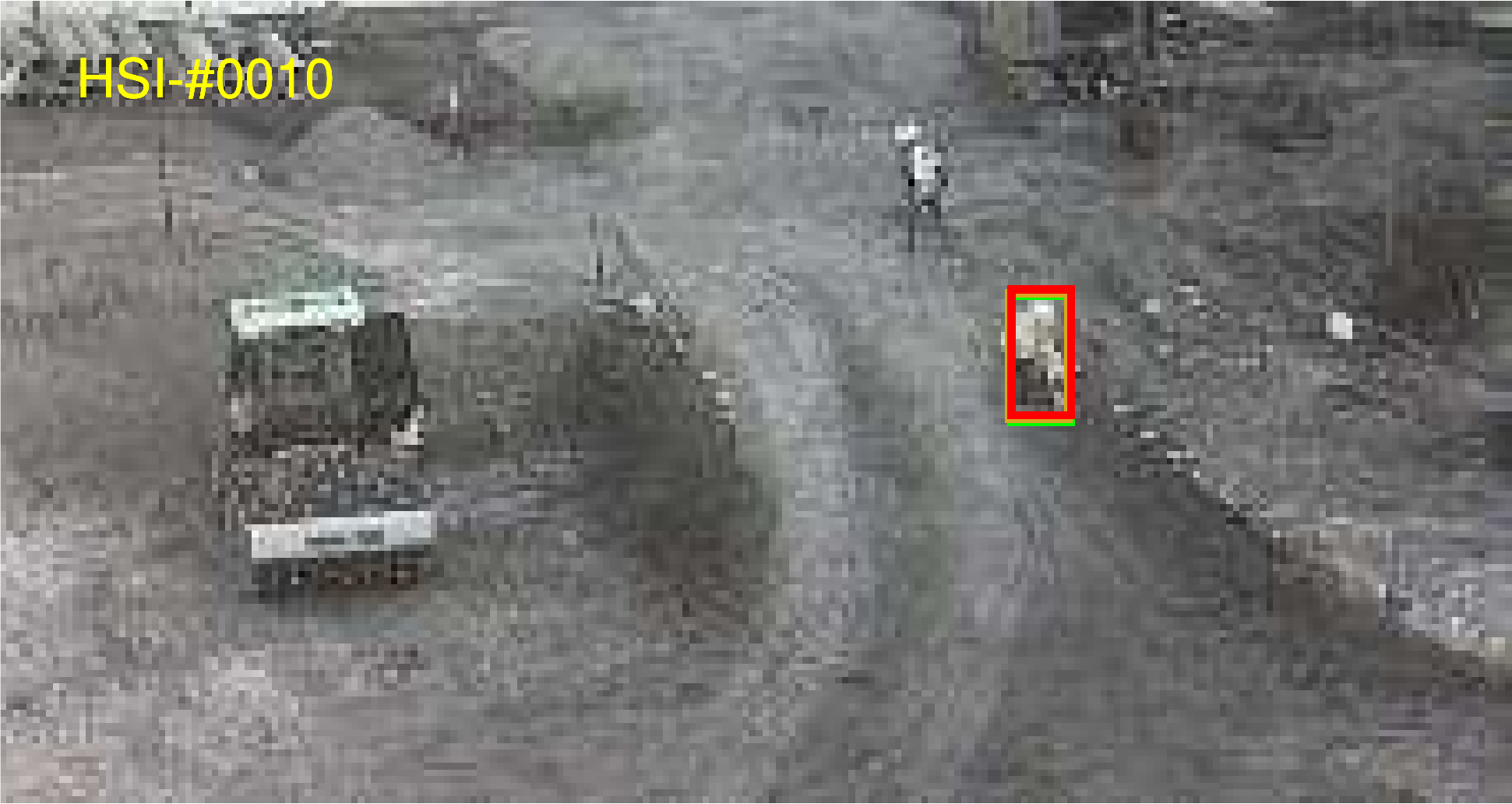}
\includegraphics[width=0.2\linewidth, height=0.13\linewidth, clip=true]{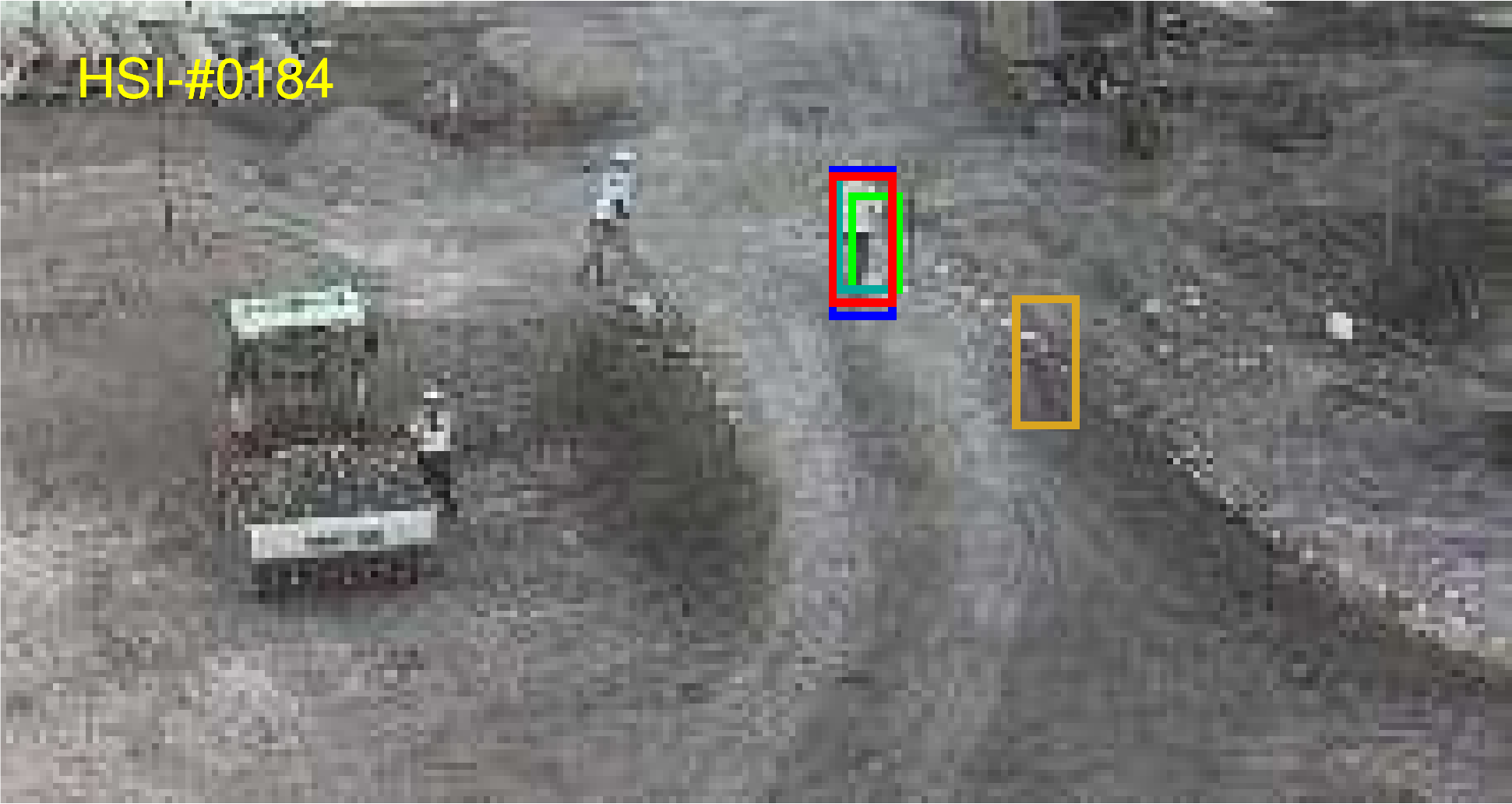}
\includegraphics[width=0.2\linewidth, height=0.13\linewidth, clip=true]{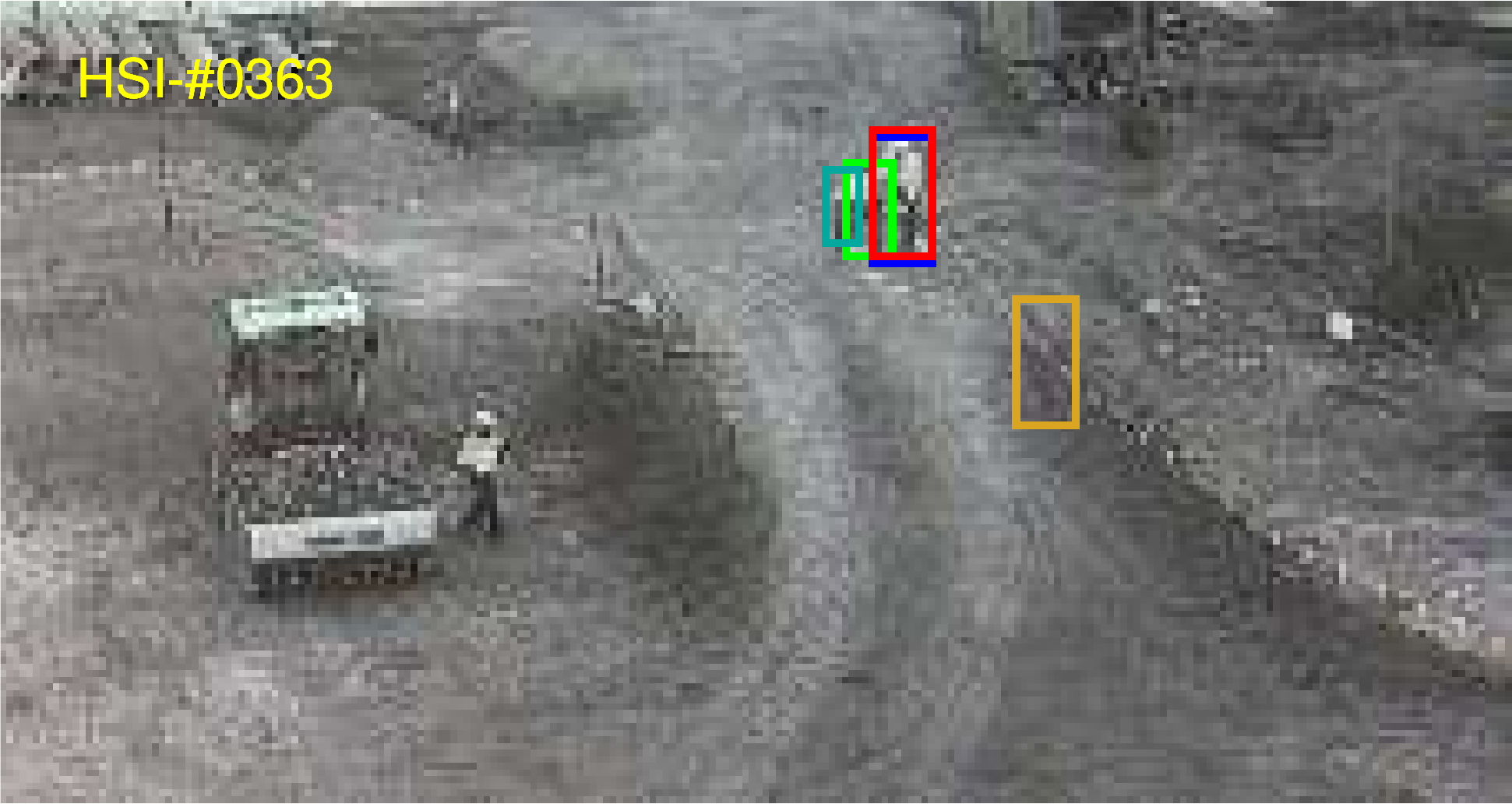}
\\
\includegraphics[width=0.2\linewidth, height=0.13\linewidth, clip=true]{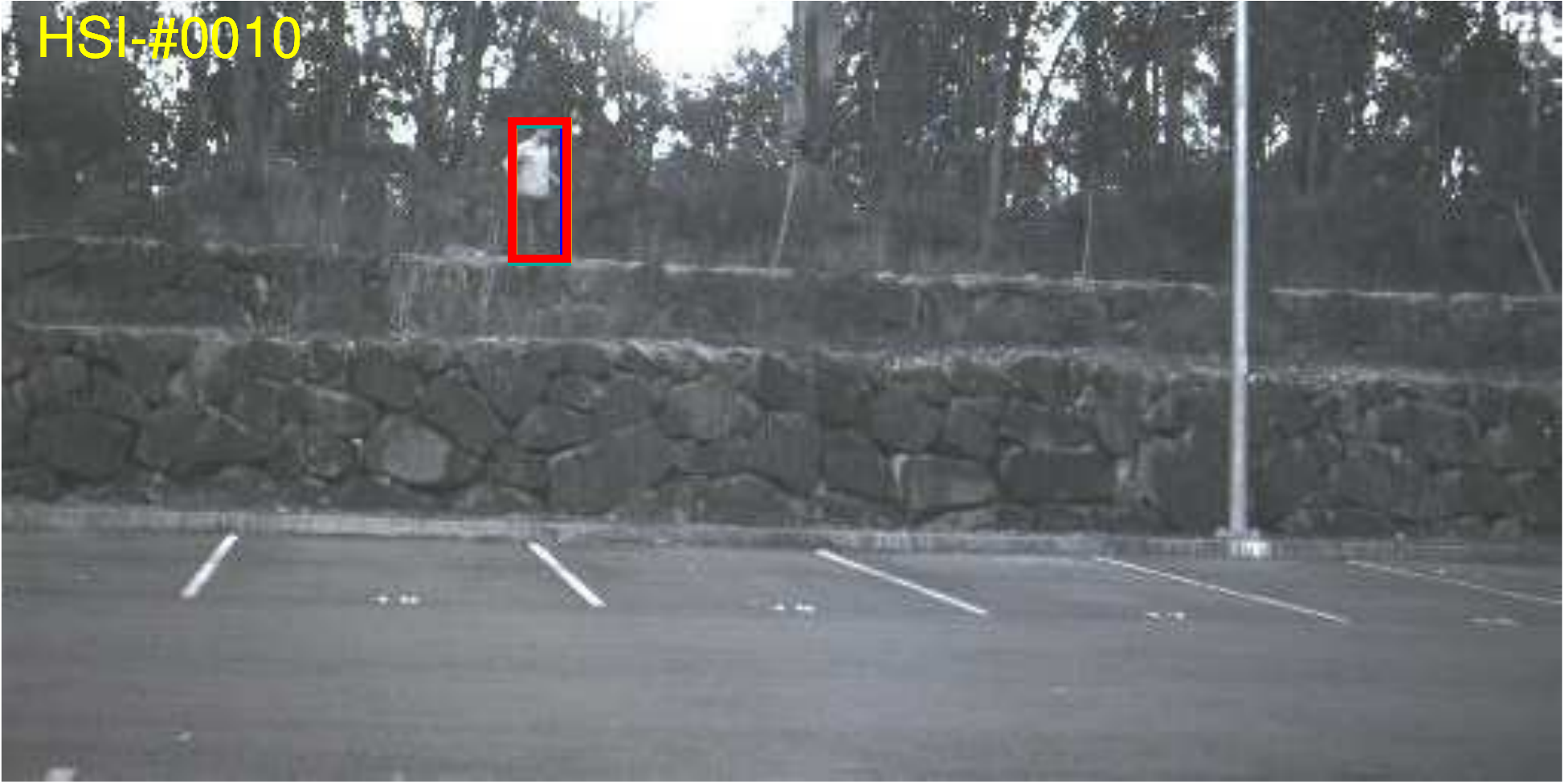}
\includegraphics[width=0.2\linewidth, height=0.13\linewidth, clip=true]{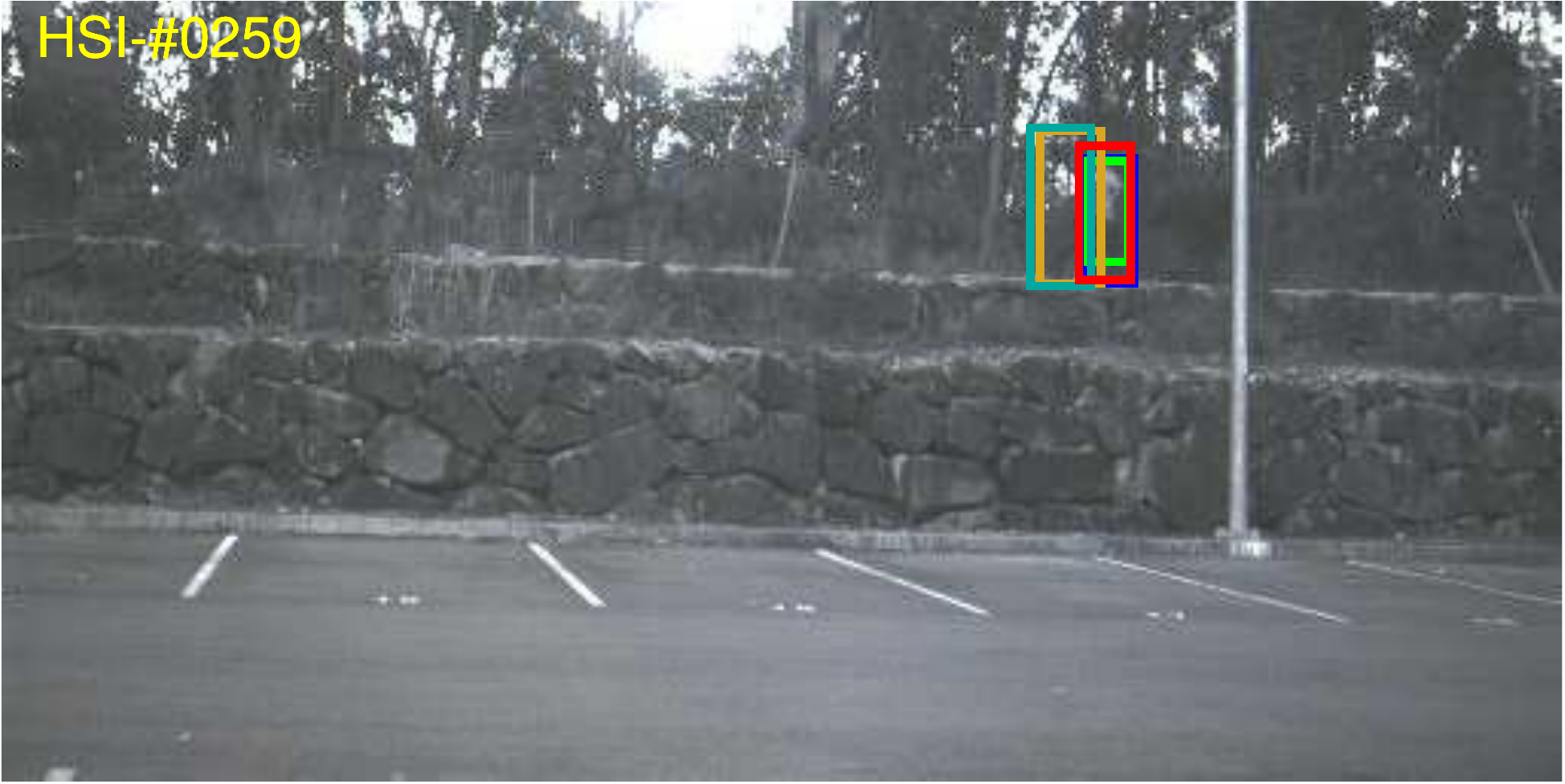}
\includegraphics[width=0.2\linewidth, height=0.13\linewidth, clip=true]{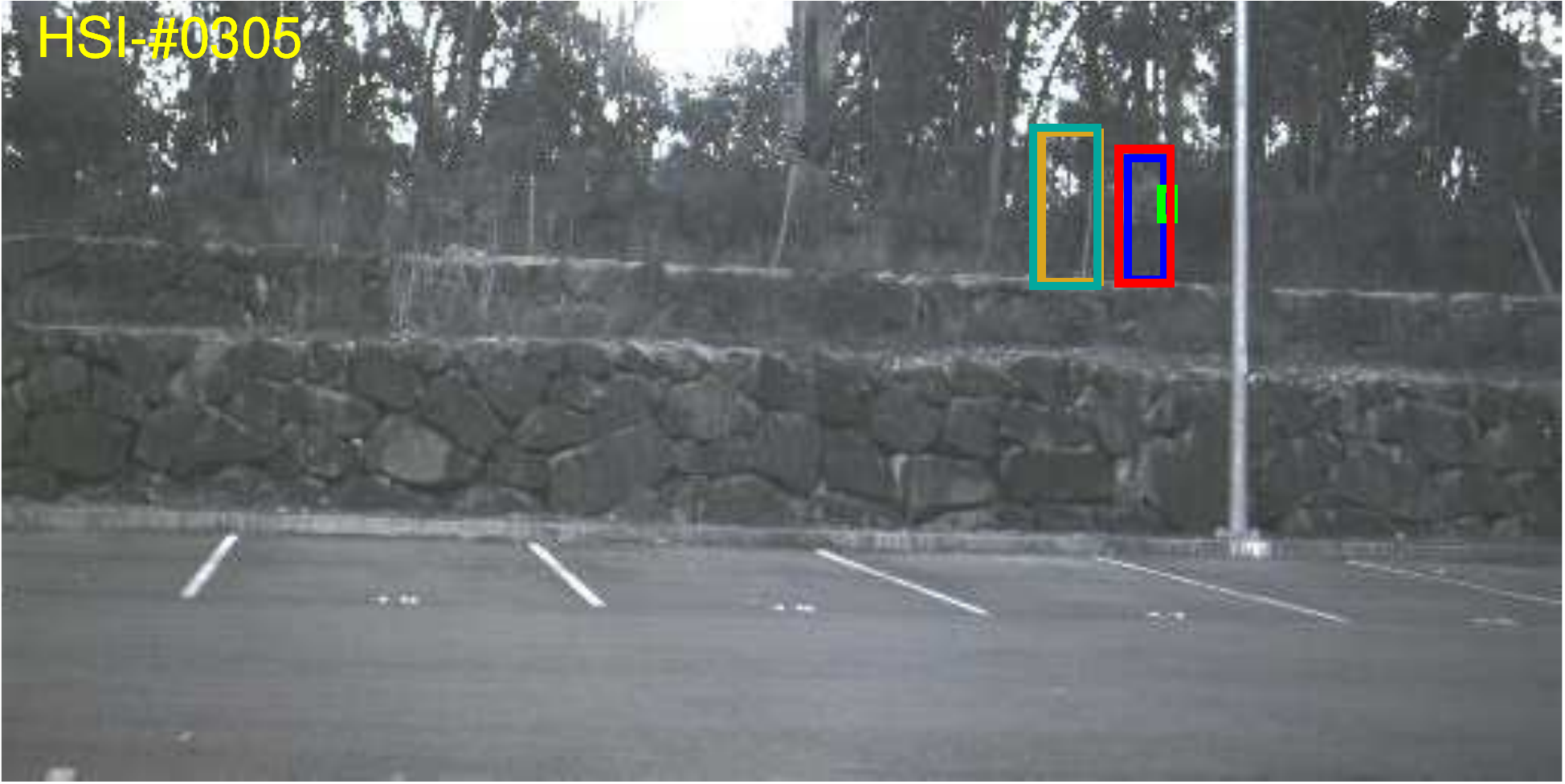}
\\
\includegraphics[width=0.4\linewidth, clip=true]{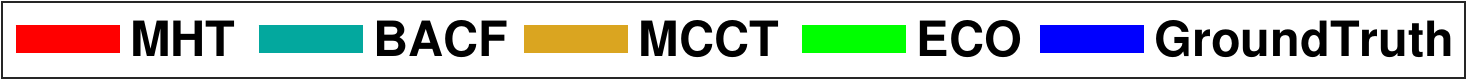}
\caption{Qualitative evaluation on 4 video sequences (i.e., \emph{rubik, drive, worker, forest}). } \label{fig:hsiQualitive}
 \end{figure*}

\subsection{Quantitative Comparison with Hyperspectral Trackers.}
We also compared our method with two recent hyperspectral trackers, CNHT~\cite{Qian2018} and DeepHKCF~\cite{Uzkent2018}. Both CNHT and DeepHKCF are based on KCF but use different features. In CNHT, normalized three-dimensional patches were selected from the target region in the initial frame as fixed convolution kernels for feature extraction in succeeding frames. In terms of DeepHKCF, an HSI was converted into false-color image to learn deep features by VGGNet.

Fig.~\ref{fig:otherhsi} presents the tracking results of all competing hyperspectral trackers. The results show that CNHT gives inferior accuracy due to the fact that it only considers fixed positive samples in learning convolutional filters. Without negative patches in the surrounding background, the features produced by the fixed convolutional filters  are not discriminative enough to learn a robust model, significantly deteriorating the prediction of object location. In contrast, VGGNet uses both positive and negative samples to learn discriminative feature representation. Therefore, DeepHKCF shows more competitive performance than CNHT. However, since HSIs are converted into three-channel false-color images before passing through the VGGNet, the complete spectral-spatial structural information in an HSI is not fully explored. This makes DeepHKCF fail to outperform the proposed MHT tracker. Combining local spectral-spatial texture information and detailed material information, the proposed MHT achieves the most appealing performance. This experiment again suggests that material information facilitates object tracking.

\subsection{Attribute-based Evaluation}
In this experiment, we report the tracking effectiveness with respect to different video attributes. For simplicity, we only present the performance of top 9 color trackers on color videos and MHT on hyperspectral videos. Table~\ref{tab:auc} reports the AUCs of all the trackers. MHT ranks the first on 5 out of 11 attributes: IPR, OPR, SV, DEF and BC. In BC situations, it is difficult to extract robust features from the target in a color image. In contrast, the material information can be captured by an HSI and represented by the proposed SSHMG and abundance features, helping our tracker to discriminate the object from the background. On the videos with DEF, OPR and IPR attributes, the tracking target is partly or fully deteriorated, which makes the spatial structure information unreliable. Compared with spatial information, the underlying material information is more robust. Thanks to the superior advantages of the proposed SSHMG and abundances in spectral-spatial material structure representation and underlying material information exploitation, MHT tracker is more capable of separating  target from surrounding environment. It is also observed that MHT fails to outperform color trackers in IV. The main reason is that the endmembers are fixed during tracking, which are in fact related to illumination condition.
\begin{table*}[!tp]
\centering
\caption{Sequences in our benchmark. } \label{tab:dataset}
\begin{tabular}{ccccc}
\includegraphics[width=0.16\linewidth, height=0.12\linewidth, clip=true]{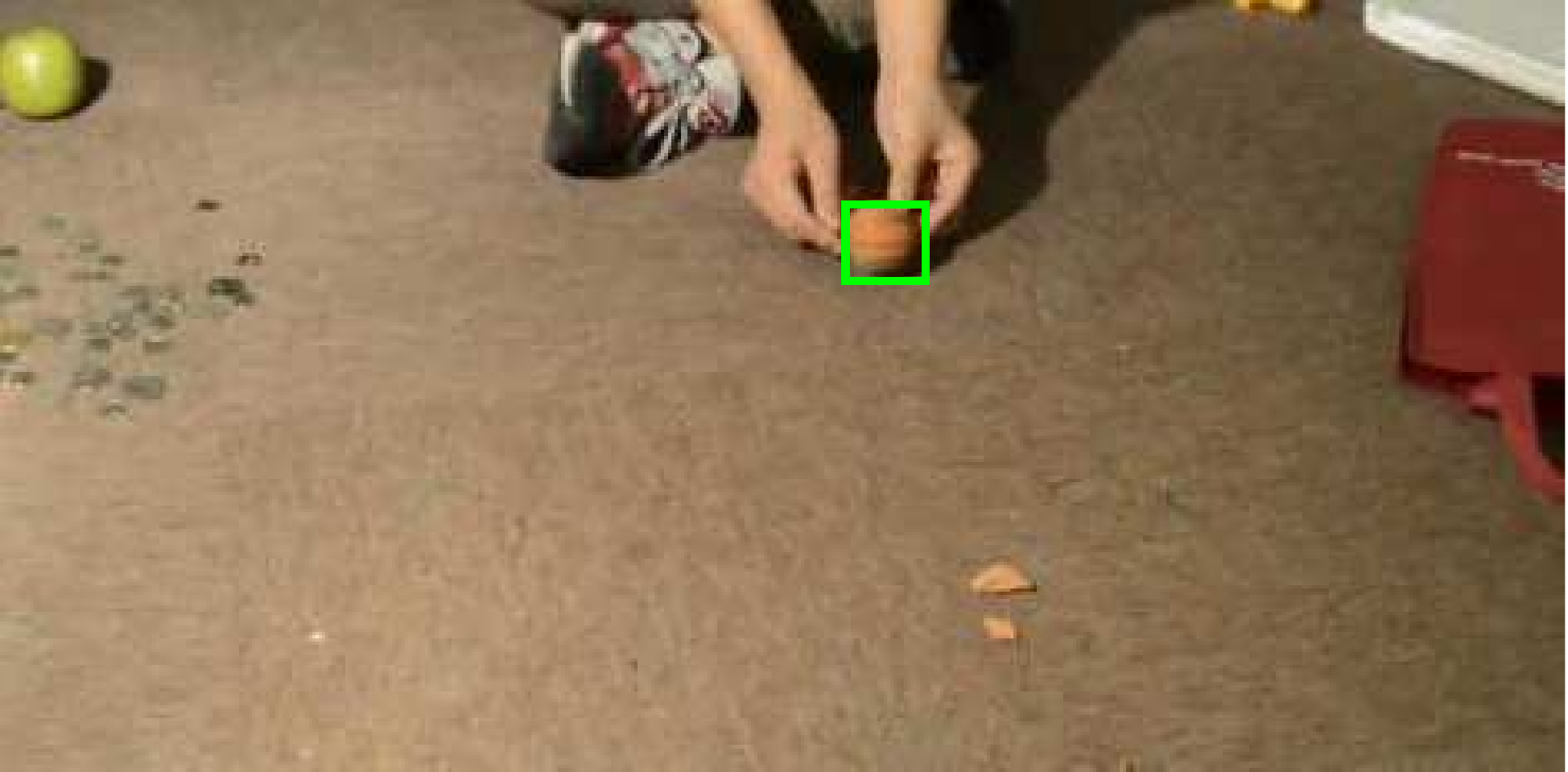}
 &\includegraphics[width=0.16\linewidth, height=0.12\linewidth, clip=true]{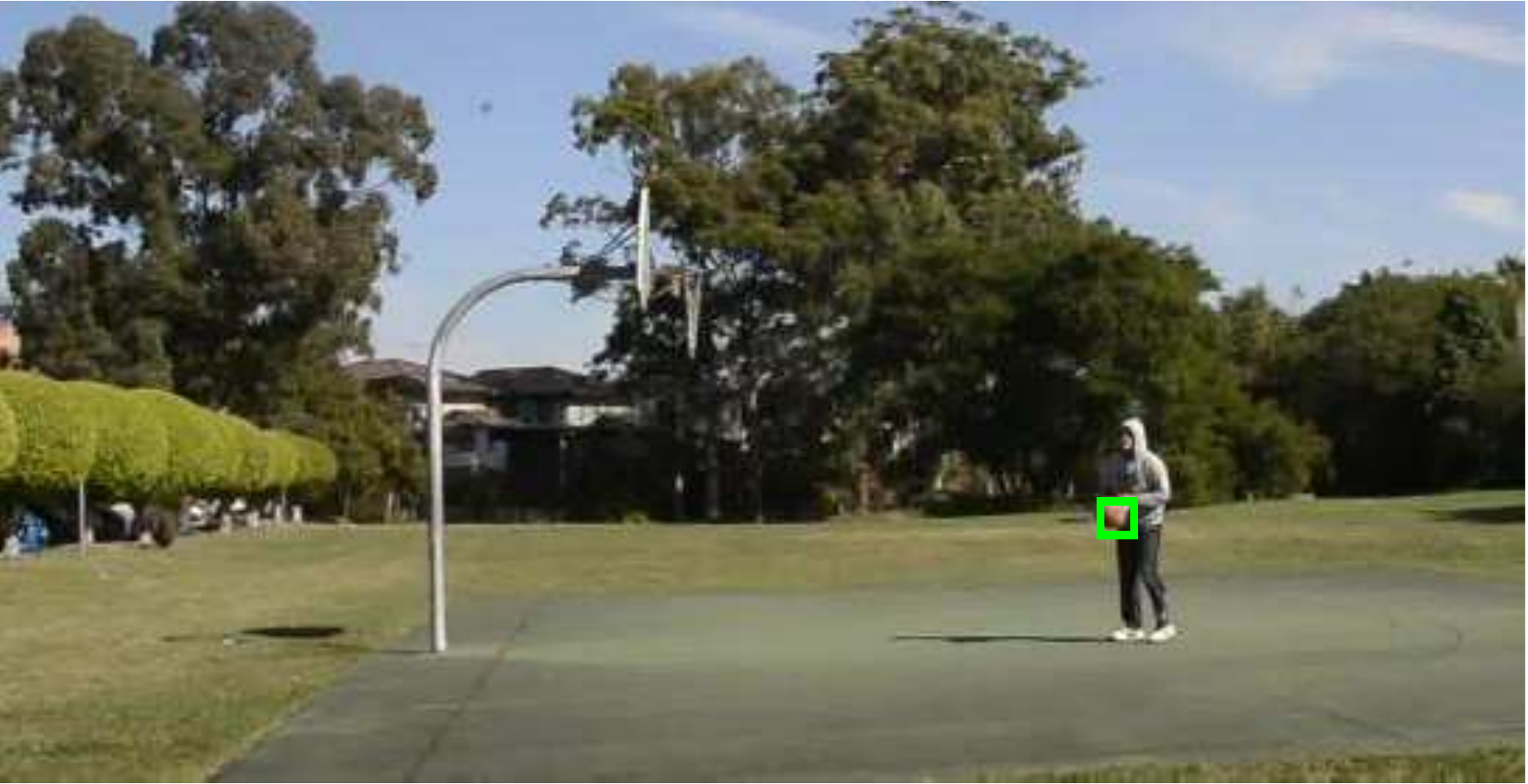}
 &\includegraphics[width=0.16\linewidth, height=0.12\linewidth, clip=true]{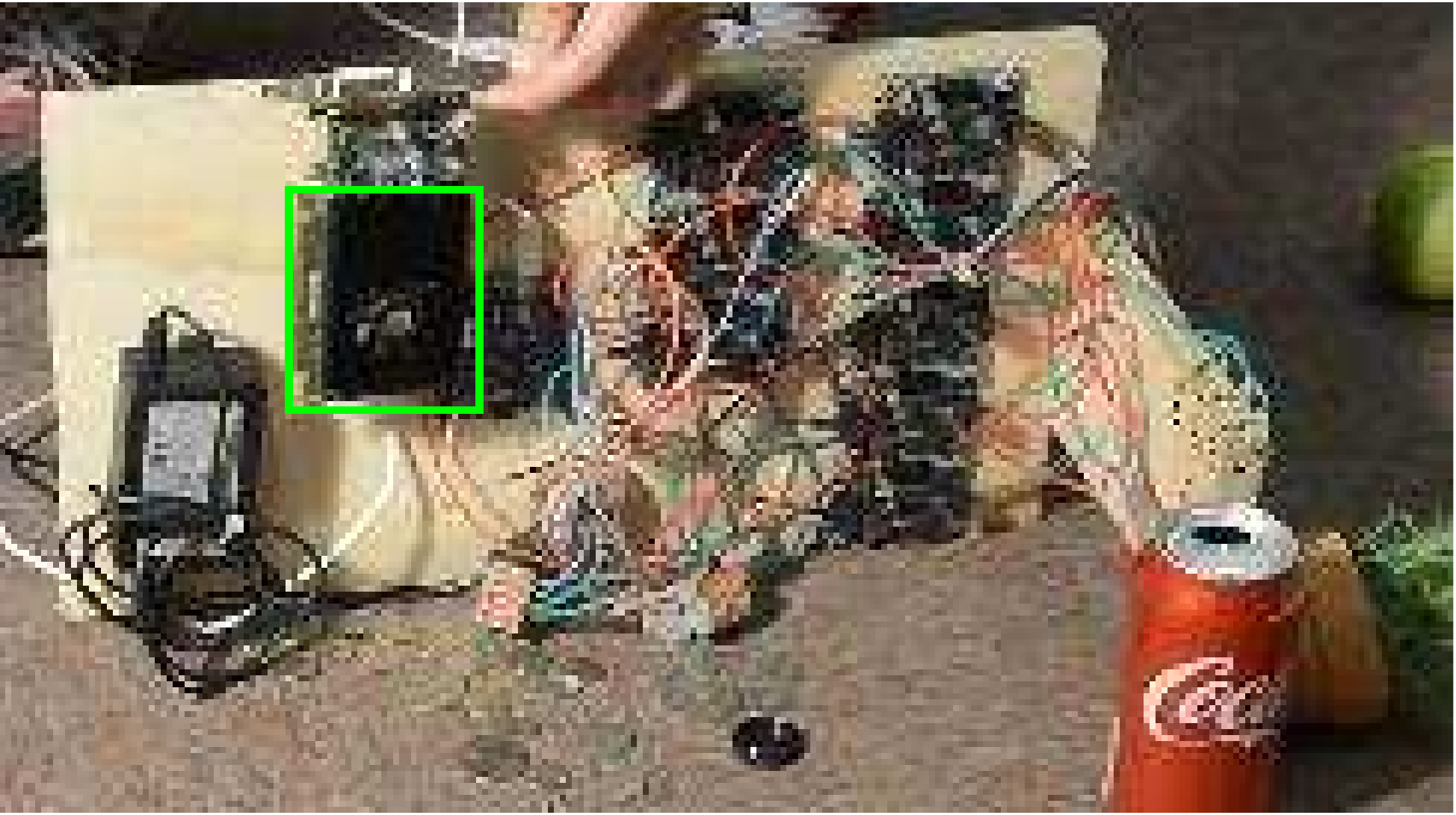}
  &\includegraphics[width=0.16\linewidth, height=0.12\linewidth, clip=true]{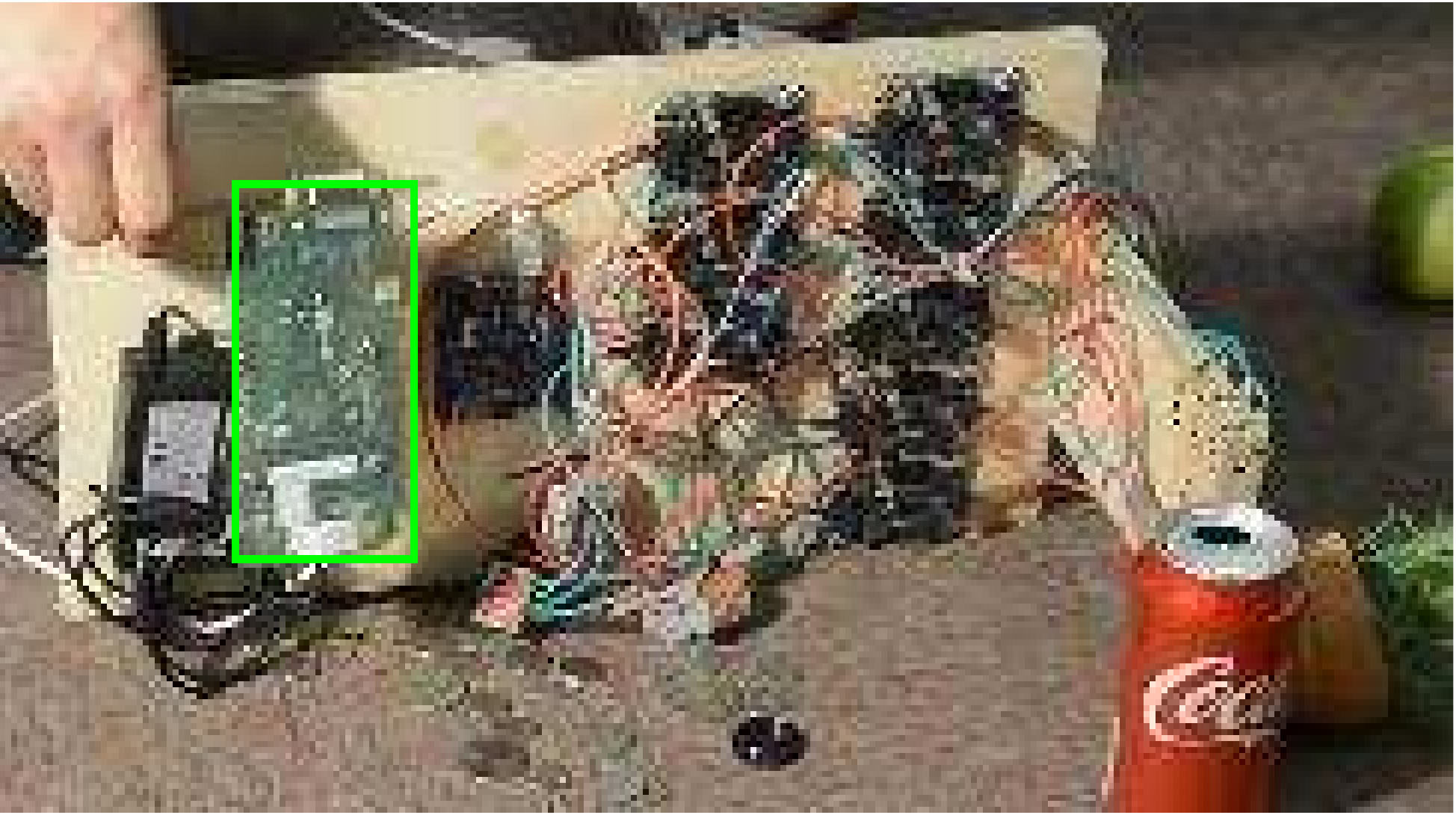}
	&\includegraphics[width=0.16\linewidth, height=0.12\linewidth, clip=true]{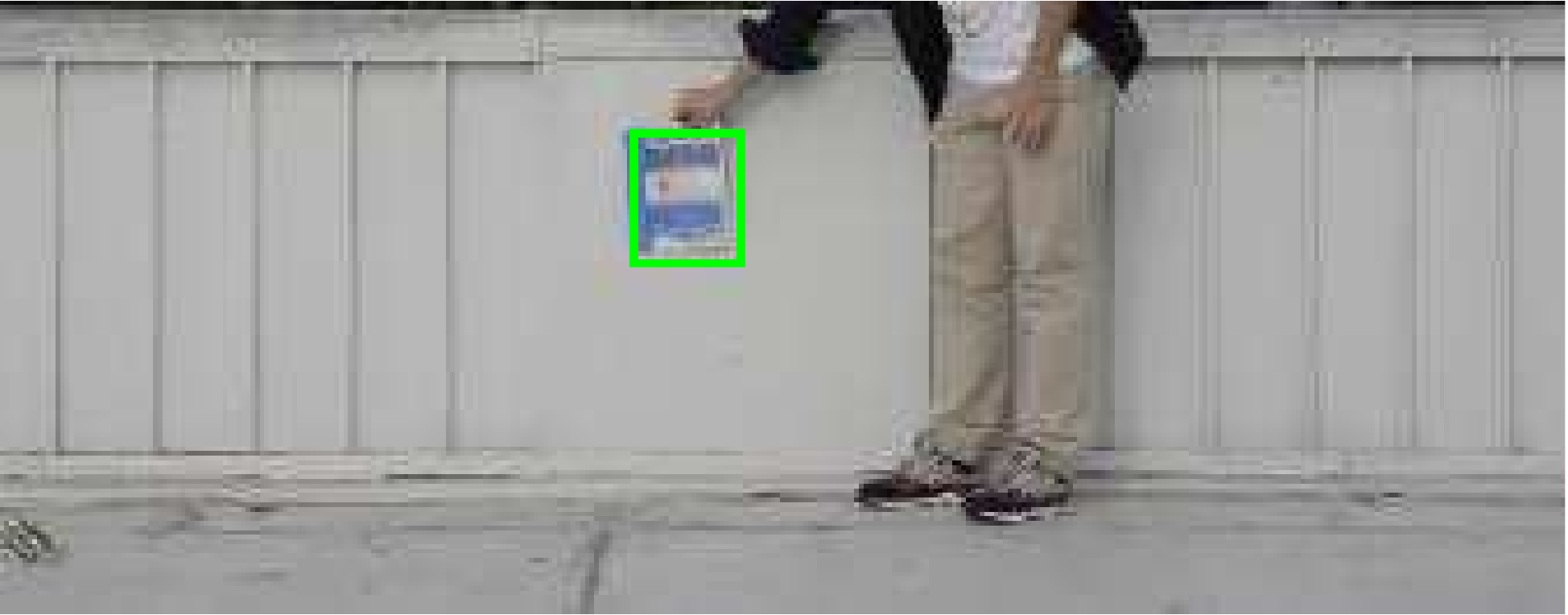}\\
  	\textbf{Ball}& \textbf{Basketball}&\textbf{Board1}&\textbf{Board2} &\textbf{Book}\\
  	SV, MB, OCC& FM, MB, OCC, LR&IPR, OPR, BC&IPR, OPR, BC&IPR, DEF, OPR\\

	\includegraphics[width=0.16\linewidth, height=0.12\linewidth, clip=true]{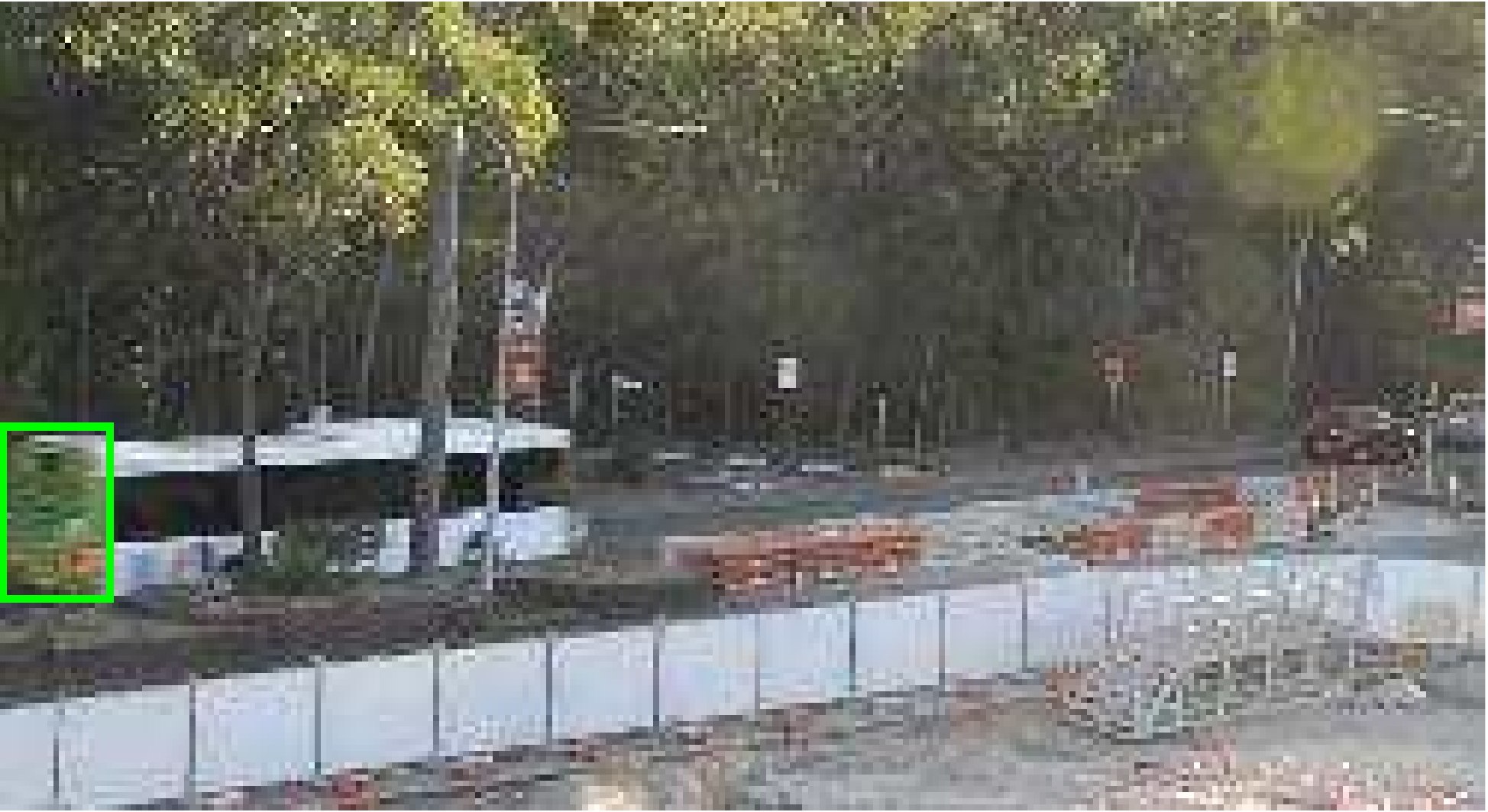}
	  &\includegraphics[width=0.16\linewidth, height=0.12\linewidth, clip=true]{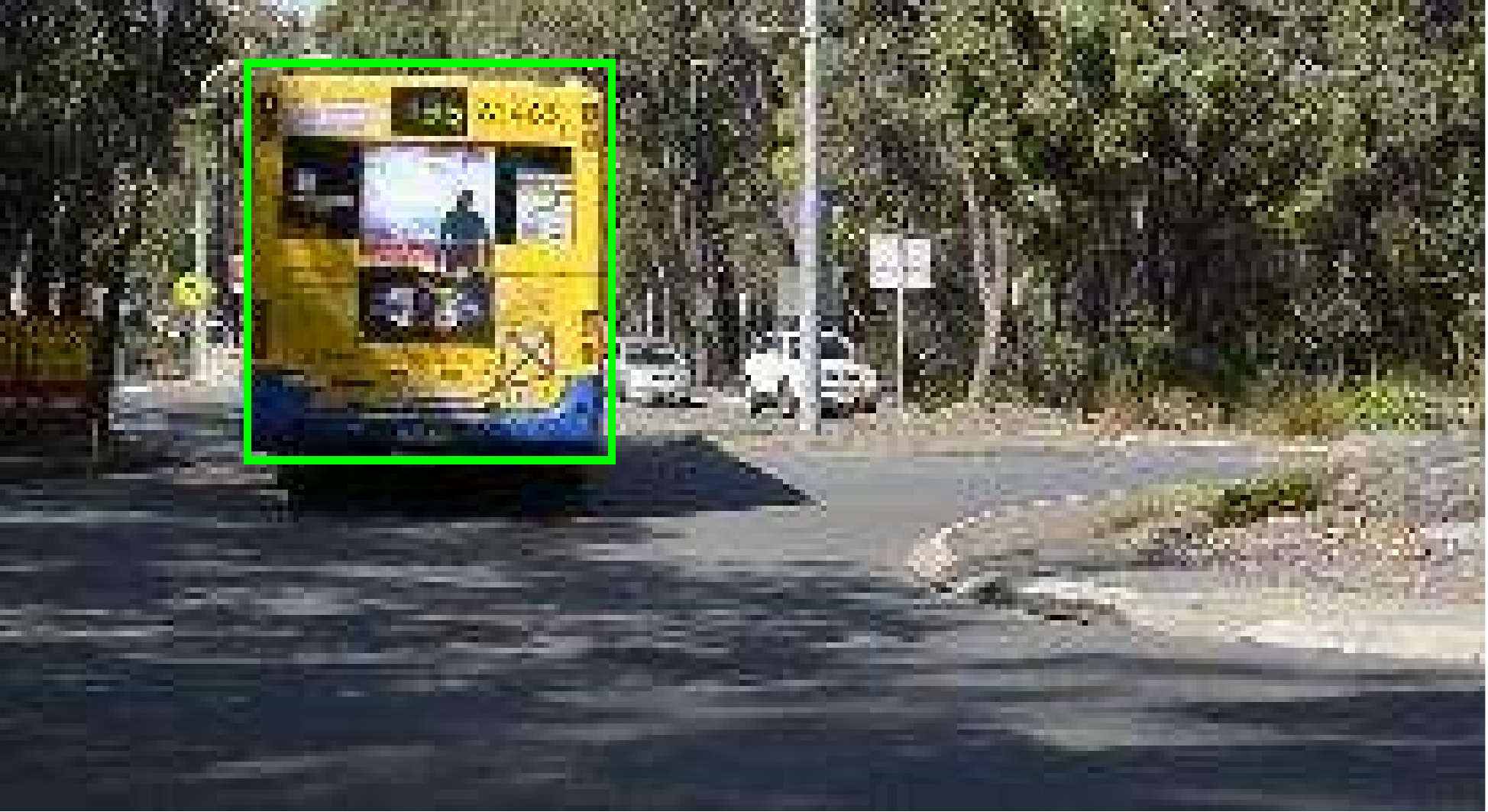}
  &\includegraphics[width=0.16\linewidth, height=0.12\linewidth, clip=true]{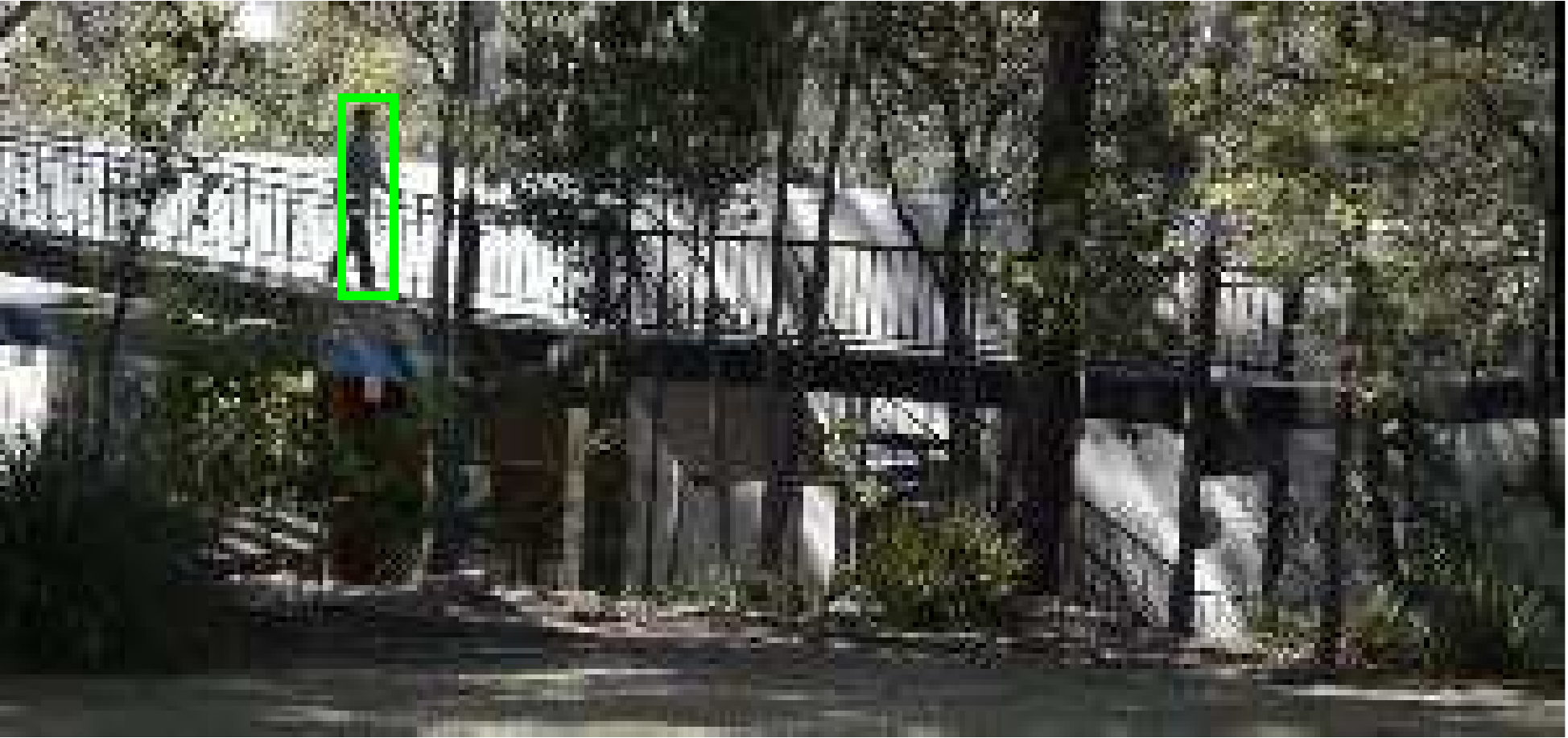}
	&\includegraphics[width=0.16\linewidth, height=0.12\linewidth, clip=true]{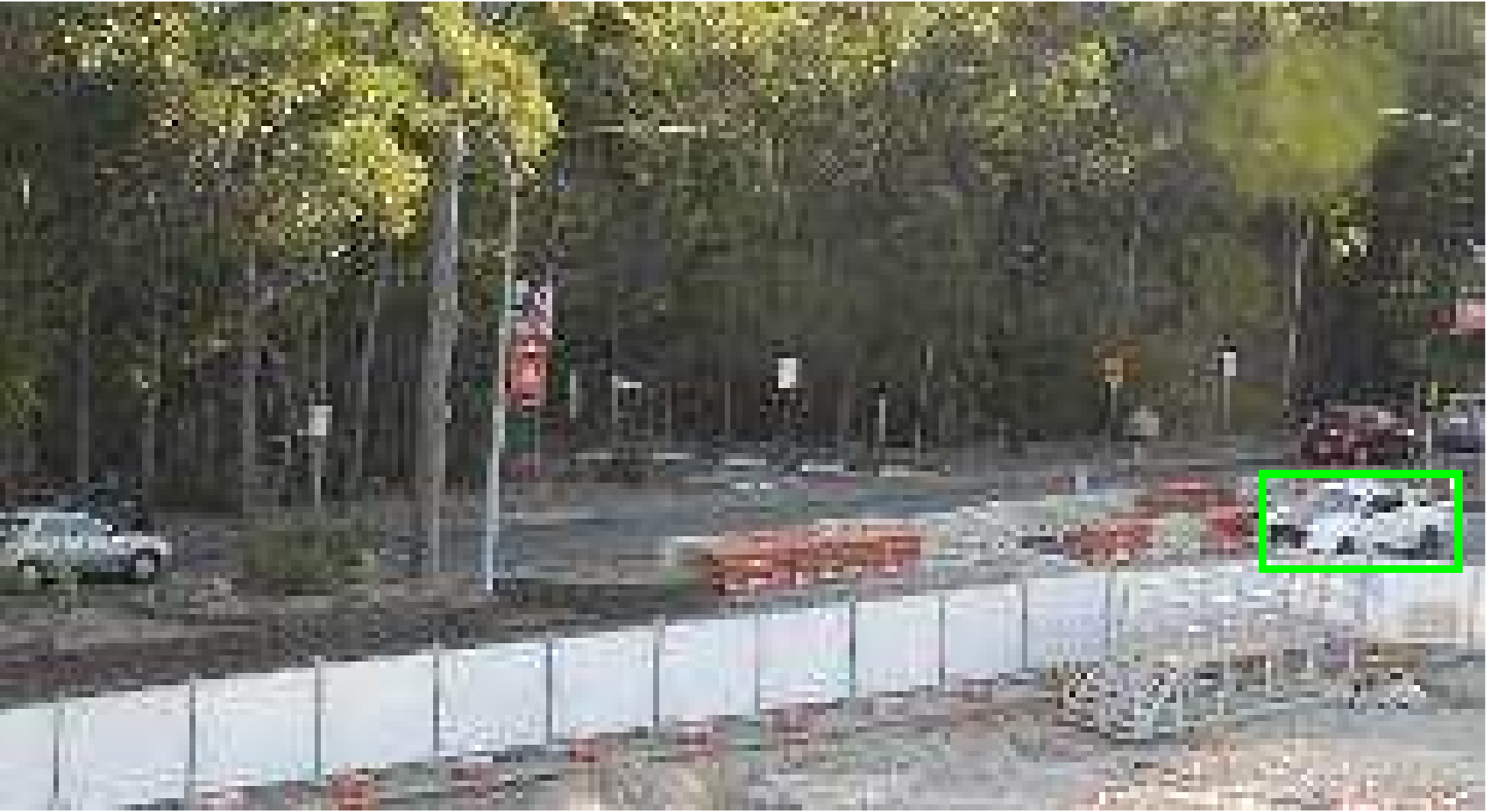}
	&\includegraphics[width=0.16\linewidth, height=0.12\linewidth, clip=true]{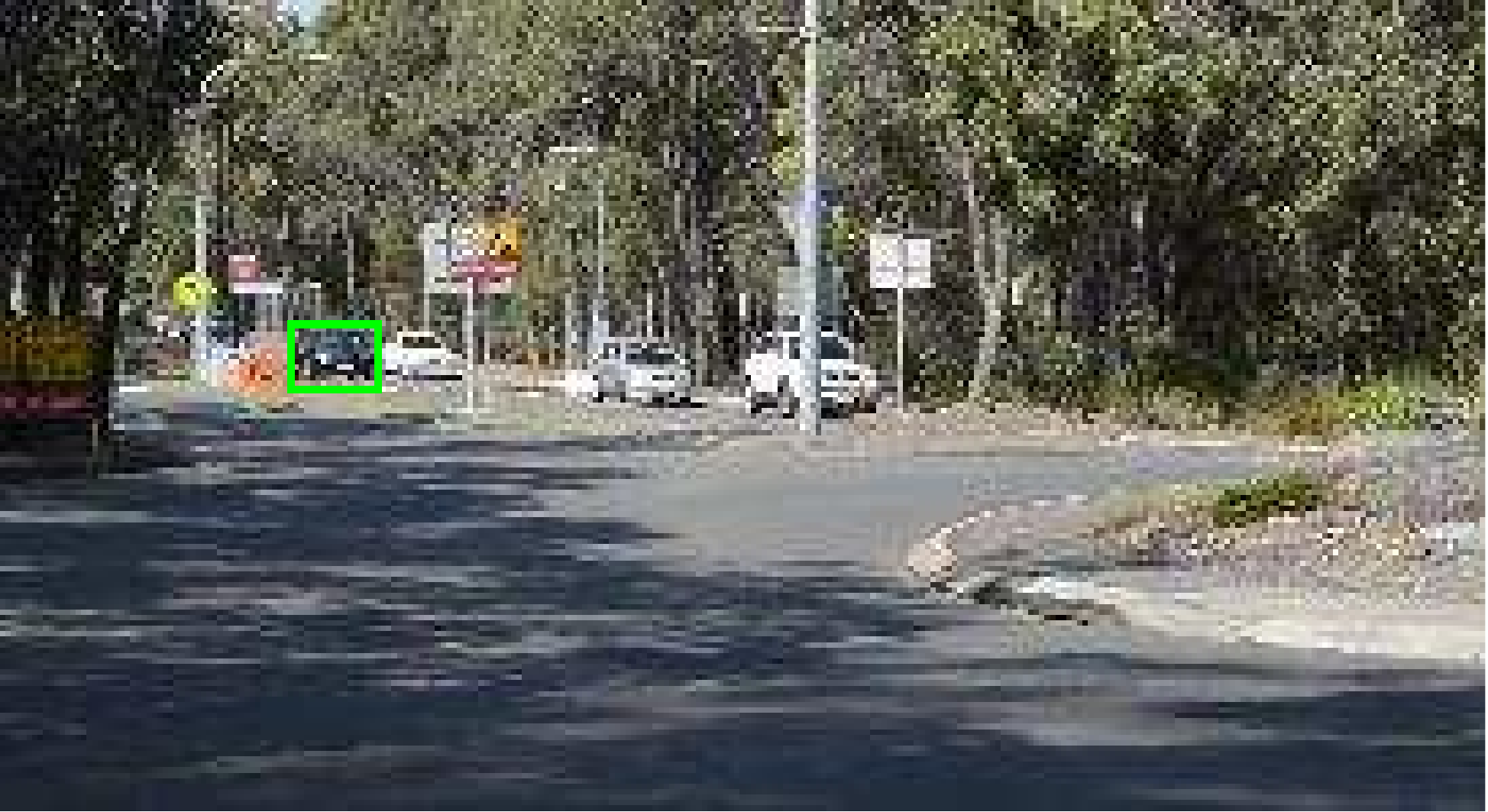}
	\\
	\textbf{Bus}&\textbf{Bus2}&\textbf{Campus} &\textbf{Car}&\textbf{Car2}\\
   LR, BC, FM&IV, SV, OCC, FM& IV, SV, OCC&SV, IPR, OPR& SV, IPR, OPR\\

  \includegraphics[width=0.16\linewidth, height=0.12\linewidth, clip=true]{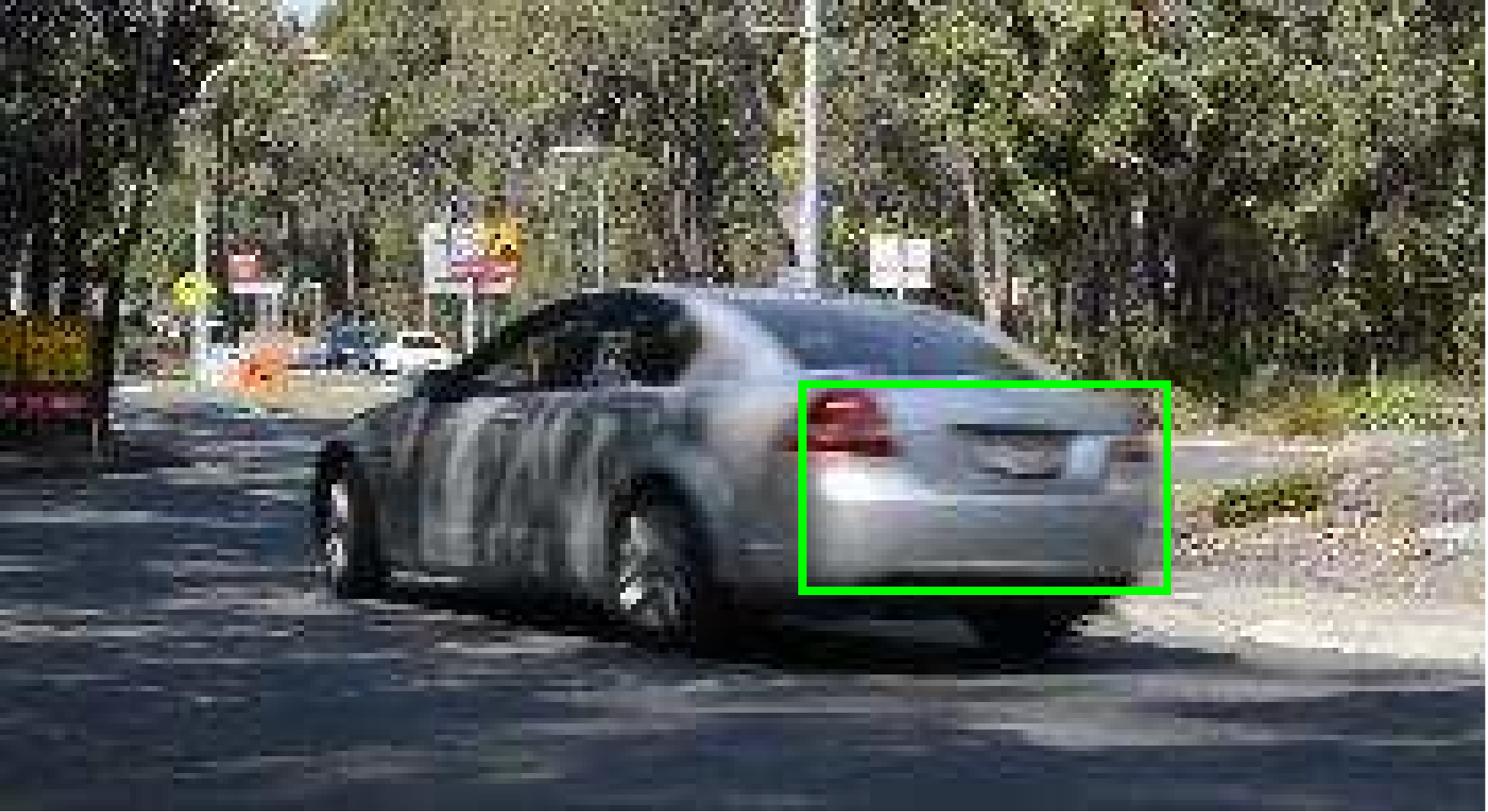}
  &\includegraphics[width=0.16\linewidth, height=0.12\linewidth, clip=true]{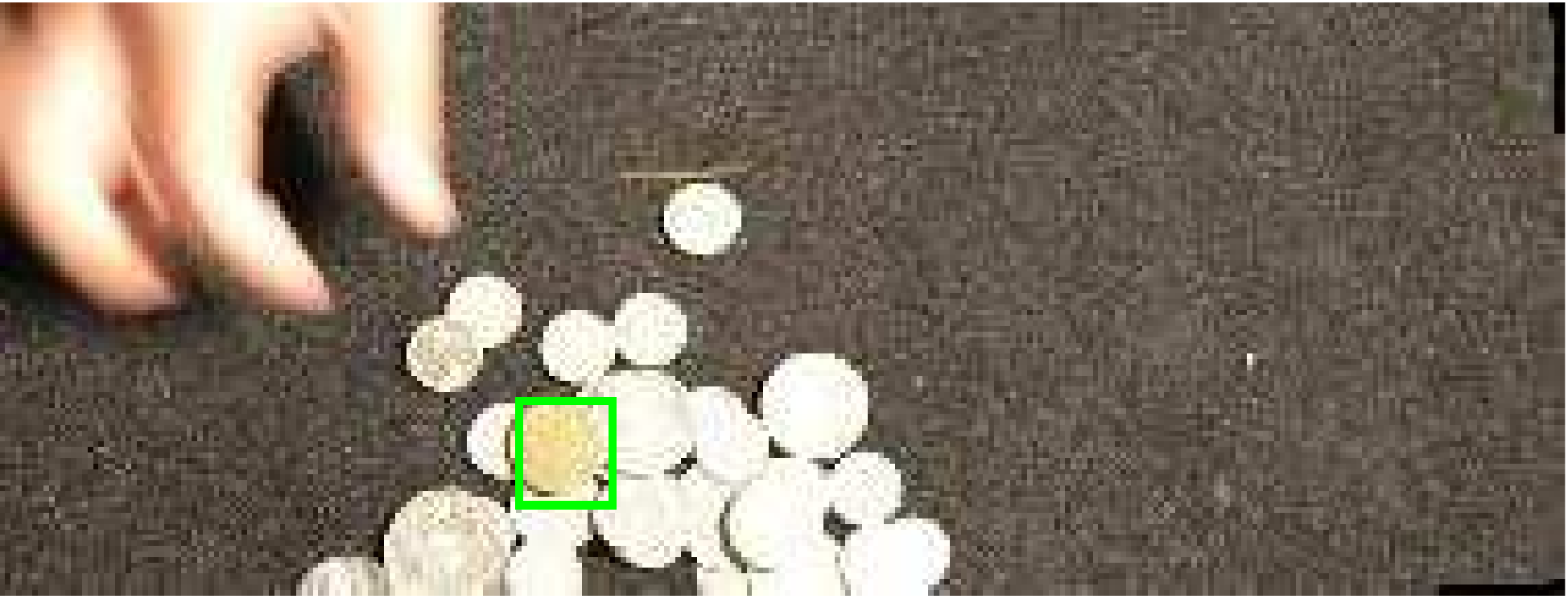}
  	 &\includegraphics[width=0.16\linewidth, height=0.12\linewidth, clip=true]{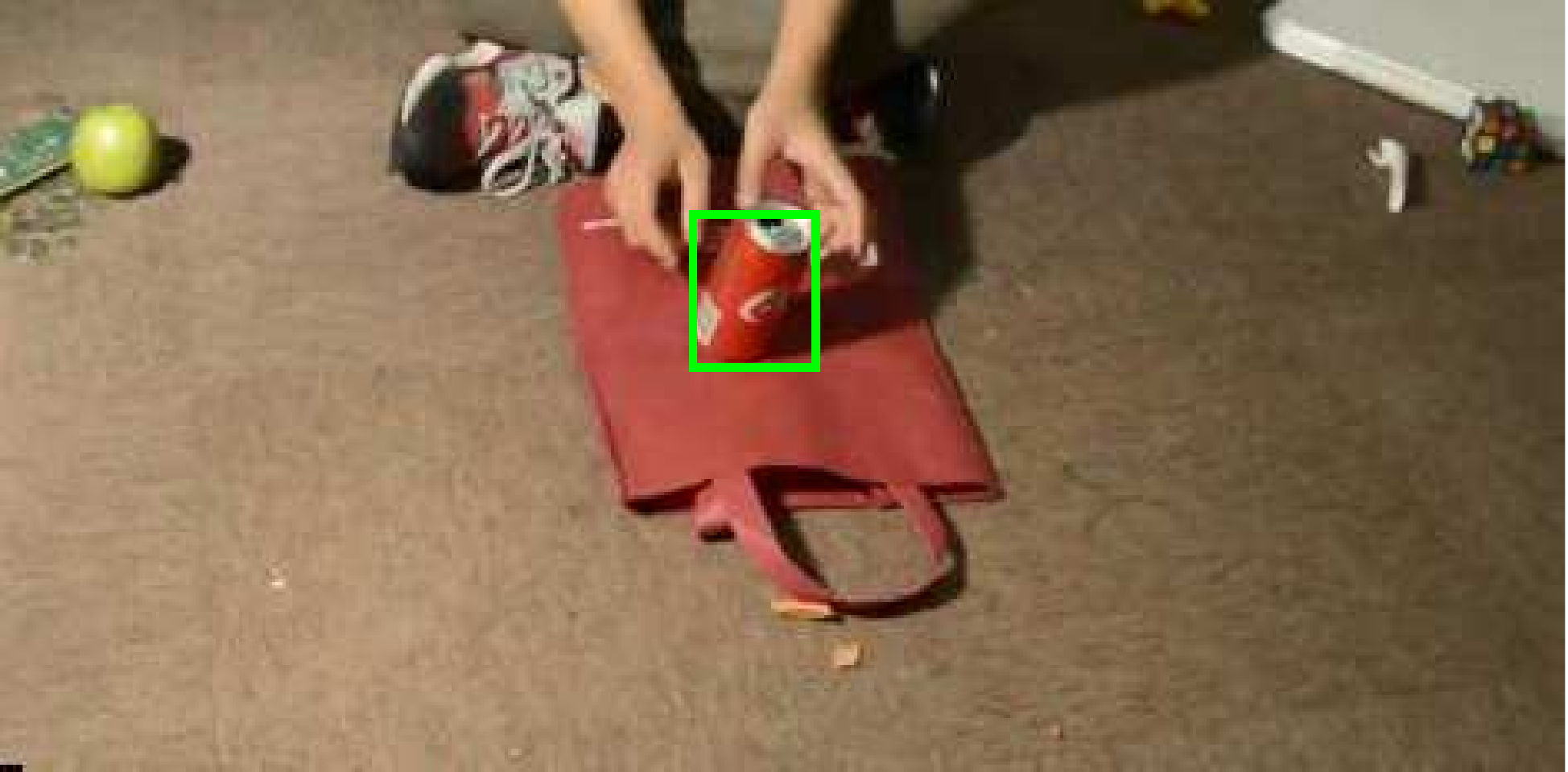}
	&\includegraphics[width=0.16\linewidth, height=0.12\linewidth, clip=true]{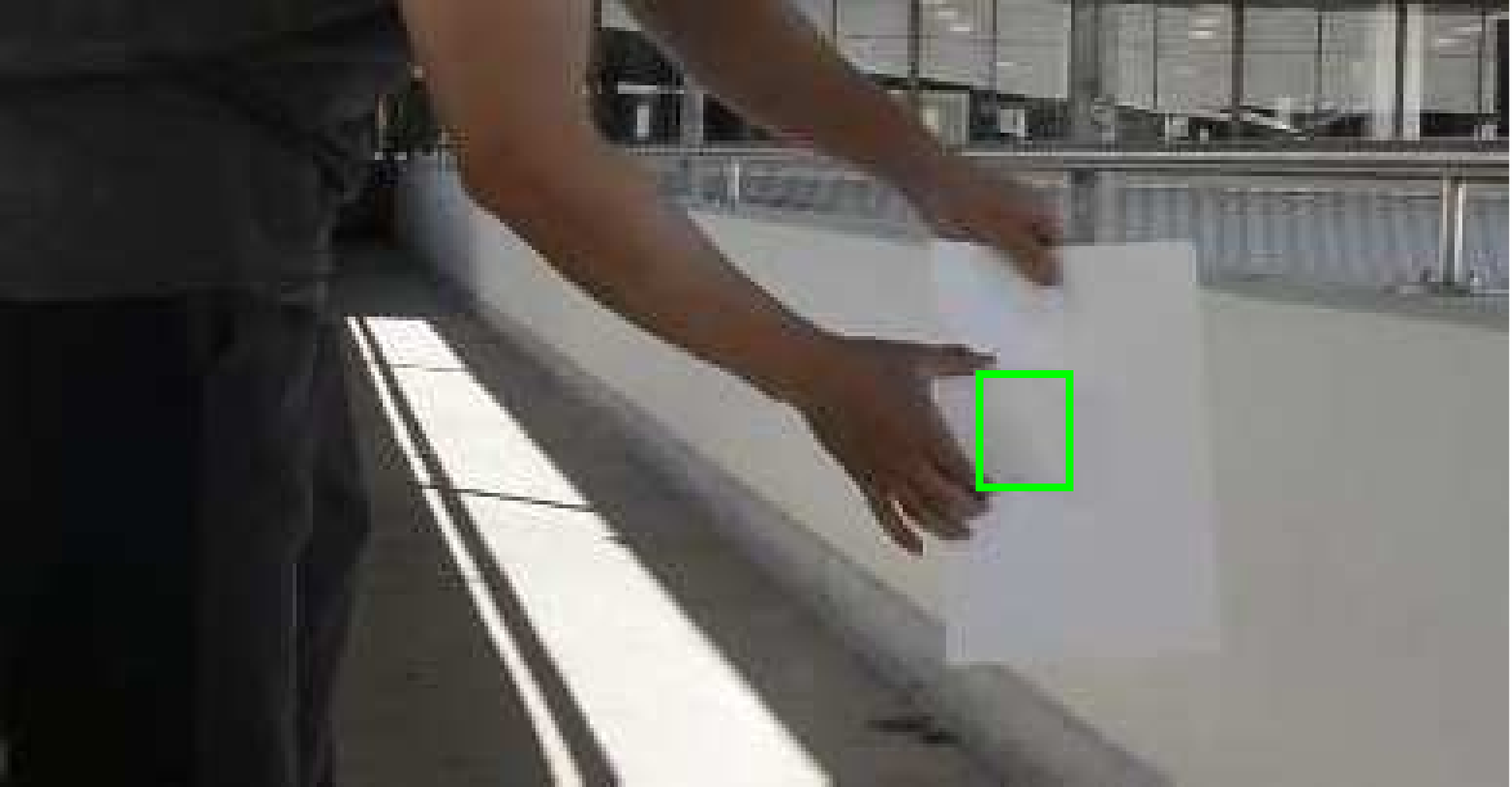}
 &\includegraphics[width=0.16\linewidth, height=0.12\linewidth, clip=true]{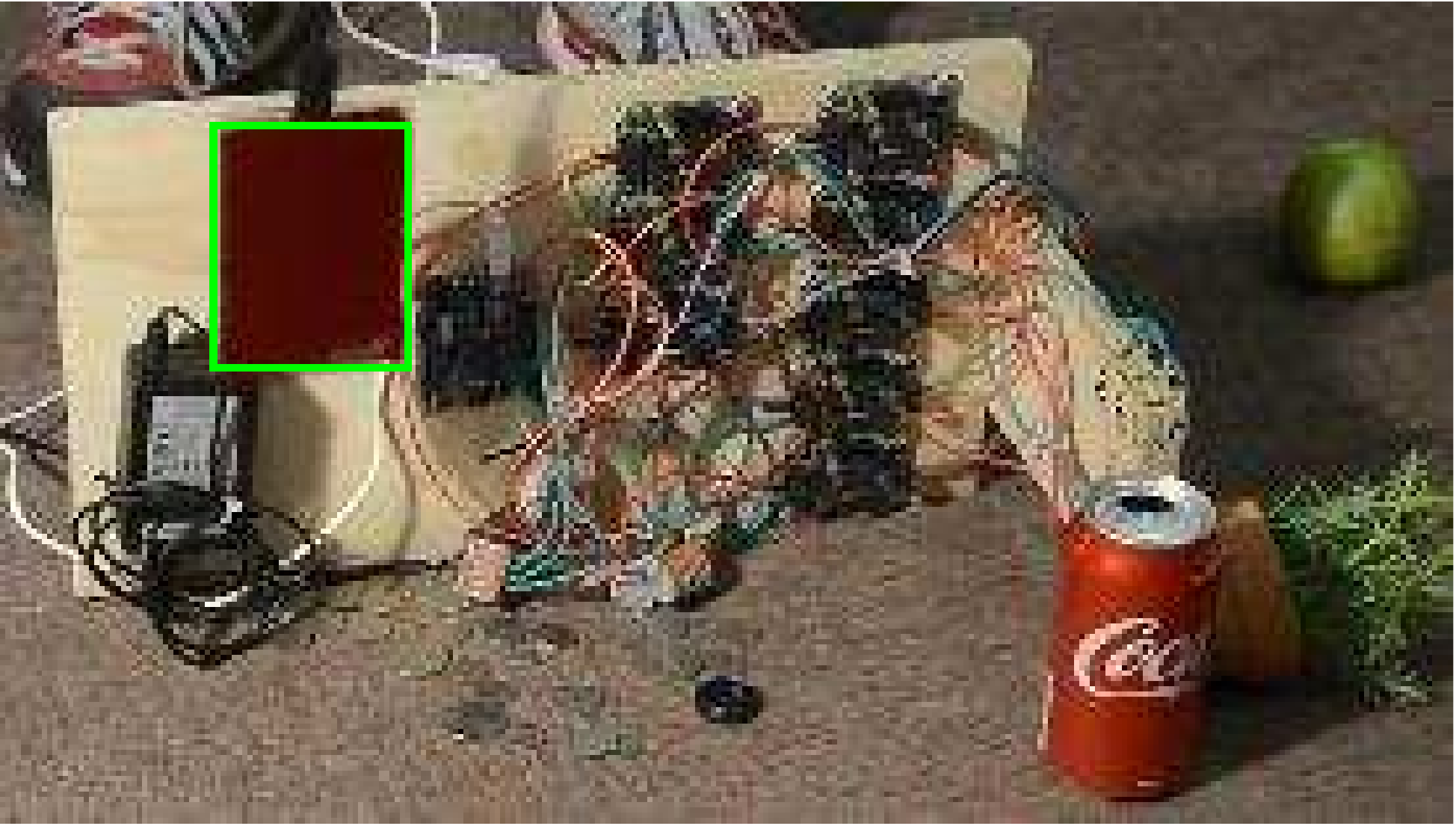}\\

\textbf{Car3} &\textbf{Coin}&\textbf{Coke}&\textbf{Card}&\textbf{Drive}\\
SV, IPR, OPR, OCC, IV&BC&BC, IPR, OPR, FM, SV&IPR, BC, OCC&BC, IPR, OPR, SV\\
\includegraphics[width=0.16\linewidth, height=0.12\linewidth, clip=true]{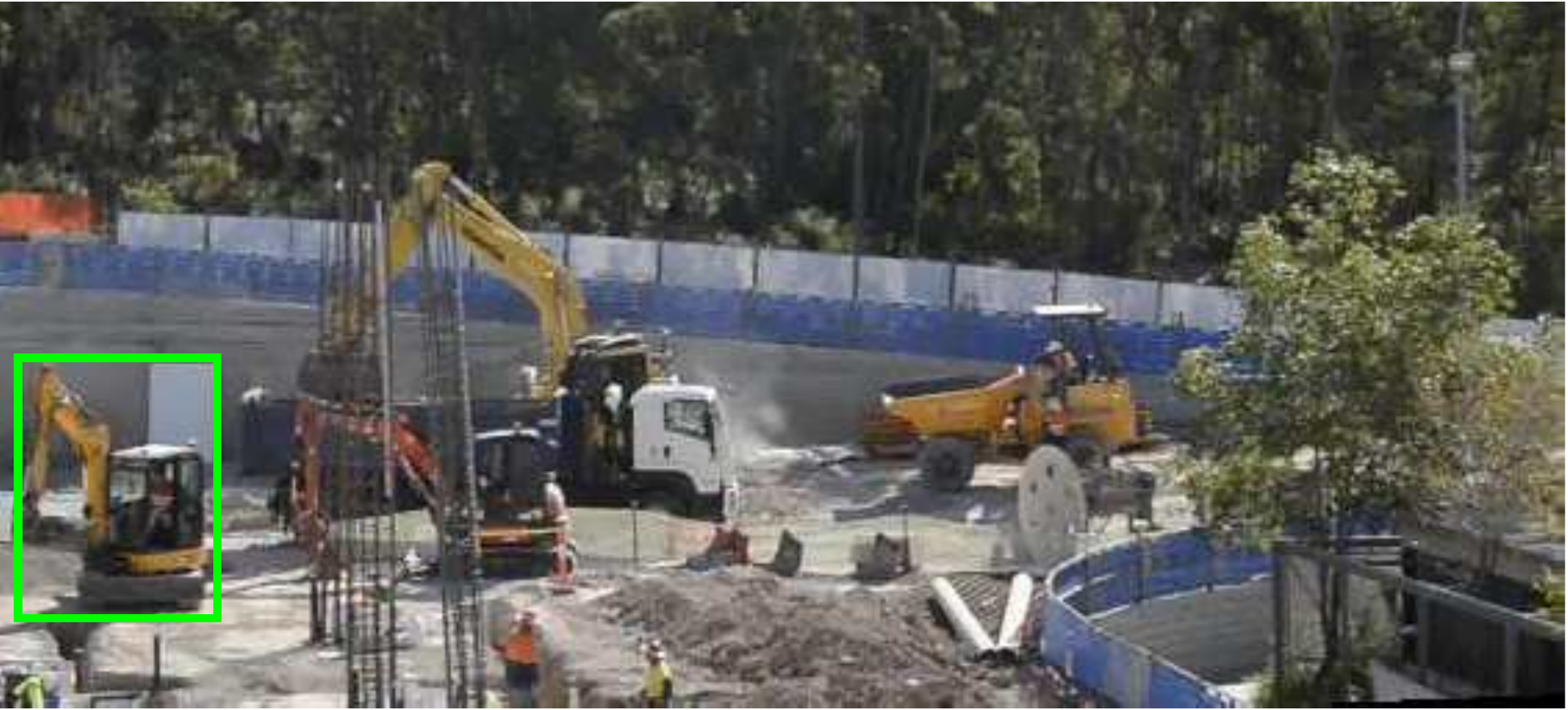}
  &\includegraphics[width=0.16\linewidth, height=0.12\linewidth, clip=true]{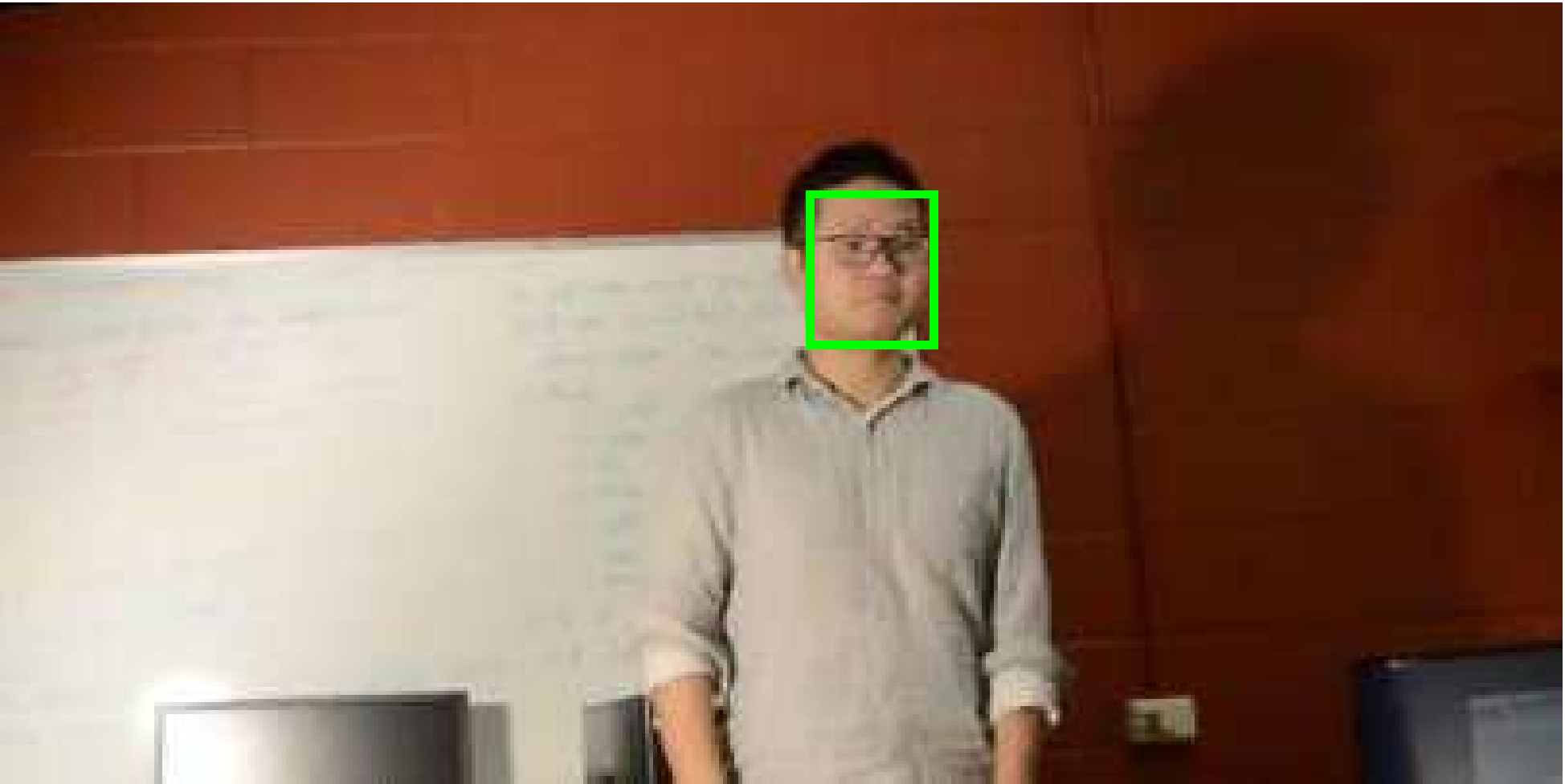}
  &\includegraphics[width=0.16\linewidth, height=0.12\linewidth, clip=true]{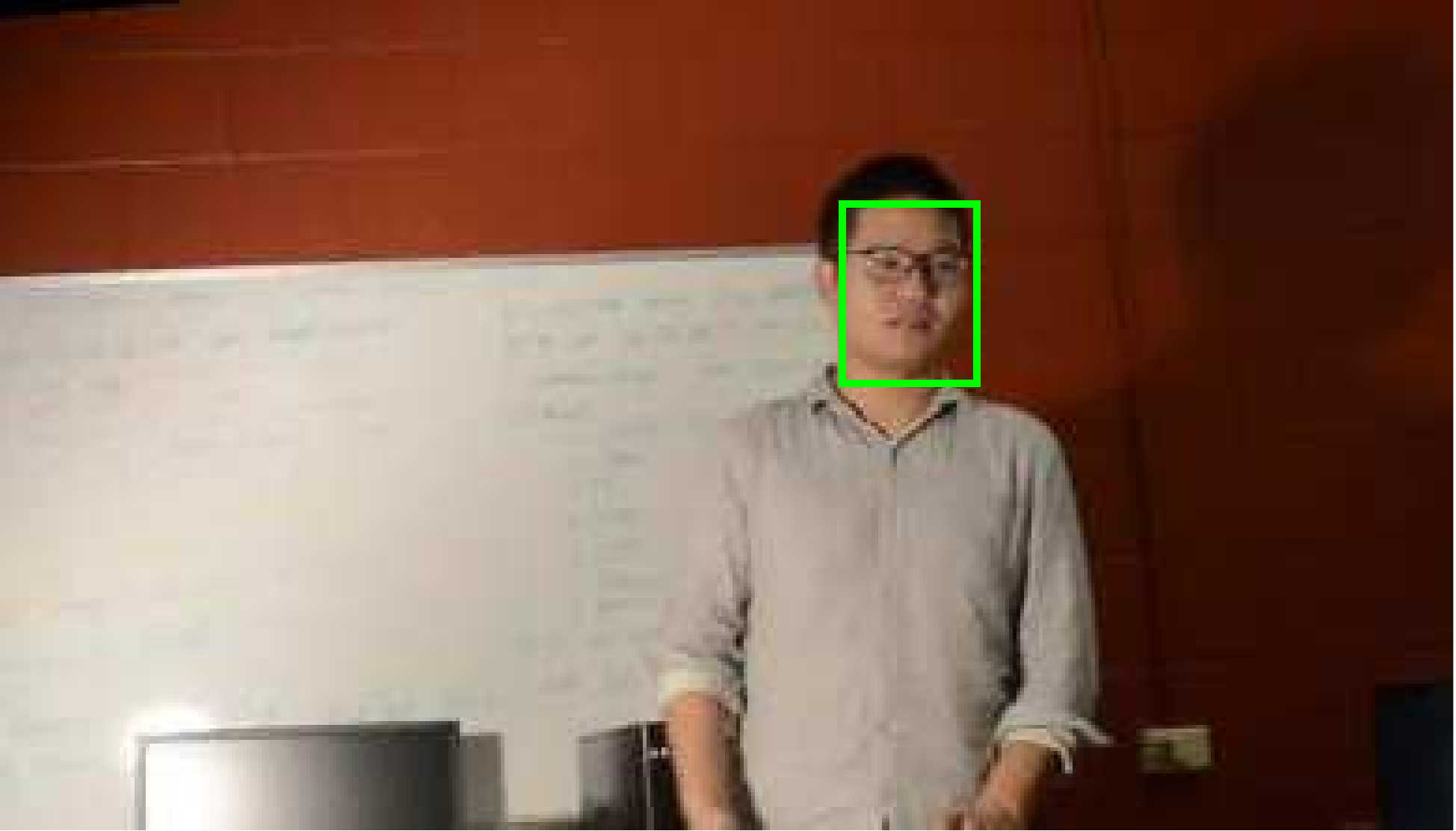}
&\includegraphics[width=0.16\linewidth, height=0.12\linewidth, clip=true]{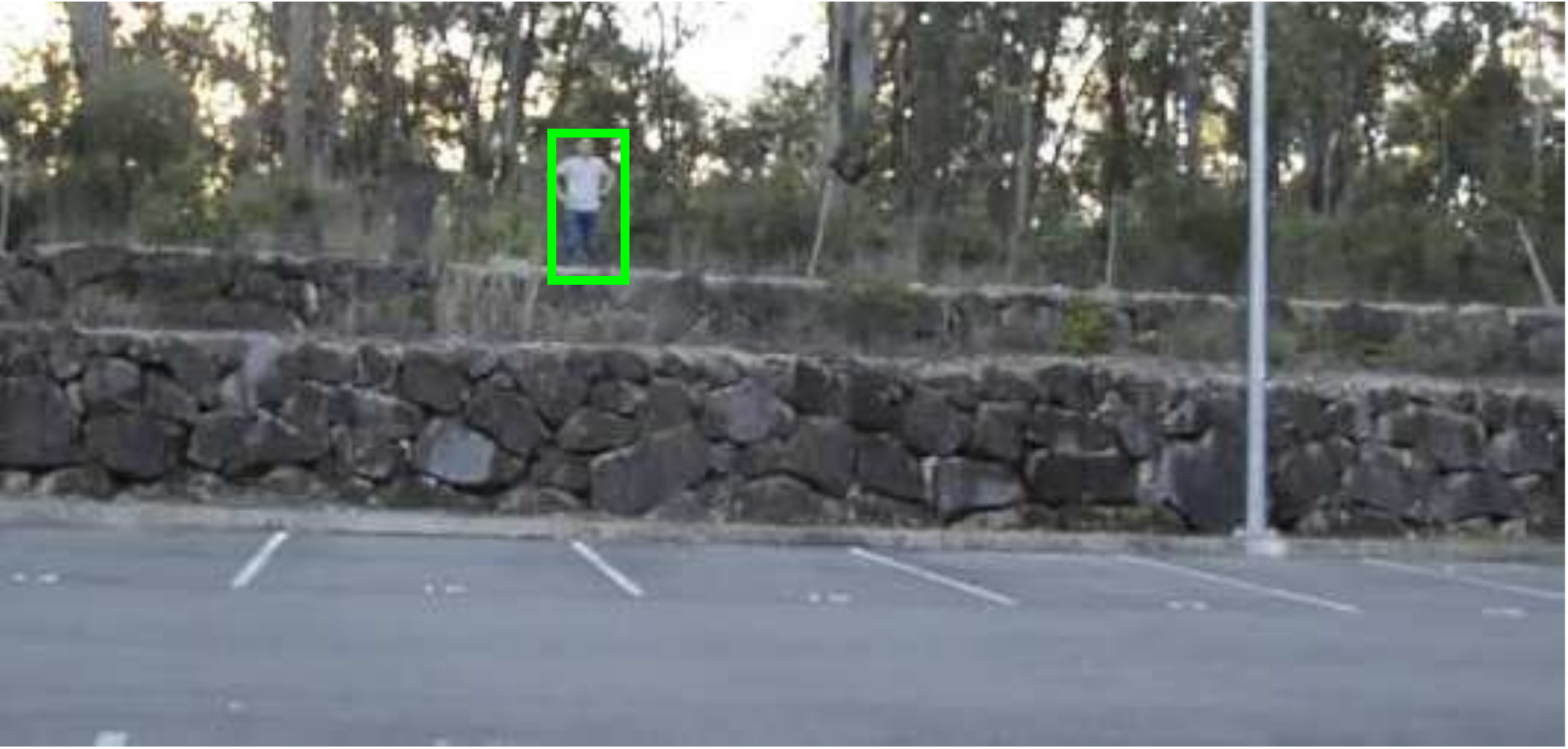}
&\includegraphics[width=0.16\linewidth, height=0.12\linewidth, clip=true]{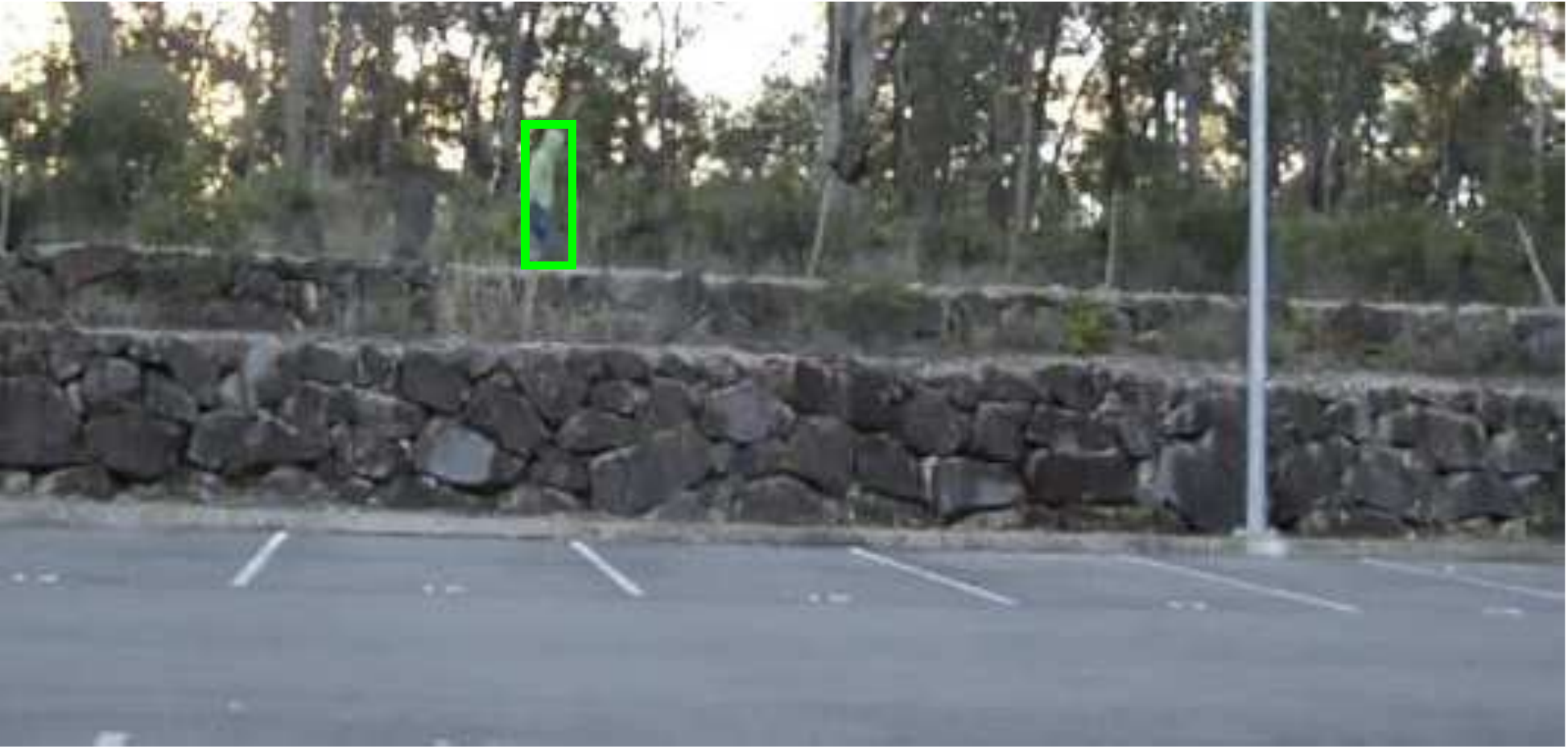}\\
  \textbf{Excavator}&\textbf{Face}&\textbf{Face2}&\textbf{Forest}&\textbf{Forest2}\\
  IPR, OPR, SV, OCC, DEF&IPR, OPR, SV, MB&IPR, OPR, SV, OCC&BC, OCC&BC, OCC\\

\includegraphics[width=0.16\linewidth, height=0.12\linewidth, clip=true]{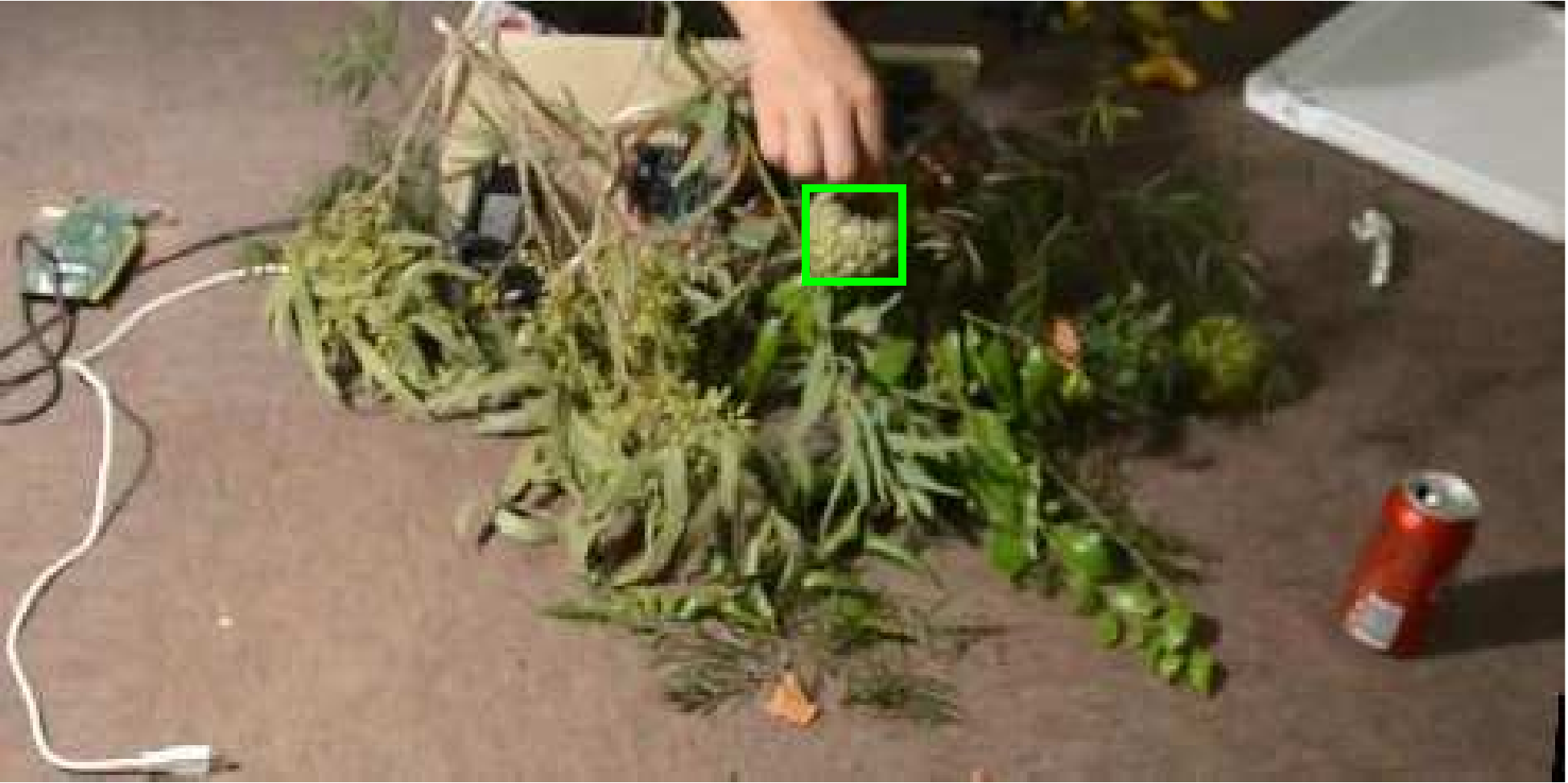}
	&\includegraphics[width=0.16\linewidth, height=0.12\linewidth, clip=true]{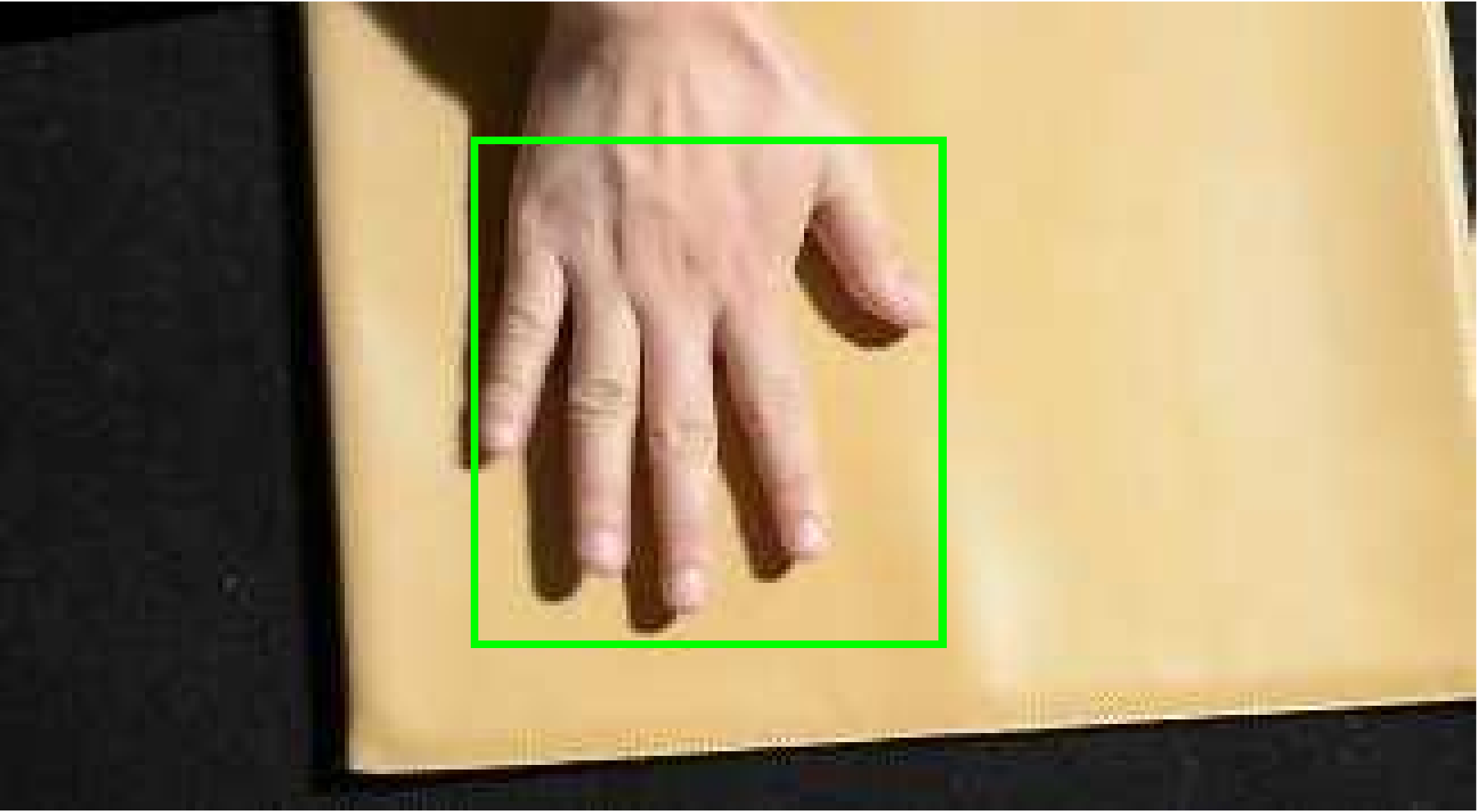}
&\includegraphics[width=0.16\linewidth, height=0.12\linewidth, clip=true]{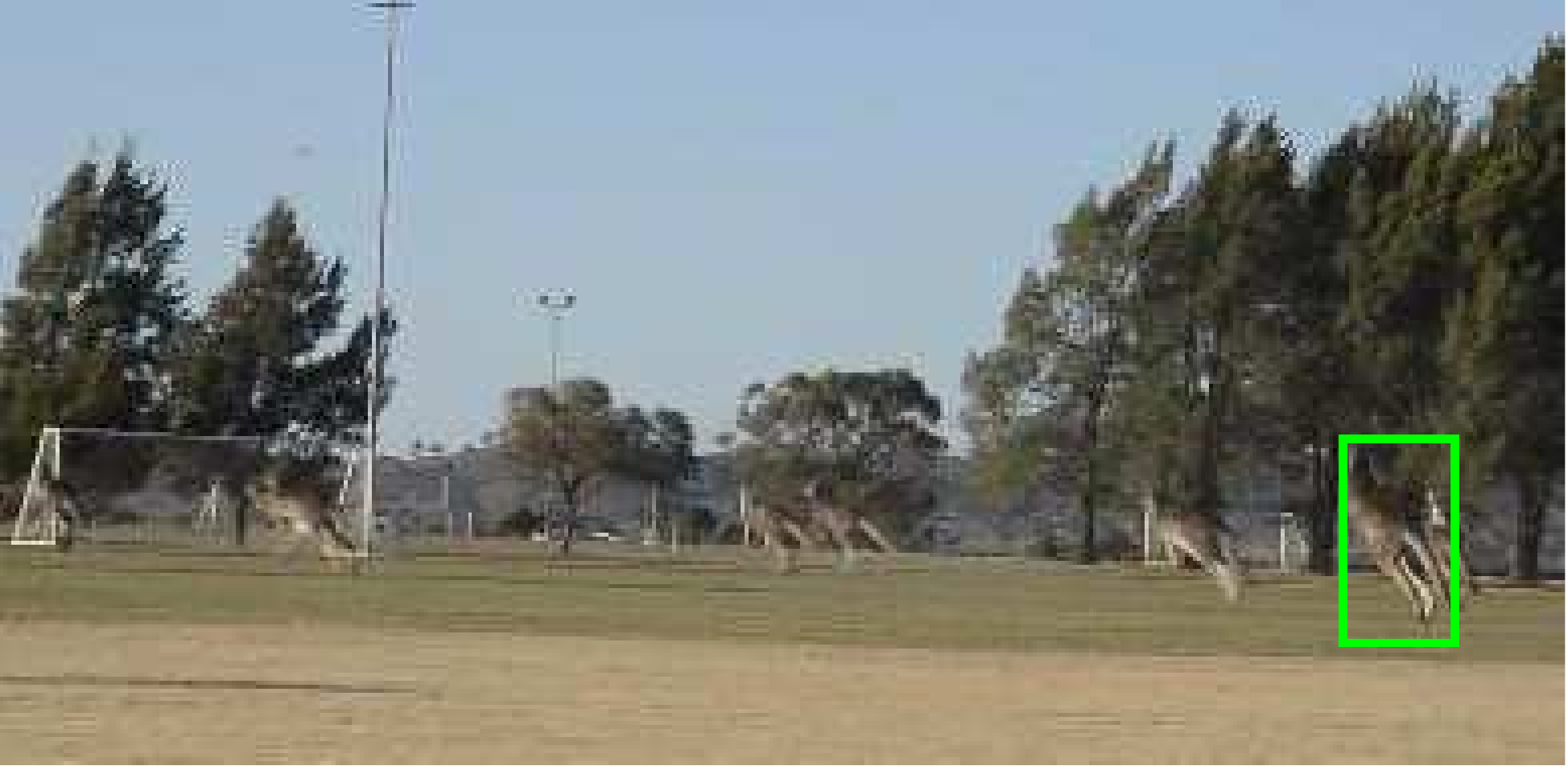}
&\includegraphics[width=0.16\linewidth, height=0.12\linewidth, clip=true]{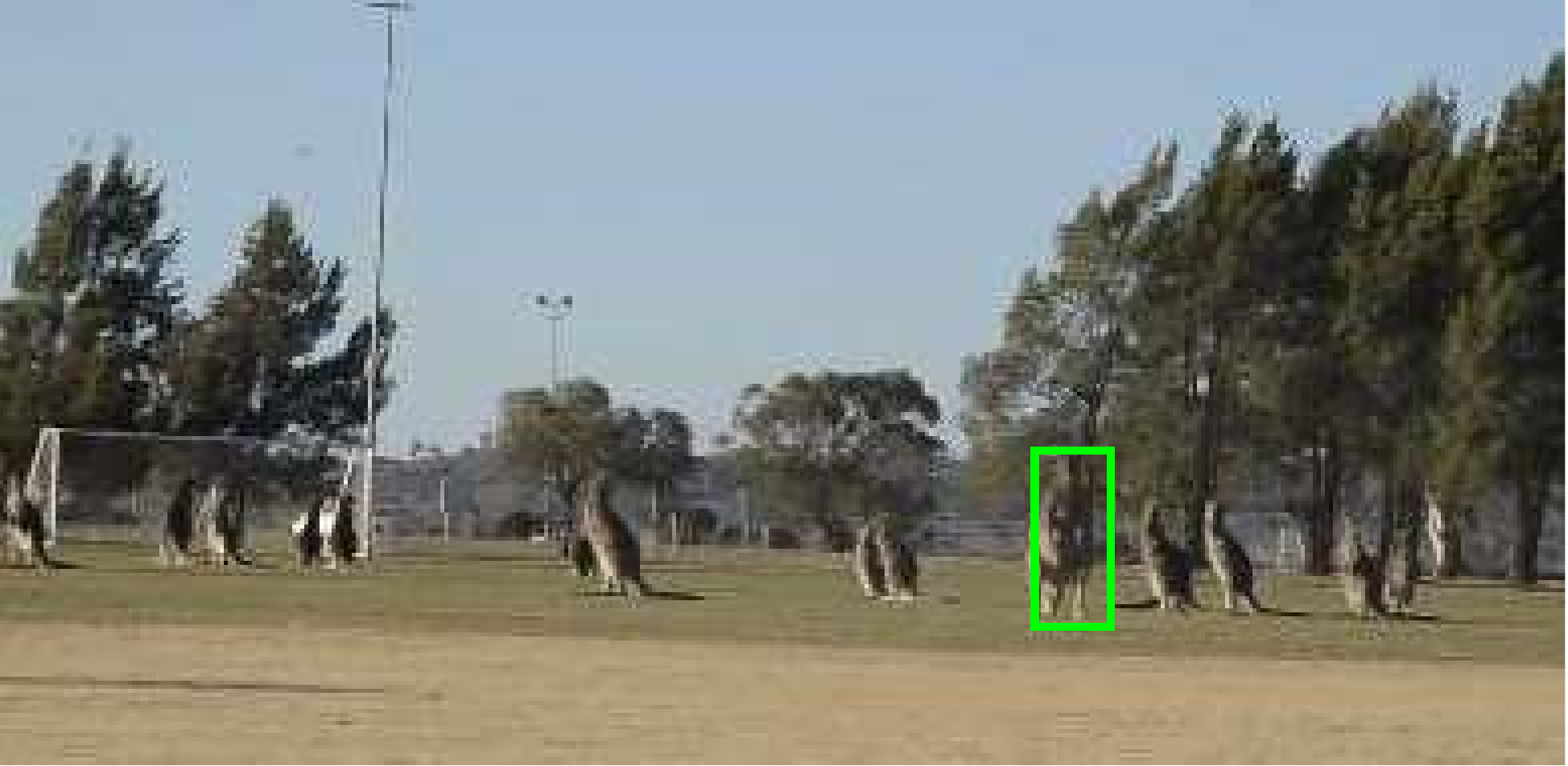}
&\includegraphics[width=0.16\linewidth, height=0.12\linewidth, clip=true]{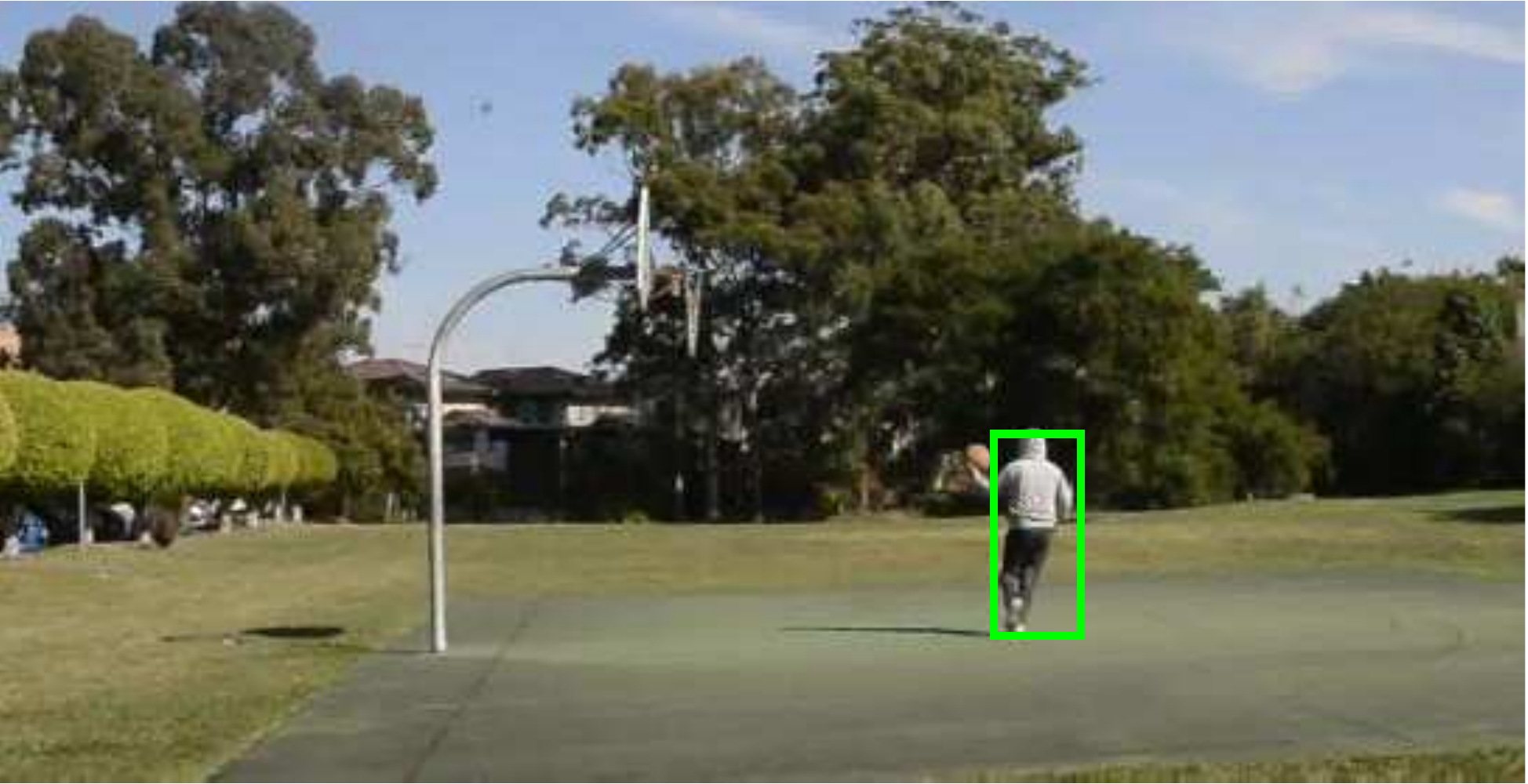}
  \\
\textbf{Fruit}&\textbf{Hand}&\textbf{Kangaroo}&\textbf{Kangaroo2}&\textbf{Player}\\
BC, OCC&BC, SV, DEF, OPR&	BC, SV, DEF, OPR, MB&BC, SV, DEF,OPR&IPR, DFF, OPR, SV\\
\includegraphics[width=0.16\linewidth, height=0.12\linewidth, clip=true]{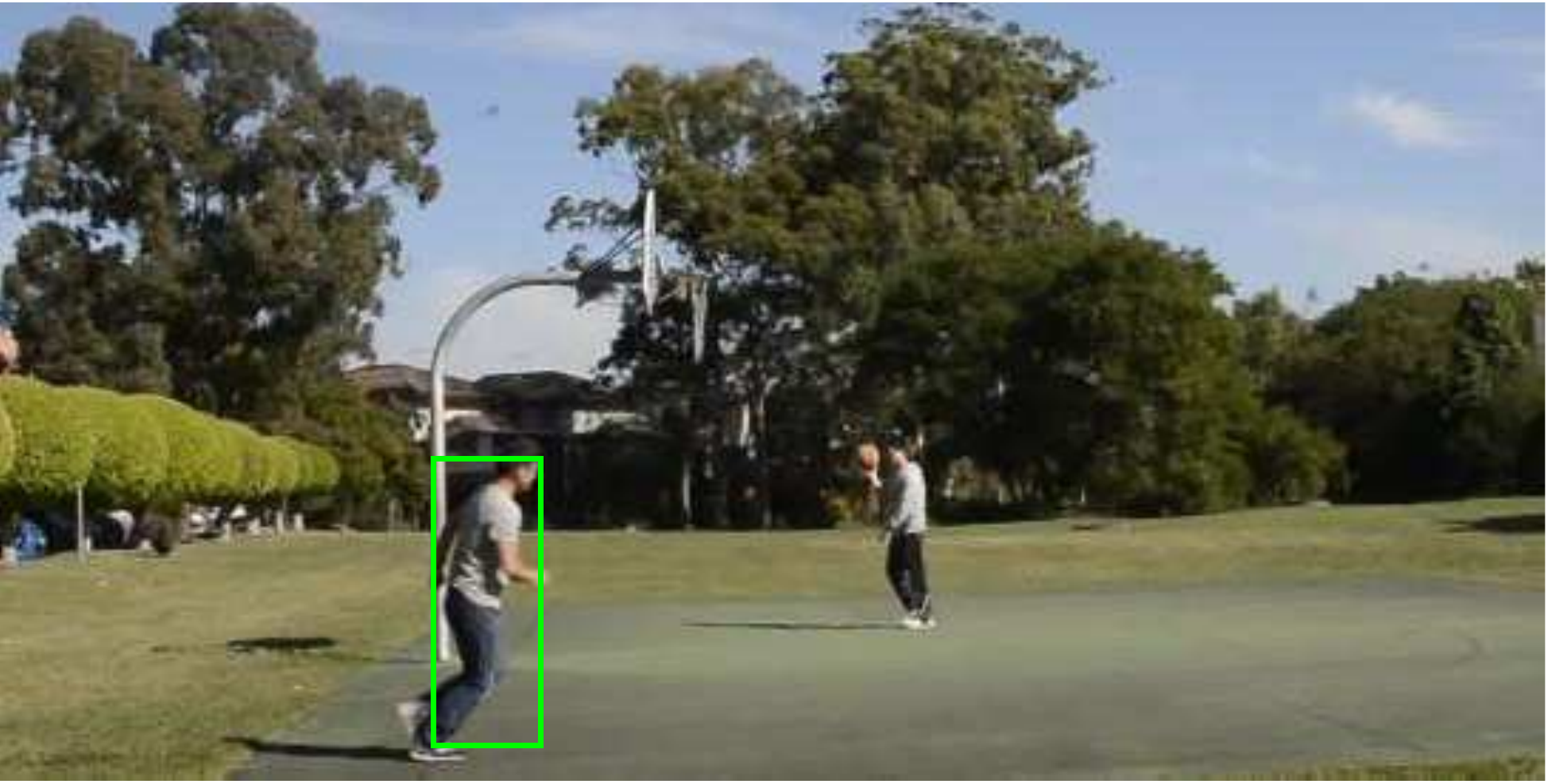}
    &\includegraphics[width=0.16\linewidth, height=0.12\linewidth, clip=true]{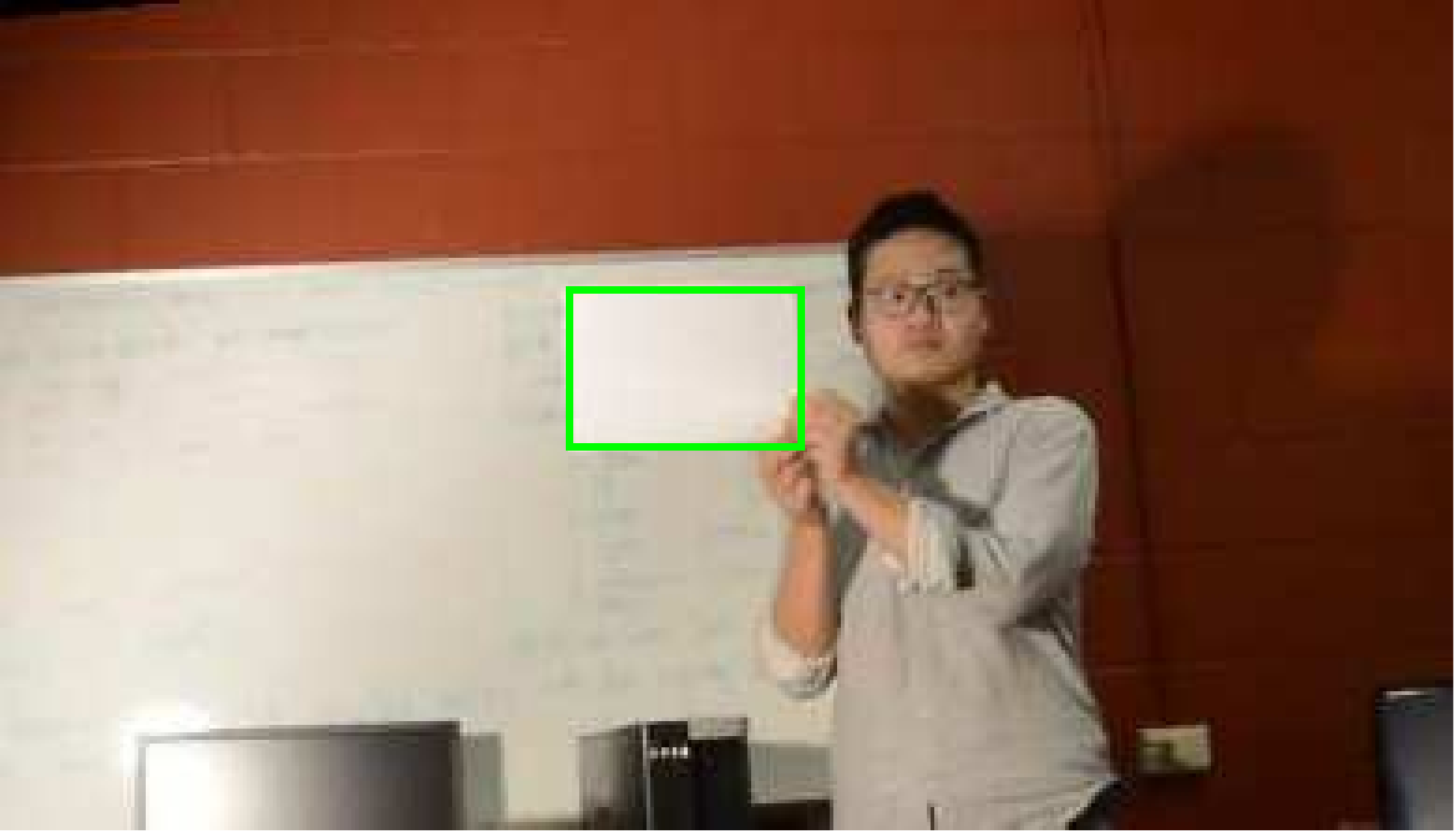}
     &\includegraphics[width=0.16\linewidth, height=0.12\linewidth, clip=true]{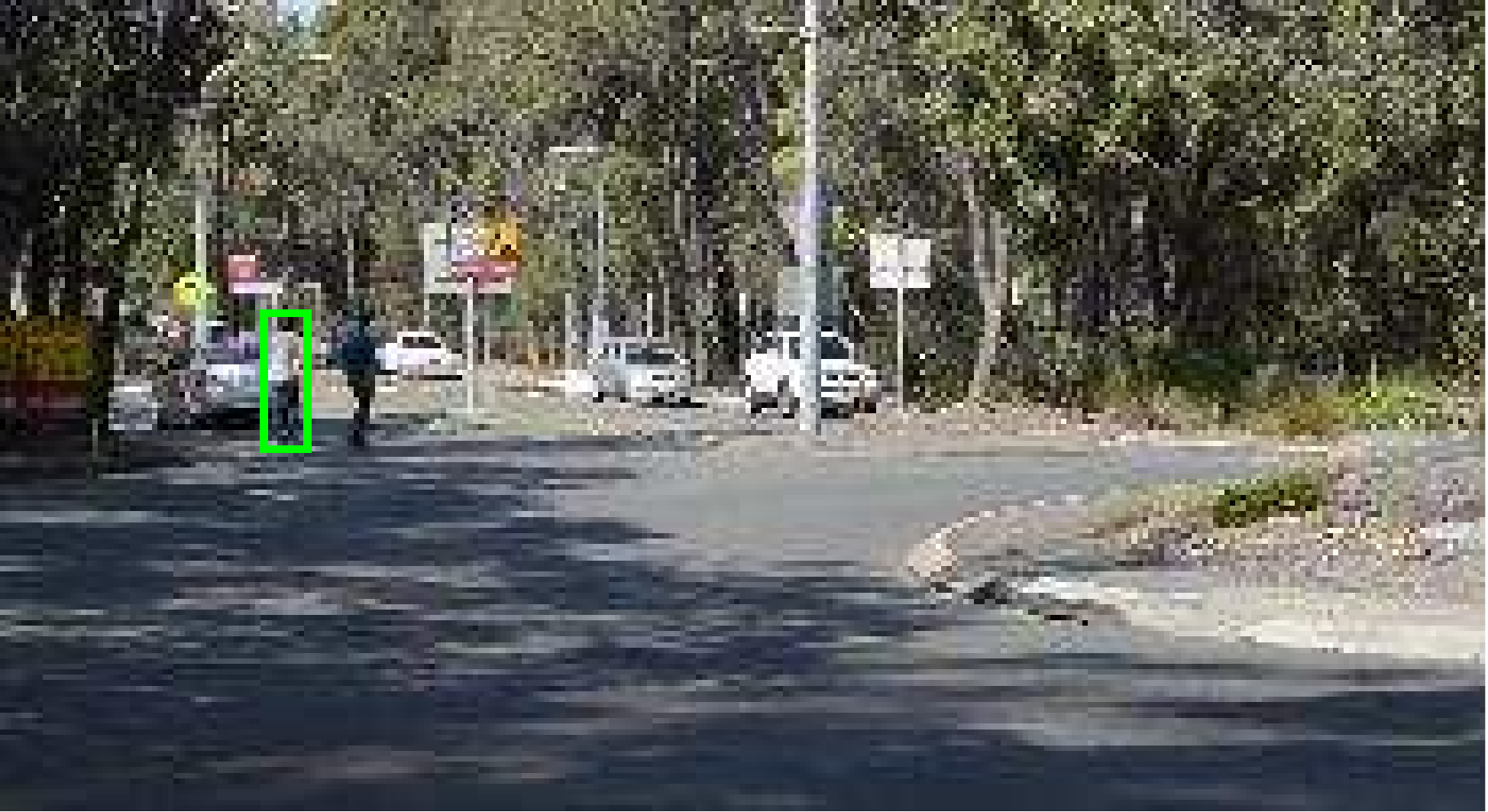}
	&\includegraphics[width=0.16\linewidth, height=0.12\linewidth, clip=true]{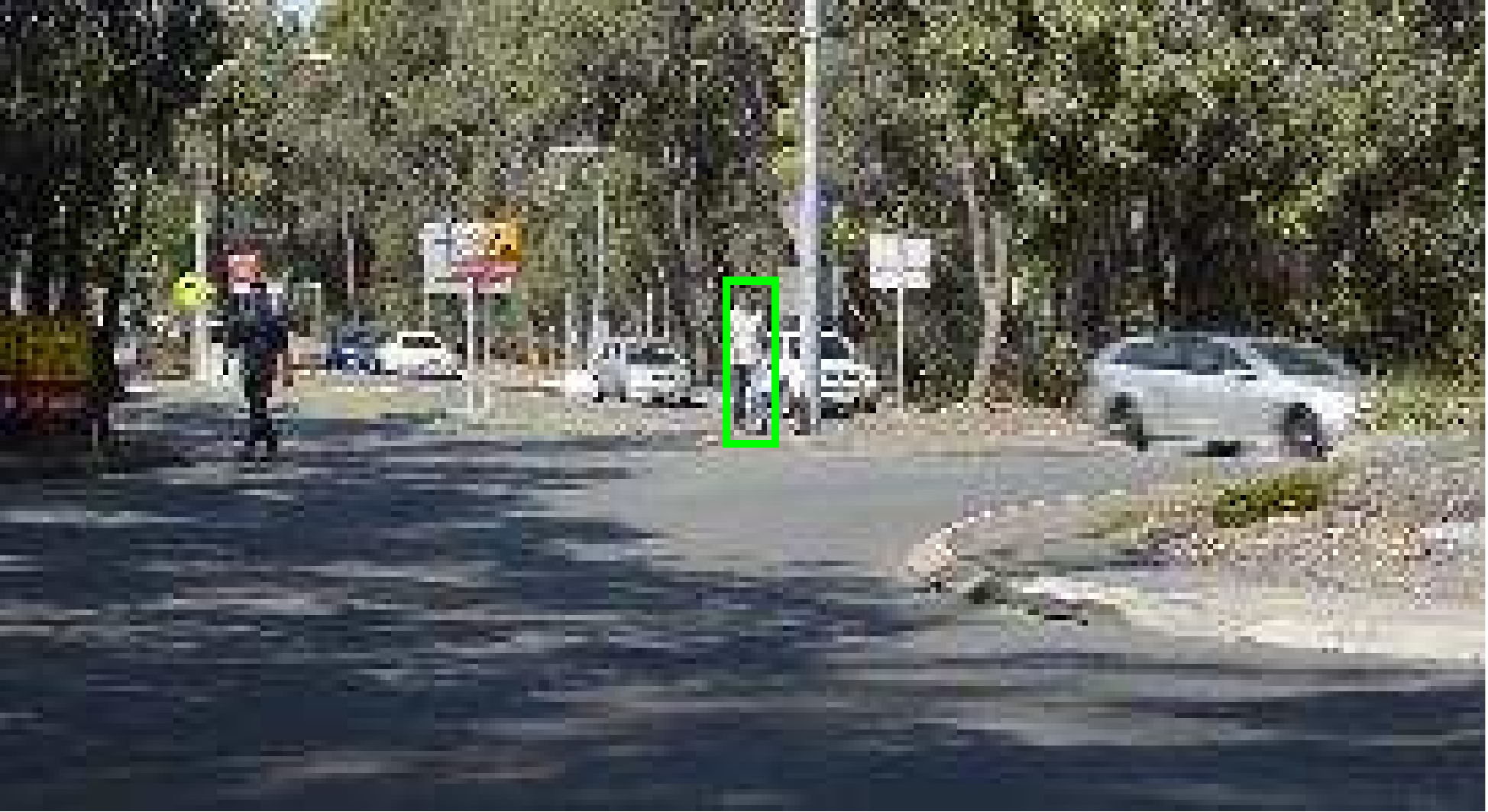}
	&\includegraphics[width=0.16\linewidth, height=0.12\linewidth, clip=true]{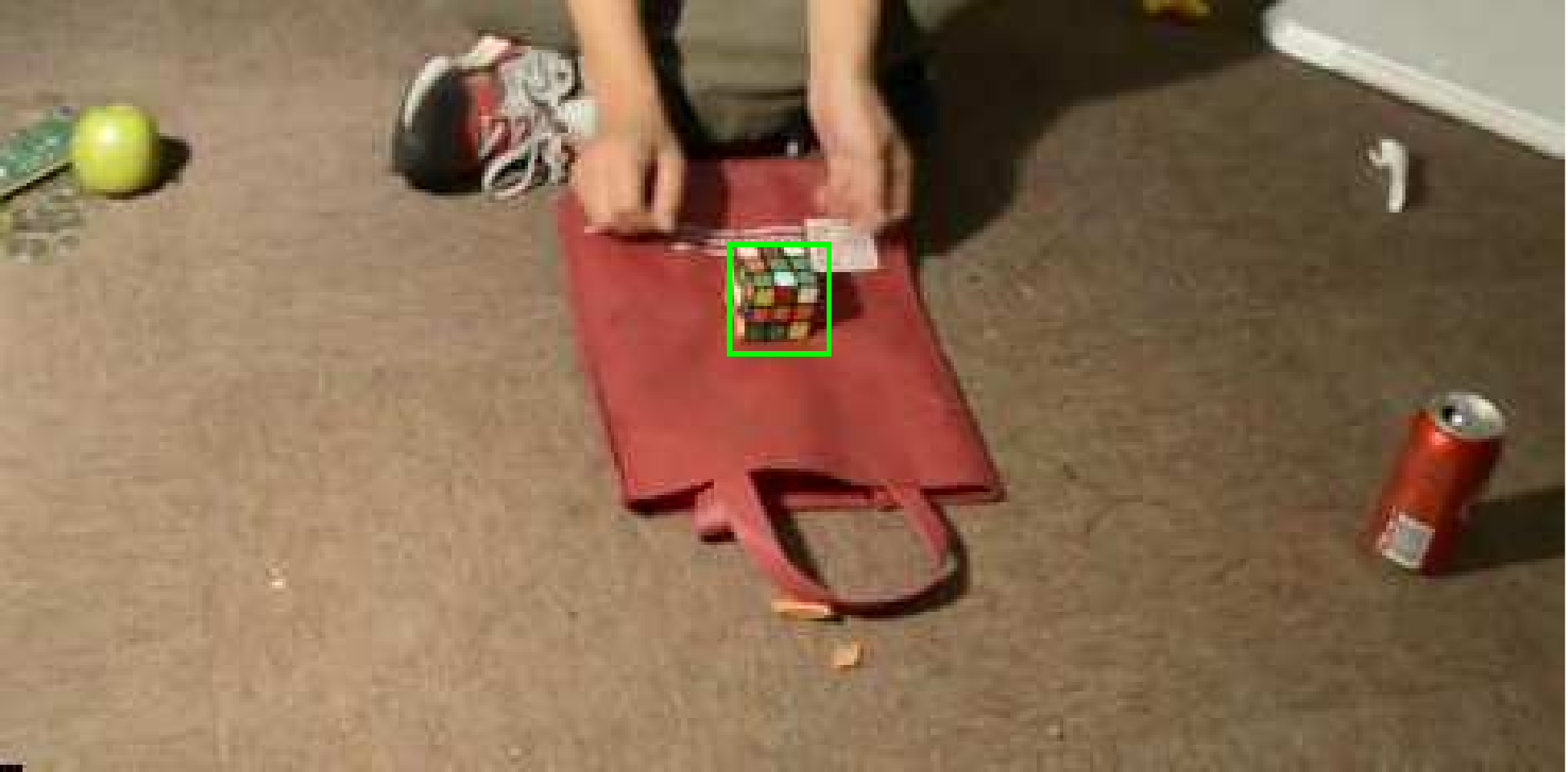}\\
\textbf{Playground}&\textbf{Paper}&\textbf{Pedestrian}&\textbf{Pedestrian2}&\textbf{Rubik}\\
SV, OCC &IPR, BC&IV, SV&IV, SV, OCC&DEF, IPR, OPR\\
	\includegraphics[width=0.16\linewidth, height=0.12\linewidth, clip=true]{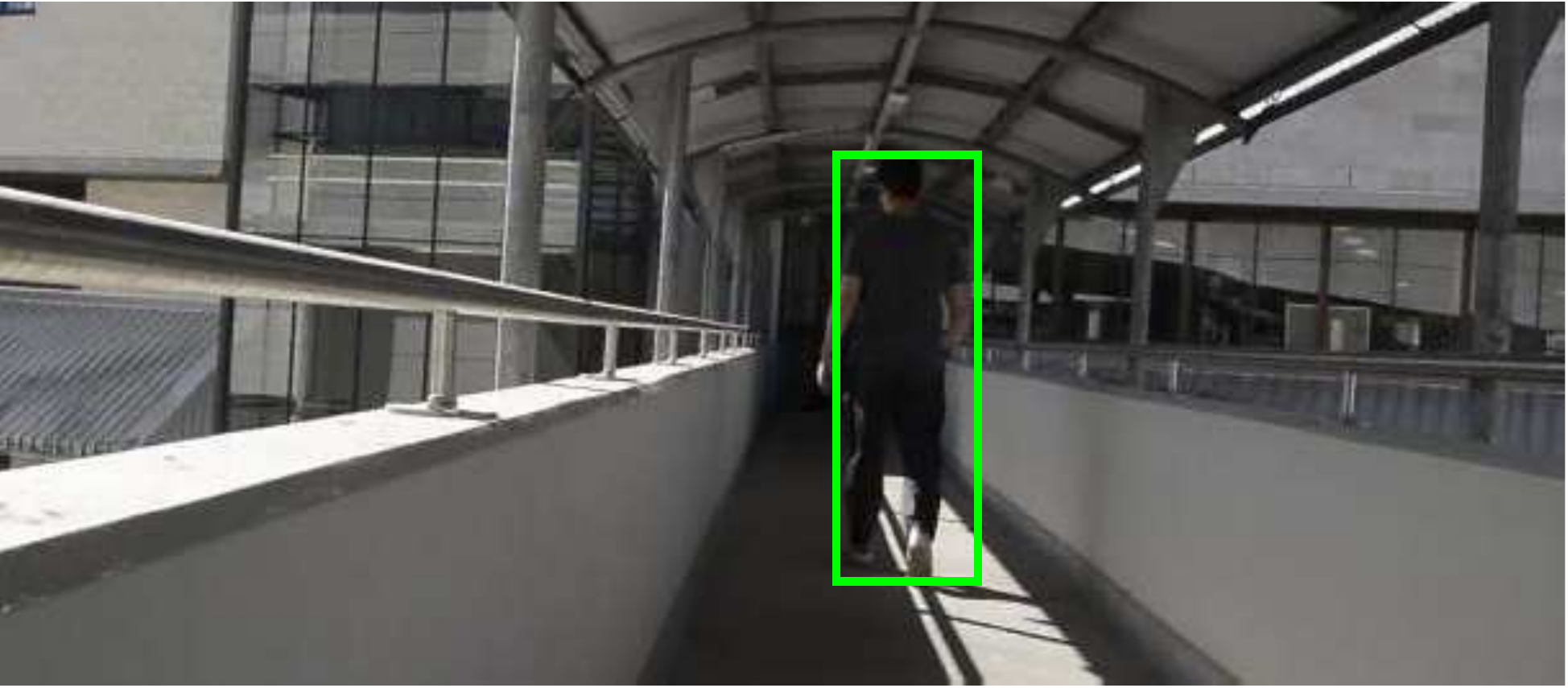}
		&\includegraphics[width=0.16\linewidth, height=0.12\linewidth, clip=true]{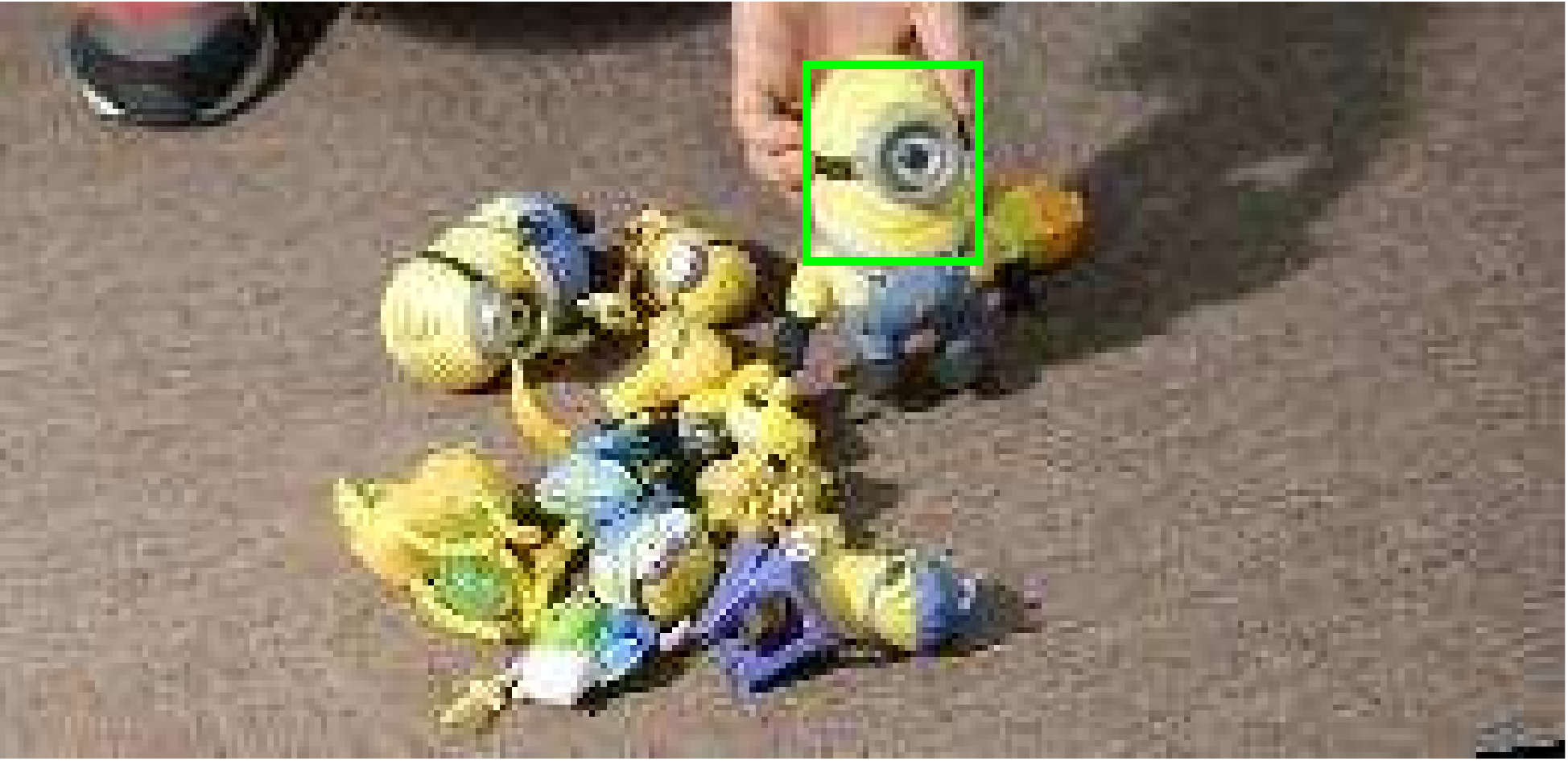}
	&\includegraphics[width=0.16\linewidth, height=0.12\linewidth, clip=true]{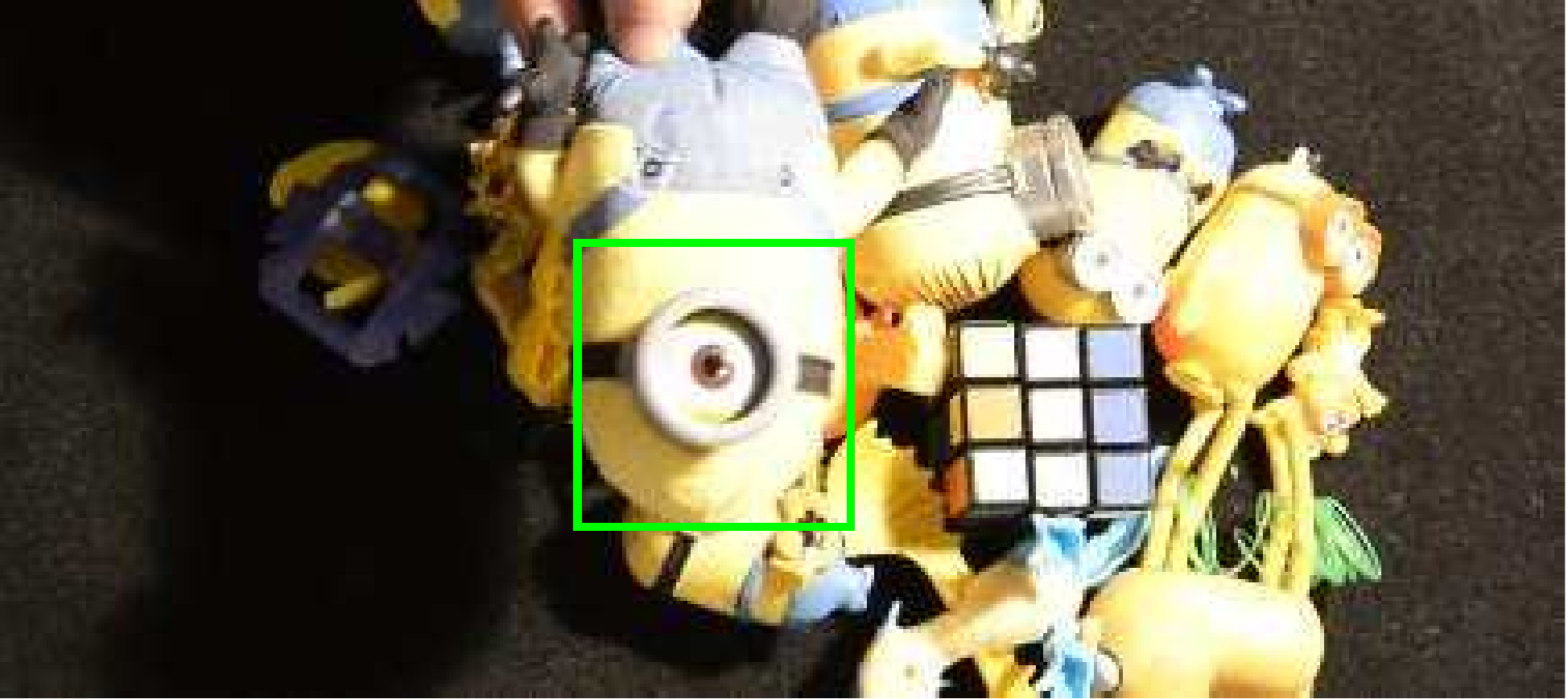}
	&\includegraphics[width=0.16\linewidth, height=0.12\linewidth, clip=true]{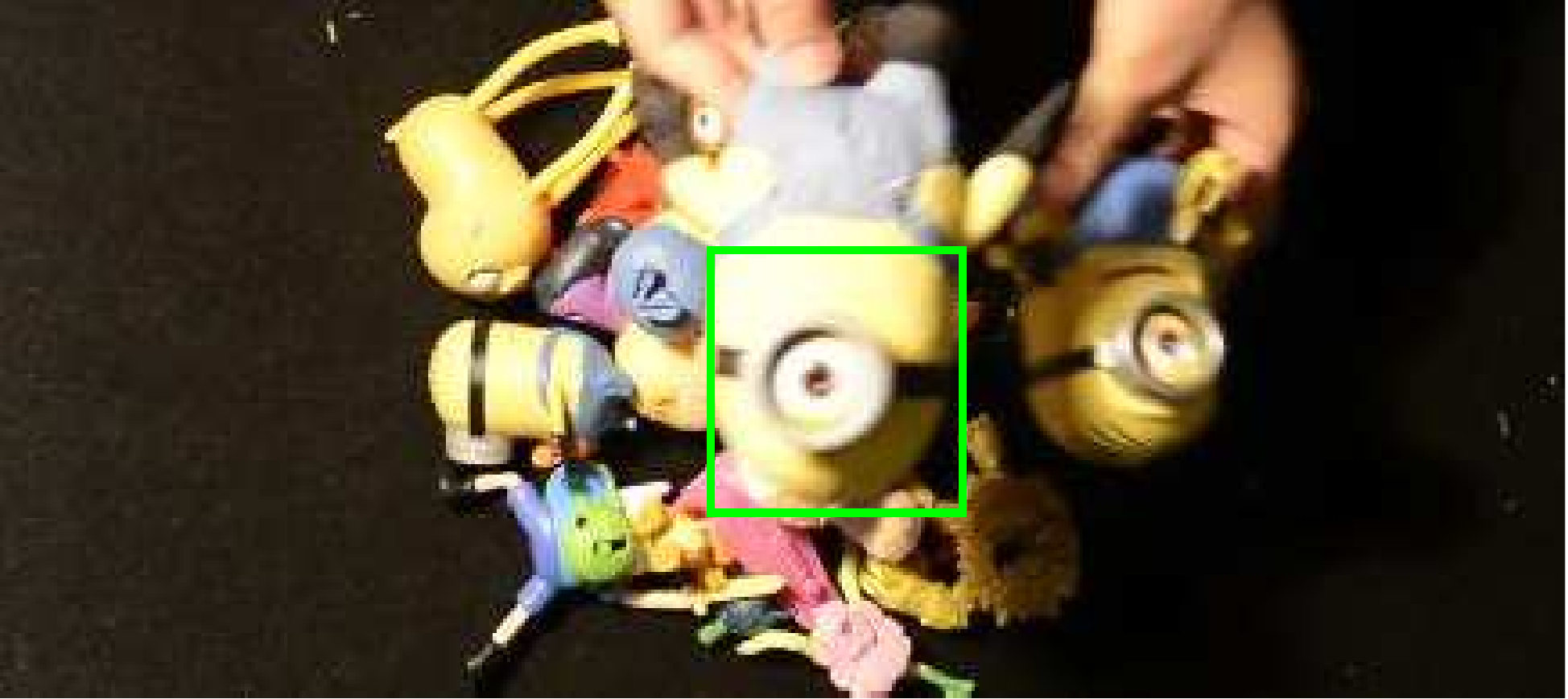}
	&\includegraphics[width=0.16\linewidth, height=0.12\linewidth, clip=true]{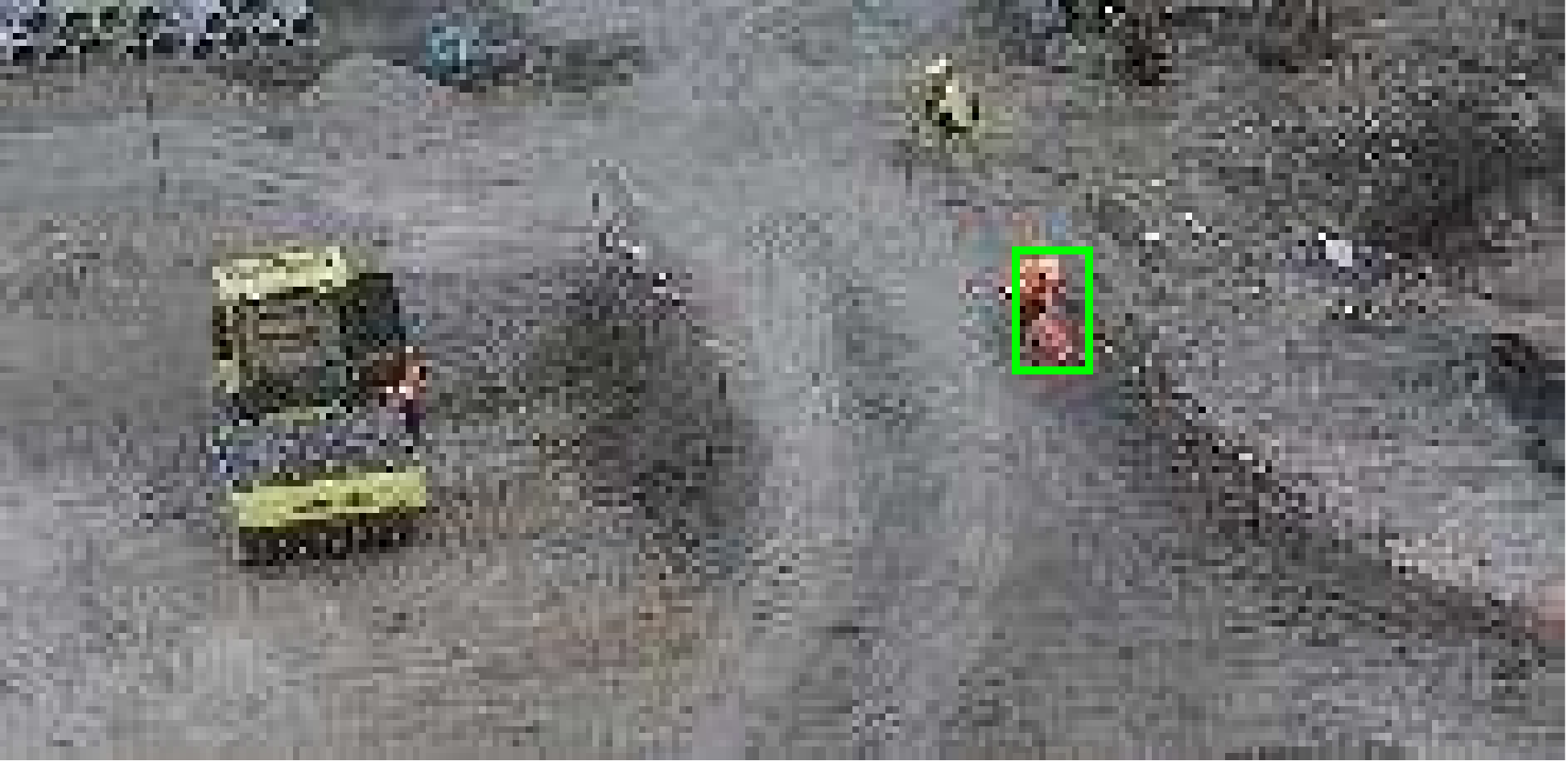}\\
	\textbf{Student}&\textbf{Toy1}&\textbf{Toy2}&\textbf{Toy3}&\textbf{Worker}\\
		IV, SV&BC, OCC&BC, SV, OV, OPR&BC, OCC, SV, IV, OPR, OV&SV, LR, BC\\
\end{tabular}
\end{table*}
\subsection{Qualitative Comparisons}
Here we provide qualitative evaluations of the competing trackers on sample hyperspectral or false-color videos, as shown in Fig.~\ref{fig:hsiQualitive}. Due to ensemble trackers, MCCT provides better performance in rotation (\emph{drive}), deformation (\emph{rubik}) and clutter conditions (\emph{forest}). BACF drifts away when a target is in low-resolution (\emph{worker}) and shares similar color as the background (\emph{forest}). Thanks to the high robustness of material properties, MHT shows higher robustness in tracking the objects  in these scenarios.

\section{Conclusion}\label{sec:con}
In this paper, we introduce a benchmark dataset for object tracking in hyperspectral videos and study the task from a new perspective where the material information is explored. The material information is embodied in the proposed SSHMG and abundance features. SSHMG encodes the local spectral-spatial texture information by summarising the occurrences of spectral-spatial multidimensional gradients orientation in local cubes of an HSI. Abundance describes detailed constituent material distribution using hyperspectral unmixing approach. Extensive experiments on the hyperspectral benchmark dataset demonstrate that the material properties contribute to moving object tracking especially in background cluster, object rotation and deformation scenes, confirming that material information in HSIs have great potential in object tracking. In our future work, we will develop a material based convolutional neural network to investigate deep spectral-spatial material information for object tracking.

\appendices
\section{Appendix}
Here, we show the first frame of all the sequences in the collected dataset in Table~\ref{tab:dataset}. We show the ground truth bounding box in the first frame of RGB video for more natural color visualization. Each video is also labeled with the challenging factors according to 11 attributes listed in~\cite{Wu2015}, including illumination variation (IV), scale variation (SV), occlusion (OCC), deformation (DEF), motion blur (MB), fast motion (FM), in-plane rotation (IPR), out-of-plane rotation (OPR), out-of-view (OV), background clutters (BC), and low resolution (LR). The whole dataset contains 35 color videos and 35 hyperspectral videos with an average of 500 frames in each sequence. The hyperspectral videos are further converted to false-color videos. The full dataset and benchmark including the sequences, annotations and associate code will be available online after the review process. Some preprocessing steps are listed as follows.

\begin{figure}[htbp]
    \centering
    \graphicspath{{figure/}}
    \includegraphics[width=0.5\linewidth]{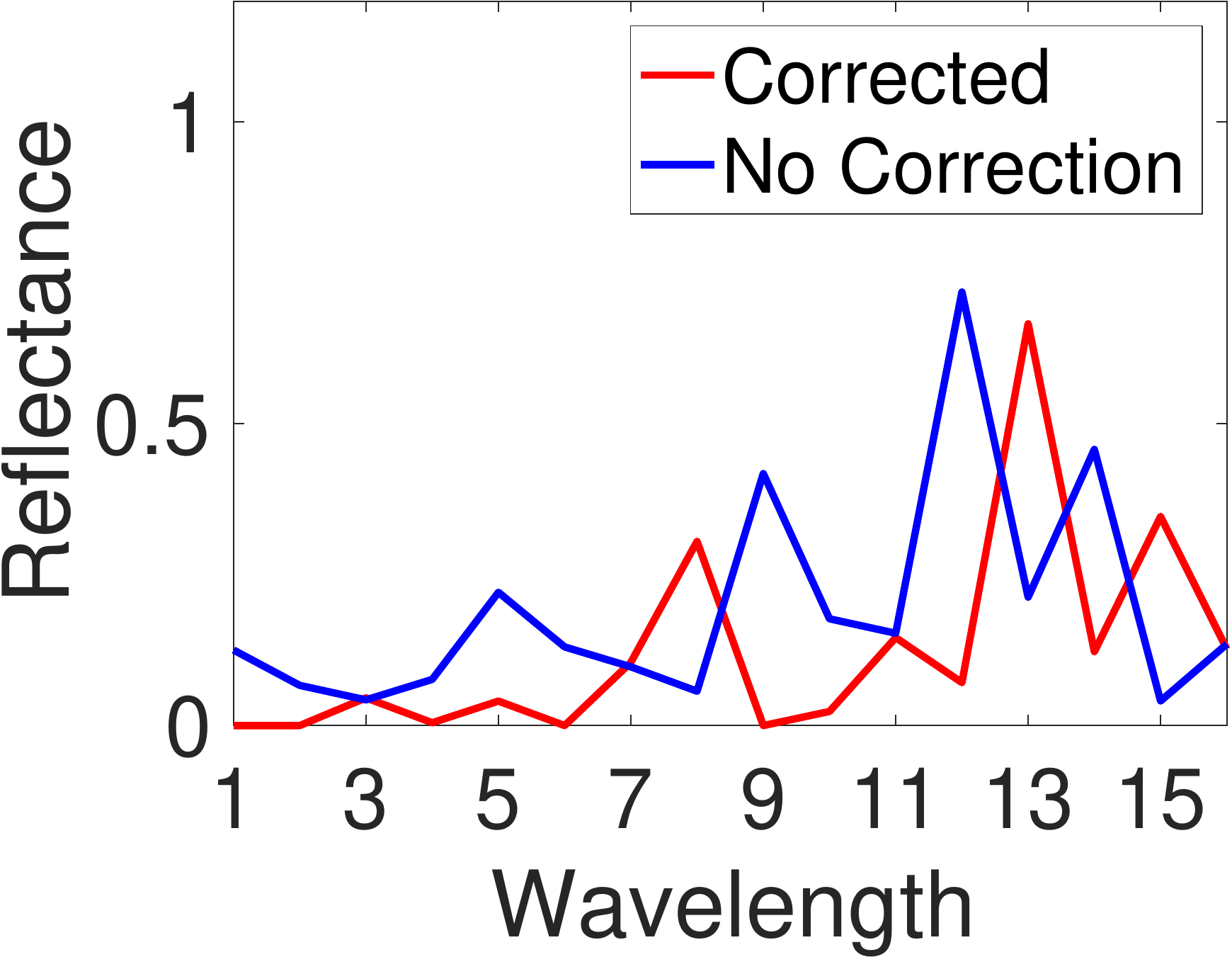}
    \caption{Effect of spectral correction. }\label{fig:SpectralCorrection}
\end{figure}

\textbf{Spectral Calibration: } Our calibration process involves two steps: dark calibration and spectral correction. Dark calibration aims to remove the effect of noises produced by the camera sensor. We performed dark calibration by subtracting a dark frame from the captured image, in which the dark frame was captured when lens was covered by a cap. The goal of spectral calibration is to suppress the contributions of unwanted second order responses, for example response to wavelengths leaking into the filters. By applying the sensor-specific spectral correction matrix on the acquired reflectance, the resulted spectrum is more consistent with the expected spectrum. An example is given in Fig.~\ref{fig:SpectralCorrection}. The blue curve is acquired spectrum without correction and the red curve is corrected spectrum.

\textbf{Color to Hyperspectral Image Registration: } We registered the hyperspectral sequences and color sequences to make them describe almost the same scene. Specifically, we manually selected some points in the initial frame of both hyperspectral and color videos as key points. These points were matched and then used for the geometrical transformation.  Finally, the resulting matrix was used to transform the color image in all the subsequent frames to align with corresponding hyperspectral frames.

\textbf{Hyperspectral to Color Image Conversion: } We converted hyperspectral videos to false color videos using CIE color matching functions (CMFs)~\footnote{http://cvrl.ioo.ucl.ac.uk/cmfs.htm}. The CMFs indicate the weights of each wavelength in a hyperspectral imagery (HSI) when they are used to generate red, green, and blue channels. Given an HSI $\X \in \Rbb^{N \times K}$ with $N$ pixels and $K$ bands, this step converts the HSI to a CIE XYZ image, formulated as
\begin{equation}
	\I=\sum \limits_{k=1}^{K} \X(:,k)\A(k,:)
\end{equation}
where $\A \in \Rbb^{K\times 3}$ are the CMFs. Here we used 10-deg XYZ CMFs transformed from the CIE (2006) with a step size of 0.1nm.  After that, $\I$ is converted to the default RGB colour space sRGB. Furthermore, the color transformation method in~\cite{Pitie2007} was applied to make the color intensity of converted image close to that of corresponding collected color frames.

\bibliographystyle{IEEEtran}
\bibliography{IEEEabrv,hsi-tracking-refs2}

\begin{thebibliography}{10}
\providecommand{\url}[1]{#1}
\csname url@samestyle\endcsname
\providecommand{\newblock}{\relax}
\providecommand{\bibinfo}[2]{#2}
\providecommand{\BIBentrySTDinterwordspacing}{\spaceskip=0pt\relax}
\providecommand{\BIBentryALTinterwordstretchfactor}{4}
\providecommand{\BIBentryALTinterwordspacing}{\spaceskip=\fontdimen2\font plus
\BIBentryALTinterwordstretchfactor\fontdimen3\font minus
  \fontdimen4\font\relax}
\providecommand{\BIBforeignlanguage}[2]{{%
\expandafter\ifx\csname l@#1\endcsname\relax
\typeout{** WARNING: IEEEtran.bst: No hyphenation pattern has been}%
\typeout{** loaded for the language `#1'. Using the pattern for}%
\typeout{** the default language instead.}%
\else
\language=\csname l@#1\endcsname
\fi
#2}}
\providecommand{\BIBdecl}{\relax}
\BIBdecl

\bibitem{Bolme2010}
D.~S. Bolme, J.~R. Beveridge, B.~A. Draper, and Y.~M. Lui, ``Visual object
  tracking using adaptive correlation filters,'' in \emph{Proc. IEEE Conf.
  Comput. Vis. Pattern Recognit. (CVPR)}, 2010.

\bibitem{Henriques2012}
J.~F. Henriques, R.~Caseiro, P.~Martins, and J.~Batista, ``Exploiting the
  circulant structure of tracking-by-detection with kernels,'' in \emph{Proc.
  European conference on computer vision (ECCV)}, 2012.

\bibitem{Danelljan2014}
M.~Danelljan, F.~S. Khan, M.~Felsberg, and J.~v.~d. Weijer, ``Adaptive color
  attributes for real-time visual tracking,'' in \emph{Proc. IEEE Conf. Comput.
  Vis. Pattern Recognit. (CVPR)}, June 2014.

\bibitem{Henriques2015}
J.~F. Henriques, R.~Caseiro, P.~Martins, and J.~Batista, ``High-speed tracking
  with kernelized correlation filters,'' \emph{IEEE Trans. Pattern Anal. Mach.
  Intell.}, vol.~37, no.~3, pp. 583--596, 2015.

\bibitem{Danelljan2015}
M.~Danelljan, G.~Häger, F.~S. Khan, and M.~Felsberg, ``Learning spatially
  regularized correlation filters for visual tracking,'' in \emph{Proc. IEEE
  Int. Conf. Comput. Vis. (ICCV)}, Dec 2015.

\bibitem{Mueller2017}
M.~Mueller, N.~Smith, and B.~Ghanem, ``Context-aware correlation filter
  tracking,'' in \emph{Proc. IEEE Conf. Comput. Vis. Pattern Recognit. (CVPR)},
  2017.

\bibitem{Danelljan2016a}
M.~Danelljan, G.~Hager, F.~Shahbaz~Khan, and M.~Felsberg, ``Adaptive
  decontamination of the training set: A unified formulation for discriminative
  visual tracking,'' in \emph{Proc. IEEE Conf. Comput. Vis. Pattern Recognit.
  (CVPR)}, June 2016.

\bibitem{Zhang2017}
S.~{Zhang}, X.~{Lan}, Y.~{Qi}, and P.~C. {Yuen}, ``Robust visual tracking via
  basis matching,'' \emph{IEEE Trans. Circuits Syst. Video Technol.}, vol.~27,
  no.~3, pp. 421--430, March 2017.

\bibitem{Choi2016}
J.~{Choi}, H.~J. {Chang}, J.~{Jeong}, Y.~{Demiris}, and J.~Y. {Choi}, ``Visual
  tracking using attention-modulated disintegration and integration,'' in
  \emph{Proc. IEEE Conf. Comput. Vis. Pattern Recognit. (CVPR)}, June 2016.

\bibitem{Gao2017}
J.~{Gao}, T.~{Zhang}, X.~{Yang}, and C.~{Xu}, ``Deep relative tracking,''
  \emph{IEEE Trans. Image Process.}, vol.~26, no.~4, pp. 1845--1858, April
  2017.

\bibitem{Feng2019}
W.~{Feng}, R.~{Han}, Q.~{Guo}, J.~{Zhu}, and S.~{Wang}, ``Dynamic
  saliency-aware regularization for correlation filter based object tracking,''
  \emph{IEEE Transactions on Image Processing}, 2019.

\bibitem{Liang2018}
J.~Liang, J.~Zhou, L.~Tong, X.~Bai, and B.~Wang, ``Material based salient
  object detection from hyperspectral images,'' \emph{Pattern Recognit.},
  vol.~76, pp. 476--490, 2018.

\bibitem{Schwartz2013}
G.~{Schwartz} and K.~{Nishino}, ``Visual material traits: Recognizing per-pixel
  material context,'' in \emph{Proc. IEEE Int. Conf. Comput. Vis. Workshops
  (ICCVW)}, Dec 2013.

\bibitem{Schwartz2015}
------, ``Automatically discovering local visual material attributes,'' in
  \emph{Proc. IEEE Conf. Comput. Vis. Pattern Recognit. (CVPR)}, 2015.

\bibitem{Degol2016}
J.~Degol, M.~Golparvar-Fard, and D.~Hoiem, ``Geometry-informed material
  recognition,'' in \emph{Proc. IEEE Conf. Comput. Vis. Pattern Recognit.
  (CVPR)}, 2016.

\bibitem{Bell2015}
S.~{Bell}, P.~{Upchurch}, N.~{Snavely}, and K.~{Bala}, ``Material recognition
  in the wild with the materials in context database,'' in \emph{Proc. IEEE
  Conf. Comput. Vis. Pattern Recognit. (CVPR)}, June 2015.

\bibitem{Cimpoi2015}
M.~{Cimpoi}, S.~{Maji}, and A.~{Vedaldi}, ``Deep filter banks for texture
  recognition and segmentation,'' in \emph{Proc. IEEE Conf. Comput. Vis.
  Pattern Recognit. (CVPR)}, June 2015.

\bibitem{Shiradkar2014}
R.~{Shiradkar}, L.~{Shen}, G.~{Landon}, S.~H. {Ong}, and P.~{Tan}, ``A new
  perspective on material classification and ink identification,'' in
  \emph{Proc. IEEE Conference on Computer Vision and Pattern Recognition}, June
  2014.

\bibitem{TominagaandTakahikoHoriuchi2010}
T.~H. Abdelhameed F.~Ibrahim, Shoji~Tominaga, ``Spectral imaging method for
  material classification and inspection of printed circuit boards,''
  \emph{Optical Engineering}, vol.~49, no.~5, 2010.

\bibitem{Tanaka2017}
K.~{Tanaka}, Y.~{Mukaigawa}, T.~{Funatomi}, H.~{Kubo}, Y.~{Matsushita}, and
  Y.~{Yagi}, ``Material classification using frequency-and depth-dependent
  time-of-flight distortion,'' in \emph{Proc. IEEE Conf. Comput. Vis. Pattern
  Recognit. (CVPR)}, July 2017.

\bibitem{Su2016}
S.~{Su}, F.~{Heide}, R.~{Swanson}, J.~{Klein}, C.~{Callenberg}, M.~{Hullin},
  and W.~{Heidrich}, ``Material classification using raw time-of-flight
  measurements,'' in \emph{Proc. IEEE Conf. Comput. Vis. Pattern Recognit.
  (CVPR)}, June 2016.

\bibitem{Uzair2015}
M.~Uzair, A.~Mahmood, and A.~Mian, ``Hyperspectral face recognition with
  spatiospectral information fusion and {PLS} regression,'' \emph{IEEE Trans.
  Image Process.}, vol.~24, no.~3, pp. 1127--1137, 2015.

\bibitem{Ye2017}
M.~Ye, Y.~Qian, J.~Zhou, and Y.~Y. Tang, ``Dictionary learning-based
  feature-level domain adaptation for cross-scene hyperspectral image
  classification,'' \emph{IEEE Trans. Geosci. Remote Sens.}, vol.~55, no.~3,
  pp. 1544--1562, 2017.

\bibitem{Al-khafaji2018}
S.~L. Al-khafaji, J.~Zhou, A.~Zia, and A.~W. Liew, ``Spectral-spatial scale
  invariant feature transform for hyperspectral images,'' \emph{IEEE Trans.
  Image Process.}, vol.~27, no.~2, pp. 837--850, 2018.

\bibitem{Wang2010}
T.~Wang, Z.~Zhu, and E.~Blasch, ``Bio-inspired adaptive hyperspectral imaging
  for real-time target tracking,'' \emph{IEEE Sens. J.}, vol.~10, no.~3, pp.
  647--654, March 2010.

\bibitem{Banerjee2009}
A.~Banerjee, P.~Burlina, and J.~Broadwater, ``Hyperspectral video for
  illumination-invariant tracking,'' in \emph{Proc. First Workshop on
  Hyperspectral Image and Signal Processing: Evolution in Remote Sensing
  (WHISPERS)}, Aug 2009.

\bibitem{Nguyen2010}
H.~V. Nguyen, A.~Banerjee, and R.~Chellappa, ``Tracking via object reflectance
  using a hyperspectral video camera,'' in \emph{Proc. IEEE Conf. Comput. Vis.
  Pattern Recognit. Workshops (CVPRW)}, 2010.

\bibitem{Uzkent2016a}
B.~Uzkent, M.~J. Hoffman, and A.~Vodacek, ``Integrating hyperspectral
  likelihoods in a multidimensional assignment algorithm for aerial vehicle
  tracking,'' \emph{IEEE J. Select. Topics Appl. Earth Observ. Remote Sens.},
  vol.~9, no.~9, pp. 4325--4333, Sept 2016.

\bibitem{Uzkent2018}
B.~Uzkent, A.~Rangnekar, and M.~J. Hoffman, ``Tracking in aerial hyperspectral
  videos using deep kernelized correlation filters,'' \emph{IEEE Transactions
  on Geoscience and Remote Sensing}, vol.~57, no.~1, pp. 449--461, 2019.

\bibitem{Qian2018}
K.~Qian, J.~Zhou, F.~Xiong, and H.~Zhou., ``Object tracking in hyperspectral
  videos with convolutional features and kernelized correlation filter,'' in
  \emph{Proc. International Conference on Smart Multimedia}, 2018.

\bibitem{Hu2011}
D.~Hu, L.~Bo, and X.~Ren, ``Toward robust material recognition for everyday
  objects.'' in \emph{Proc. British Machine Vision Conference (BMVC)}, 2011.

\bibitem{Wu2015}
Y.~Wu, J.~Lim, and M.~Yang, ``Object tracking benchmark,'' \emph{IEEE Trans.
  Pattern Anal. Mach. Intell.}, vol.~37, no.~9, pp. 1834--1848, 2015.

\bibitem{Qian2013}
Y.~Qian, M.~Ye, and J.~Zhou, ``Hyperspectral image classification based on
  structured sparse logistic regression and three-dimensional wavelet texture
  features,'' \emph{IEEE Trans. Geosci. Remote Sens.}, vol.~51, no.~4, pp.
  2276--2291, 2013.

\bibitem{Jia2017}
S.~Jia, J.~Hu, J.~Zhu, X.~Jia, and Q.~Li, ``Three-dimensional local binary
  patterns for hyperspectral imagery classification,'' \emph{IEEE Trans.
  Geosci. Remote Sens.}, vol.~55, no.~4, pp. 2399--2413, 2017.

\bibitem{Sun2018}
C.~Sun, D.~Wang, H.~Lu, and M.-H. Yang, ``Learning spatial-aware regressions
  for visual tracking,'' in \emph{Proc. IEEE Conf. Comput. Vis. Pattern
  Recognit. (CVPR)}, 2018.

\bibitem{Sui2018}
Y.~Sui, G.~Wang, L.~Zhang, and M.~Yang, ``Exploiting spatial-temporal locality
  of tracking via structured dictionary learning,'' \emph{IEEE Trans. Image
  Process.}, vol.~27, no.~3, pp. 1282--1296, 2018.

\bibitem{Choi2018}
J.~Choi, H.~J. Chang, T.~Fischer, S.~Yun, K.~Lee, J.~Jeong, Y.~Demiris, and
  J.~Y. Choi, ``Context-aware deep feature compression for high-speed visual
  tracking,'' in \emph{Proc. IEEE Conf. Comput. Vis. Pattern Recognit. (CVPR)},
  2018, pp. 479--488.

\bibitem{Nascimento2005}
J.~M.~P. Nascimento and J.~M.~B. Dias, ``Vertex component analysis: a fast
  algorithm to unmix hyperspectral data,'' \emph{IEEE Trans. Geosci. Remote
  Sens.}, vol.~43, no.~4, pp. 898--910, 2005.

\bibitem{Chang2010}
C.~Chang, C.~Wu, C.~Lo, and M.~Chang, ``Real-time simplex growing algorithms
  for hyperspectral endmember extraction,'' \emph{IEEE Trans. Geosci. Remote
  Sens.}, vol.~48, no.~4, pp. 1834--1850, 2010.

\bibitem{Chouzenoux2014}
E.~Chouzenoux, M.~Legendre, S.~Moussaoui, and J.~Idier, ``Fast constrained
  least squares spectral unmixing using primal-dual interior-point
  optimization,'' \emph{IEEE J. Sel. Top. Appl. Earth Obs. Remote Sens.},
  vol.~7, no.~1, pp. 59--69, 2014.

\bibitem{Heylen2011}
R.~Heylen, D.~Burazerovic, and P.~Scheunders, ``Fully constrained least squares
  spectral unmixing by simplex projection,'' \emph{IEEE Trans. Geosci. Remote
  Sens.}, vol.~49, no.~11, pp. 4112--4122, 2011.

\bibitem{Wang2015c}
Y.~Wang, C.~Pan, S.~Xiang, and F.~Zhu, ``Robust hyperspectral unmixing with
  correntropy-based metric,'' \emph{IEEE Trans. Image Process.}, vol.~24,
  no.~11, pp. 4027--4040, 2015.

\bibitem{Qian2017}
Y.~Qian, F.~Xiong, S.~Zeng, J.~Zhou, and Y.~Y. Tang, ``Matrix-vector
  nonnegative tensor factorization for blind unmixing of hyperspectral
  imagery,'' \emph{IEEE Trans. Geosci. Remote Sens.}, vol.~55, no.~3, pp.
  1776--1792, 2017.

\bibitem{Iordache2014}
M.~Iordache, J.~M. Bioucas-Dias, and A.~Plaza, ``Collaborative sparse
  regression for hyperspectral unmixing,'' \emph{IEEE Trans. Geosci. Remote
  Sens.}, vol.~52, no.~1, pp. 341--354, 2014.

\bibitem{Zhang2018a}
S.~Zhang, J.~Li, H.~Li, C.~Deng, and A.~Plaza, ``Spectral-spatial weighted
  sparse regression for hyperspectral image unmixing,'' \emph{IEEE Trans.
  Geosci. Remote Sens.}, vol.~56, no.~6, pp. 3265--3276, 2018.

\bibitem{Bioucas-Dias2008}
J.~M. Bioucas-Dias and J.~M.~P. Nascimento, ``Hyperspectral subspace
  identification,'' \emph{IEEE Trans. Geosci. Remote Sens.}, vol.~46, no.~8,
  pp. 2435--2445, 2008.

\bibitem{Galoogahi2017}
H.~K. Galoogahi, A.~Fagg, and S.~Lucey, ``Learning background-aware correlation
  filters for visual tracking,'' in \emph{Proc. IEEE Conf. Comput. Vis. Pattern
  Recognit. (CVPR)}, 2017.

\bibitem{Lukezic2017}
A.~Lukežic, T.~Vojír, L.~C. Zajc, J.~Matas, and M.~Kristan, ``Discriminative
  correlation filter with channel and spatial reliability,'' in \emph{Proc.
  IEEE Conf. Comput. Vis. Pattern Recognit. (CVPR)}, July 2017.

\bibitem{Felzenszwalb2010}
P.~F. Felzenszwalb, R.~B. Girshick, D.~McAllester, and D.~Ramanan, ``Object
  detection with discriminatively trained part-based models,'' \emph{IEEE
  Trans. Pattern Anal. Mach. Intell.}, vol.~32, no.~9, pp. 1627--1645, 2010.

\bibitem{Danelljan2017a}
M.~Danelljan, G.~Häger, F.~S. Khan, and M.~Felsberg, ``Discriminative scale
  space tracking,'' \emph{IEEE Trans. Pattern Anal. Mach. Intell.}, vol.~39,
  no.~8, pp. 1561--1575, 2017.

\bibitem{Hong2015}
Z.~Hong, Z.~Chen, C.~Wang, X.~Mei, D.~Prokhorov, and D.~Tao, ``Multi-store
  tracker ({MUSTer}): A cognitive psychology inspired approach to object
  tracking,'' in \emph{Proc. IEEE Conf. Comput. Vis. Pattern Recognit. (CVPR)},
  2015.

\bibitem{Li2014}
Y.~Li and J.~Zhu, ``A scale adaptive kernel correlation filter tracker with
  feature integration,'' in \emph{Proc. European conference on computer vision
  Workshop (ECCVW)}, 2014.

\bibitem{Hare2016}
S.~Hare, S.~Golodetz, A.~Saffari, V.~Vineet, M.~Cheng, S.~L. Hicks, and
  P.~H.~S. Torr, ``Struck: Structured output tracking with kernels,''
  \emph{IEEE Trans. Pattern Anal. Mach. Intell.}, vol.~38, no.~10, pp.
  2096--2109, 2016.

\bibitem{Zhang2016}
K.~Zhang, Q.~Liu, Y.~Wu, and M.~Yang, ``Robust visual tracking via
  convolutional networks without training,'' \emph{IEEE Trans. Image Process.},
  vol.~25, no.~4, pp. 1779--1792, 2016.

\bibitem{Wang2018}
N.~Wang, W.~Zhou, Q.~Tian, R.~Hong, M.~Wang, and H.~Li, ``Multi-cue correlation
  filters for robust visual tracking,'' in \emph{Proc. IEEE Conf. Comput. Vis.
  Pattern Recognit. (CVPR)}, 2018.

\bibitem{Danelljan2016}
M.~Danelljan, A.~Robinson, F.~S. Khan, and M.~Felsberg, ``Beyond correlation
  filters: Learning continuous convolution operators for visual tracking,'' in
  \emph{Proc. European Conference on Computer Vision (ECCV)}, 2016.

\bibitem{Danelljan2017}
M.~Danelljan, G.~Bhat, F.~S. Khan, and M.~Felsberg, ``{ECO}: Efficient
  convolution operators for tracking.'' in \emph{Proc. IEEE Conf. Comput. Vis.
  Pattern Recognit. (CVPR)}, 2017.

\bibitem{Ma2015}
C.~Ma, J.~Huang, X.~Yang, and M.~Yang, ``Hierarchical convolutional features
  for visual tracking,'' in \emph{Proc. IEEE Int. Conf. Comput. Vis. (ICCV)},
  Dec 2015.

\bibitem{Valmadre2017a}
J.~Valmadre, L.~Bertinetto, J.~Henriques, A.~Vedaldi, and P.~H.~S. Torr,
  ``End-to-end representation learning for correlation filter based tracking,''
  in \emph{Proc. IEEE Conf. Comput. Vis. Pattern Recognit. (CVPR)}, 2017.

\bibitem{Qi2016}
Y.~Qi, S.~Zhang, L.~Qin, H.~Yao, Q.~Huang, J.~Lim, and M.~Yang, ``Hedged deep
  tracking,'' in \emph{Proc. IEEE Conf. Comput. Vis. Pattern Recognit. (CVPR)},
  June 2016.

\bibitem{Guo2017a}
Q.~Guo, W.~Feng, C.~Zhou, R.~Huang, L.~Wan, and S.~Wang, ``Learning dynamic
  siamese network for visual object tracking,'' in \emph{Proc. IEEE Int. Conf.
  Comput. Vis. (ICCV)}, 2017.

\bibitem{Danelljan2015a}
M.~Danelljan, G.~Häger, F.~S. Khan, and M.~Felsberg, ``Convolutional features
  for correlation filter based visual tracking,'' in \emph{Proc. IEEE Int.
  Conf. Comput. Vis. Workshop (ICCVW)}, 2015.

\bibitem{Pitie2007}
F.~Pitie and A.~Kokaram, ``The linear monge-kantorovitch linear colour mapping
  for example-based colour transfer,'' in \emph{Proc. 4th European Conference
  on Visual Media Production}, Nov 2007.

\end{thebibliography}
\end{document}